\documentclass[pdflatex,sn-nature]{sn-jnl}
\usepackage{mhchem}
\usepackage{geometry}
\usepackage{graphicx} 
\usepackage{multirow} 
\usepackage{amsmath,amssymb,amsfonts} 
\usepackage{amsthm} 
\usepackage{mathrsfs} 
\usepackage[title]{appendix} 
\usepackage{xcolor} 
\usepackage{textcomp} 
\usepackage{caption} 
\usepackage{capt-of}
\usepackage{textcomp} 
\usepackage{manyfoot} 
\usepackage{booktabs} 
\usepackage{algorithm} 
\usepackage{algorithmicx} 
\usepackage{algpseudocode} 
\usepackage{listings} 
\usepackage{subcaption}
\usepackage{longtable}
\usepackage{array}
\usepackage{enumitem}
\usepackage{fontawesome5}

\usepackage[most]{tcolorbox}
\usepackage{xcolor}
\usepackage{fvextra}
\usepackage{caption}
\definecolor{inputbg}{HTML}{F7F8FA}
\definecolor{cotbg}{HTML}{FFFFFF}
\definecolor{finalbg}{HTML}{F2F8F2}
\definecolor{boxline}{HTML}{222222}

\newtcolorbox{fullcotsection}[2]{
enhanced,
breakable,
colback=#1,
colframe=boxline,
boxrule=0.35pt,
arc=1pt,
left=6pt,
right=6pt,
top=5pt,
bottom=5pt,
title={#2},
fonttitle=\bfseries\small,
coltitle=black,
colbacktitle=gray!12,
attach boxed title to top left={xshift=6pt,yshift=-2pt},
boxed title style={boxrule=0.3pt,arc=1pt}
}

\DefineVerbatimEnvironment{cotverbatim}{Verbatim}{
  breaklines=true,
  breakanywhere=true,
  breaksymbolleft={},
  breaksymbolright={},
  fontsize=\scriptsize,
  baselinestretch=0.93,
  commandchars=\\\{\}
}

\DefineVerbatimEnvironment{PromptBlock}{Verbatim}{
  breaklines=true,
  breaksymbolleft={},
  fontsize=\scriptsize,
  frame=single,
  rulecolor=\color{black!25},
  framesep=2mm
}

\definecolor{Ink}{HTML}{202733}
\definecolor{MutedInk}{HTML}{596579}
\definecolor{RuleGray}{HTML}{CBD5E1}
\definecolor{TableStripe}{HTML}{F6F8FC}
\definecolor{TableHeader}{HTML}{243B53}
\definecolor{PanelBg}{HTML}{F8FAFC}
\definecolor{PromptBg}{HTML}{F4F7FB}
\definecolor{MaterialsMain}{HTML}{207567}
\definecolor{MaterialsSoft}{HTML}{E8F5F1}
\definecolor{GOMain}{HTML}{7952B3}
\definecolor{GOSoft}{HTML}{F2ECFA}
\definecolor{RetroMain}{HTML}{B85C00}
\definecolor{RetroSoft}{HTML}{FFF3E2}
\definecolor{ScoreMain}{HTML}{1F6FEB}
\definecolor{ScoreSoft}{HTML}{EAF2FF}

\newcommand{\sampletitle}[2]{\subsection{#1: #2}}

\newcommand{\scoreblank}{\underline{\hspace{1.2cm}}}
\newcommand{\smiles}[1]{\begingroup\scriptsize\ttfamily\nolinkurl{#1}\endgroup}
\newcolumntype{P}[1]{>{\raggedright\arraybackslash}p{#1}}

\newcommand{\projName}[0]{SciReasoner}
\newcommand{\totalTasks}[0]{86}
\newcommand{\totalSOTATasks}[0]{67}
\newcommand{\totalSOTATasksExperts}[0]{26}

\makeatletter
\newcommand{\printprojectpage}{ 
  \par\vskip-12pt
  {\centering\large\bfseries \href{https://scireasoner.github.io}{\faIcon{globe}\hspace{0.35em}SciReasoner.github.io}\par} 
  \vskip-10pt
}
\let\originalprintabstract\printabstract
\renewcommand{\printabstract}{\printprojectpage\originalprintabstract}
\makeatother

\theoremstyle{thmstyleone}

\theoremstyle{thmstyletwo}

\theoremstyle{thmstylethree}

\begin{document}

\title[\projName]{Accurate, Interdisciplinary and Transparent Structure-property Understanding with Deep Native Structural Reasoning}

\author[1,2]{\fnm{Chen} \sur{Tang}}
\equalcont{These authors contributed equally to this work.}

\author[1,2]{\fnm{Yizhou} \sur{Wang}}
\equalcont{These authors contributed equally to this work.}

\author[1,3]{\fnm{Jianyu} \sur{Wu}}
\equalcont{These authors contributed equally to this work.}

\author[1,5]{\fnm{Lintao} \sur{Wang}}
\equalcont{These authors contributed equally to this work.}

\author[2,1]{\fnm{Shixiang} \sur{Tang}}\email{shixiangtang@cuhk.edu.hk}

\author[1,4]{\fnm{Pengze} \sur{Li}}

\author[1,8]{\fnm{Encheng} \sur{Su}}

\author[1,8]{\fnm{Jun} \sur{Yao}}

\author[1,2]{\fnm{Jiabei} \sur{Xiao}}

\author[9,10]{\fnm{Yuqi} \sur{Shi}}

\author[1]{\fnm{Jielan} \sur{Li}}

\author[1]{\fnm{Hongxia} \sur{Hao}}

\author[1]{\fnm{Zhangyang} \sur{Gao}}

\author[11]{\fnm{Fang} \sur{Wu}}

\author[1,2]{\fnm{Ben} \sur{Fei}}

\author[2]{\fnm{Xiangyu} \sur{Yue}}

\author[1]{\fnm{Pan} \sur{Tan}}

\author[3]{\fnm{Bozitao} \sur{Zhong}}

\author[1]{\fnm{Jinouwen} \sur{Zhang}}

\author[1]{\fnm{Aoran} \sur{Wang}}

\author[2,1]{\fnm{Yan} \sur{Lu}}

\author[6,1]{\fnm{Jiaheng} \sur{Liu}}

\author[1]{\fnm{Xinzhu} \sur{Ma}}

\author[3]{\fnm{Liang} \sur{Hong}}

\author[9,10]{\fnm{Mingyue} \sur{Zheng}}

\author[7]{\fnm{Phil} \sur{Torr}}

\author[1]{\fnm{Bowen} \sur{Zhou}}

\author[1,2]{\fnm{Wanli} \sur{Ouyang}}\email{ouyangwanli@pjlab.org.cn}

\author[1]{\fnm{Lei} \sur{Bai}}\email{bailei@pjlab.org.cn}

\affil[1]{Shanghai Artificial Intelligence Laboratory, China}
\affil[2]{\orgname{The Chinese University of Hong Kong}, Hong Kong}
\affil[3]{Shanghai Jiao Tong University, China}
\affil[4]{\orgname{Fudan University}, China}
\affil[5]{\orgname{University of Sydney}, Australia}
\affil[6]{\orgname{Nanjing University}, China}
\affil[7]{\orgname{University of Oxford}, UK}
\affil[8]{\orgname{The University of Science and Technology of China}, China}
\affil[9]{\orgname{Drug Discovery and Design Center, State Key Laboratory of Drug Research, Shanghai Institute of Materia Medica, Chinese Academy of Sciences}, China}

\affil[10]{\orgname{University of Chinese Academy of Sciences}, China}
\affil[11]{\orgname{Stanford University}, USA}

\abstract{

Structure--property relationships are foundational to biology, chemistry and materials science, where function, reactivity and physical response emerge from spatial, chemical and periodic organization. Mechanistically explaining these relationships requires interpreting structural evidence through scientific principles and physical constraints, from stereochemistry and bonding to symmetry, energetics and periodic order. However, applying artificial intelligence to this process presents a joint challenge of representation and reasoning: models must preserve domain-native structural information while showing how specific evidence supports predictions under these constraints. 
Here we introduce \projName{}, a multimodal scientific foundation model for native structural reasoning across proteins, small molecules and inorganic crystals. \projName{} discretizes coordinates, topologies and periodic connectivities into a unified structure-aware vocabulary, treating structural tokens as addressable evidence units within autoregressive reasoning trajectories. We evaluate \projName{} in settings where shortcut correlations are weakened and structure-grounded inference is essential. 
In homology-controlled Gene Ontology prediction, \projName{} improves Cellular Component annotation for low-homology and orphan-like proteins, increasing $F_{\max}$ from 0.42 to 0.55.
In chemistry, it raises single-step retrosynthesis accuracy from 0.63 to 0.72 while generating fragment-level disconnection and precursor-verification traces. In materials science, its representations separate elemental and compound phases and resolve high- and low-band-gap regimes. Across \totalTasks{} benchmarks, \projName{} achieves state-of-the-art performance on \totalSOTATasks{} tasks. Double-blind expert evaluation rates its reasoning traces as preferred or at least comparable to those of a frontier large language model in 98\% of cases. By making structure an inspectable substrate for reasoning under scientific constraints, \projName{} connects accurate prediction with interpretable scientific inference.

}

\maketitle

\section{Introduction}\label{sec1}

Structure--property relationships~\citep{thakur2026composition,le2012quantitative,van2008complete,creighton1993proteins,tai2022leveraging,osadchy2011maps} are foundational to the physical and biological sciences. Across proteins, small molecules and crystalline materials, observable functions and properties arise from the spatial, chemical and periodic organization of matter. 
In proteins, protein structures and conformations, long-range interactions and active-site geometry shape their biological functions~\citep{anfinsen1973principles,abramson2024accurate}; in chemicals, bonding, functional groups and stereochemistry contribute to their reactivity and molecular behaviour; and in materials, lattice symmetry, coordination motifs and periodic bonding networks dictate stability and electronic properties~\citep{xie2018crystal,merchant2023scaling}. 
Unraveling how these organizing principles connect structure to function, reactivity and physical response is therefore central to functional annotation, molecular design and materials discovery.

Yet, explaining these relationships mechanistically is rarely a straightforward pattern-matching problem. Evidence for a function, reaction or material property is distributed across local motifs, non-local contacts, chemical environments, conformational geometry and long-range periodic order. When annotating a protein or planning a synthesis route~\citep{strieth2024artificial,segler2018planning}, human experts integrate heterogeneous cues, compare competing structural hypotheses and apply biochemical, chemical or crystallographic constraints~\citep{duek2021functionathon,zhao2024efficient}. Artificial intelligence (AI) can accelerate this process by retrieving knowledge and screening large structural spaces, but it faces a deeper representational challenge: to support expert-like scientific reasoning, a model must represent biological, chemical and materials structures as native objects and use them as evidence in explicit, inspectable reasoning.

Current scientific AI systems only partially meet this requirement because they often separate native structural representation from evidence-linked reasoning. Large language models provide a flexible interface to scientific knowledge, but when proteins, molecules and crystals are cast primarily as text, structural organization is compressed into strings or descriptions, making the resulting explanations depend largely on linguistic associations rather than directly addressable physical evidence~\citep{achiam2023gpt,mirza2025superhuman,narayanan2026ether0}. Agentic systems extend such models through retrieval, tool use and workflow coordination~\citep{boiko2023autonomous,ghareeb2026multi,aygun2026ai,gottweis2026accelerating}, yet their scientific depth remains constrained by the structural competence of the foundation models they orchestrate. Domain-specific models, in contrast, often encode molecular graphs, protein structures or crystal lattices directly~\citep{wang2023scientific,xie2018crystal,yang2019analyzing}, but are usually optimized as task-specific predictors that output scores, labels or scalar properties without exposing the intermediate evidence behind their decisions. Thus, a critical gap remains: building a new reasoning paradigm that natively represents multidimensional scientific structures while producing reasoning traces that can be inspected against those structures.

Here we propose native structural reasoning as a foundation-model paradigm for structure--property analysis across biology, chemistry and materials science. We instantiate this paradigm in \projName{}, a multimodal scientific foundation model that represents proteins, small molecules and periodic crystals through a unified structure-aware vocabulary. By native structural reasoning, we mean that structural tokens are not auxiliary descriptors appended to language, but addressable evidence units that can be combined, cited and checked within a generated reasoning trajectory. To this end, \projName{} discretizes coordinates, topologies and periodic connectivities into scientific tokens that preserve domain-native information, including local geometry, stereochemistry, residue-level structural motifs, lattice symmetry and crystallographic bonding patterns. These tokens are integrated with language instructions in a single autoregressive model, enabling trajectories that interleave explanatory text with specific residues, molecular fragments, conformational cues or crystal descriptors.

This design changes the role of a scientific foundation model. Rather than predicting an answer after reading a structural prompt, \projName{} generates an inspectable chain in which intermediate claims can be traced to explicit structural evidence. 
We test this capability in regimes where shortcut correlations are unreliable. 
 
In the Cellular Component aspect of protein Gene Ontology prediction, \projName{} shows its largest gains in low-homology and orphan-like proteins, where evolutionary transfer provides weak evidence, improving $F_{\max}$ from 0.42 to 0.55.
Its attention is enriched at contact-defined DNA-binding residues and localizes to protein--DNA interfaces, suggesting that functional predictions are supported by residues that physically mediate the molecular interaction. 
In retrosynthesis, \projName{} raises single-step accuracy from 0.63 to 0.72 while decomposing products into reaction-relevant fragments, identifying strategic bond disconnections and verifying chemically valid precursors. 
In crystalline materials, the reasoning trajectories of \projName{} assign higher importance to symmetry-distinct sites, local coordination environments and periodic bonding motifs associated with stability and band-gap variation, suggesting that it links electronic and thermodynamic behaviour to the structural features that physically govern them. 
Across these examples, the common behaviour is that structural organization acts as the working substrate of inference rather than a peripheral input descriptor.

Broad evaluation supports the generality of this structure-grounded modelling strategy. Across \totalTasks{} benchmarks spanning proteins, DNA, RNA, small molecules, inorganic crystals, scientific question answering, property prediction and generation tasks, \projName{} achieves state-of-the-art performance on \totalSOTATasks{} tasks. Double-blind expert evaluation further indicates that its reasoning traces are useful for scientific inspection, with experts preferring or tying \projName{} over a frontier large language model in the overwhelming majority of comparisons. By representing structures as evidence-bearing units and linking them to inspectable conclusions, \projName{} bridges accurate prediction and interpretable scientific inference, suggesting a route toward scientific AI systems that do not merely predict what property a structure has, but begin to explain why that structure gives rise to a particular function, reaction or material behaviour.

\captionsetup[figure]{
  justification=raggedright,
  singlelinecheck=false
}
\begin{center}
    \includegraphics[width=1\linewidth]{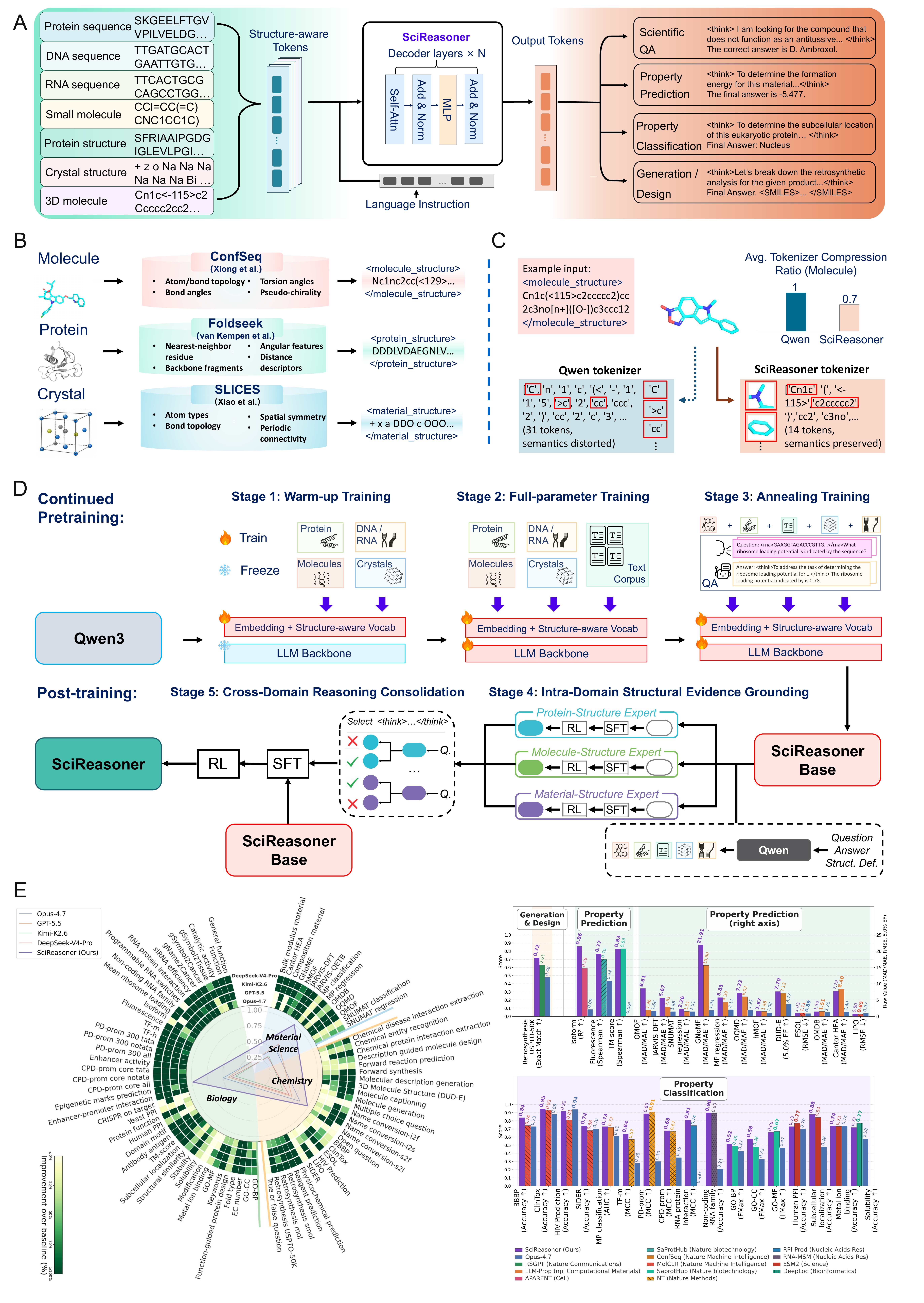}
\end{center}

\captionof{figure}{\textbf{Overview of SciReasoner, a unified multimodal foundation model for native structural reasoning across biomolecules, small molecules, and periodic crystals.} \textbf{(A)} Structure-aware scientific reasoning interface. SciReasoner encodes heterogeneous scientific entities, including protein/DNA/RNA sequences, small molecules, protein structures, crystal structures, and 3D molecules, into structure-aware tokens. These tokens are combined with language instructions and processed by an LLM backbone to support structure-grounded scientific QA, property prediction, property classification, and generation/design tasks. 
\textbf{(B)} Structure-aware token construction. Domain-specific encoders convert scientific objects into structured textual representations: ConfSeq for molecules, Foldseek for proteins, and SLICES for crystals. These representations preserve chemically and physically meaningful information, including molecular topology, spatial geometry, local motifs, lattice features, symmetry, and space-group information. 
\textbf{(C)} Tokenizer compression and semantic preservation. Compared with the Qwen tokenizer, the \projName{} tokenizer reduces token length for molecular inputs while preserving chemically meaningful substructures. This indivisible, structure-aware vocabulary mitigates semantic fragmentation caused by standard sub-word tokenization. 
\textbf{(D)} Continued Pretraining and post-training pipeline. \projName{} is initialized from a Qwen model and trained through warm-up alignment, full-parameter multimodal training, and annealing tuning across scientific modalities. During post-training, intra-domain structural evidence grounding first trains task-specialized experts to use structural tokens as reasoning evidence within each domain. Cross-domain reasoning consolidation then integrates the expert-generated reasoning traces and expert capabilities into the final \projName{} model.
\textbf{(E)} Benchmark performance. \projName{} is evaluated across diverse scientific QA, biology, chemistry, materials science, and generation/design benchmarks. The results demonstrate broad improvements over generalist language models and domain-specific baselines, highlighting \projName{}’s ability to connect structural evidence with function, property, and design decisions.}
\label{fig:main_figure}

\section{Results}\label{sec2}
\subsection{Overview of \projName}

We present \projName, a unified multimodal foundation model capable of explicit scientific sequence, structure, and textual reasoning within a single coherent output trajectory (Fig.~\ref{fig:main_figure}A). \projName~serves as a unified reasoning engine that maps diverse scientific inputs, including proteins, DNA/RNA, small molecules, and periodic crystals, into a shared cross-modal latent space. Unlike conventional computational approaches in materials and biology, this architecture provides a unified, mechanism-aware framework. 

To integrate explicit 3D structural information into a linguistic token space, \projName~designs a structure-aware vocabulary designed to preserve physical and biochemical integrity (Fig.~\ref{fig:main_figure}A). The model utilizes unified tokens representing local motifs, 3D geometry, atomic bond types, and crystal space groups (Fig.~\ref{fig:main_figure}B). This resolves the severe limitations of conventional sub-word tokenization algorithms, which arbitrarily fragment molecular graphs and motifs, thereby destroying scientific semantics and corrupting the training process (Fig.~\ref{fig:main_figure}C).

To align this structural token space with natural language, we perform continued pretraining with a multi-stage training pipeline (Fig.~\ref{fig:main_figure}D, Top). 
Stage~1 serves as a warm-up training phase that anchors the newly introduced structure-aware tokens to basic topological, geometric, and chemical semantics without destabilizing the pretrained language space. 
This establishes a stable structural--linguistic interface, which is further developed through Stage~2 full-parameter multimodal training and Stage~3 annealed training with an increased proportion of question--answer (QA)-style data, thereby supporting native structural reasoning. 

To enable robust multi-step reasoning, we developed a post-training framework named self-bootstrapped native structural reasoning (Fig.~\ref{fig:main_figure}D, Bottom). The post-training procedure uses reinforcement learning to connect the core semantic meanings of the structural vocabulary with a chain-of-thought (CoT) strategy. This allows the model to explicitly decompose complex inputs into interpretable physical factors, such as composition, symmetry, and connectivity, before executing final property predictions or design decisions. This generative approach circumvents a major bottleneck in the field: the near-complete absence of ground-truth reasoning trajectories that simultaneously bridge sequence, structure, and text. 
Intra-domain structural evidence grounding first teaches domain-specific experts how structural tokens serve as evidence for reasoning. 
Cross-domain reasoning consolidation then integrates these domain-local evidence-use patterns into unified capabilities, reducing the trajectory collapse frequently observed in standard alignment techniques.

The radar plot in Fig.~\ref{fig:main_figure}E (left) compares \projName{} with four contemporary generalist LLMs (Opus-4.7~\citep{opus47}, GPT-5.5~\citep{openai2026gpt55}, DeepSeek-V4-Pro~\citep{deepseekai2026deepseekv4} and Kimi-K2.6~\citep{kimik26}) across five capability sectors and four scientific modalities (proteins, DNA/RNA, small molecules and crystals). The suite spans numerical property prediction, sequence- and structure-level classification, generation and design, modality translation and scientific QA. \projName{} expands the polygon further than every generalist baseline in every sector, with the largest margins on tasks that demand explicit structural reasoning, where text-only LLMs collapse to near-trivial scores.

We next compared \projName{} with the published specialist that defines the prior state of the art on each benchmark (Fig.~\ref{fig:main_figure}E, right). Unlike these specialists, each of which is restricted to a fixed numerical or categorical output, \projName{} produces free-form natural language alongside its predictions, yet still matches or exceeds them on the majority of tasks across the three task panels on the right: \projName{} reaches a mean $F_{\max}=0.59$ on DeepFRI-GO (averaged over the three aspects), surpassing SaProt~\citep{su2025saprothub} ($0.52$); attains $0.88$ accuracy on subcellular localization, exceeding the ESM2~\citep{esm2} ($0.84$); is on par with or above the specialist on DNA promoter and transcription-factor detection; and substantially outperforms RNA-function specialists, raising Isoform R$^2$ from $0.59$ to $0.86$ and RNA protein interaction MCC from $0.74$ to $0.81$. On the DUD-E virtual-screening suite, \projName{} matches the previously reported best AUC of $0.76$ and improves the $5.0\%$ enrichment factor from $7.12$ to $7.70$. Beyond closed-set predictions, the same model handles open-ended scientific language tasks that lie entirely outside the specialists' output space, scoring $0.85$ BertScore on biomedical QA and $0.77$~ROUGE-L on protein general function description, and supplies an interpretable natural-language rationale for each predicted label.

High benchmark scores can be obtained by exploiting homology, scaffold or template-level shortcuts, and do not by themselves demonstrate that a model has captured the underlying science. 
Sections~2.2.1--2.2.4 therefore probe four structural behaviours of \projName{} beyond average accuracy. 
On protein GO prediction (\S2.2.1), \projName{} shows its largest gain in the low-homology regime , improving Cellular Component $F_{\max}$ at $\leq30\%$ identity, by $+0.21$ over BLAST~\citep{altschul1990blast} and by $+0.13$ over ESM2, while maintaining stable reasoning quality across similarity bins, arguing against homology memorization. 
On retrosynthesis (\S2.2.2), the chain-of-thought interleaves language with SMILES sub-fragments along a canonical \emph{analysis $\rightarrow$ disconnection $\rightarrow$ verification $\rightarrow$ feasibility} workflow, recovering literature reactants in $5/5$ representative Retrosynthesis USPTO-50K cases versus $2/5$ for RSGPT~\citep{deng2025rsgpt}. On 3D molecular similarity (\S2.2.3), the per-molecule embedding clusters DUD-E~\citep{mysinger2012directory} actives by binding-pocket geometry rather than 2D scaffold. 
On inorganic crystals (\S2.2.4), the latent space separates C, Si and SiC into disjoint compositional clusters and, within each, orders polymorphs along smooth band-gap gradients. 
The accompanying reasoning traces explicitly invoke coordination changes, bonding topology and symmetry constraints as mechanistic evidence for linking crystal structure to properties. 
Consistent with this physically aligned representation, the formation-energy parity plot (Fig.~\ref{fig:mat_results}C) reaches R$^{2}=0.895$, well above competing models even where their absolute MAE is comparable, showing that \projName{} captures the stability ranking rather than the marginal label distribution.

Across scientific tasks spanning proteins, DNA, RNA, small molecules and inorganic crystals, \projName{} outperforms previous state-of-the-art methods on \textbf{\totalSOTATasks~tasks}, and matches or exceeds published domain specialists on \textbf{\totalSOTATasksExperts~benchmarks} wherever such a baseline exists. 
To further exclude reliance on sequence, scaffold or composition-level shortcuts, we ablated structural inputs in representative protein, small-molecule and materials tasks and observed a consistent drop in performance, with the largest effects in problems governed by folded geometry, binding-site organization, molecular shape or crystal topology (Fig.~\ref{fig:ab_study}). The same perturbation changed the reasoning traces: without structure, the reasoning process relied mainly on sequence motifs or chemical composition, whereas structural tokens led \projName{} to use binding pockets, coordination environments and periodic connectivity as evidence for its predictions. 
To our knowledge, this is the \textbf{first foundation model to enable sequence, structure and natural-language reasoning} across all four scientific modalities within a single autoregressive trajectory, delivering specialist-level accuracy, interpretable structure-grounded chains of thought, and open-ended scientific language generation that no prior specialist or generalist model can provide.

\subsection{Benchmarking \projName~across proteomic, molecular, and materials tasks}

\subsubsection{\projName~precisely predicts protein GO term and learns biologically meaningful structure}

\begin{figure}[!htbp]
    \centering
    \includegraphics[width=1\linewidth]{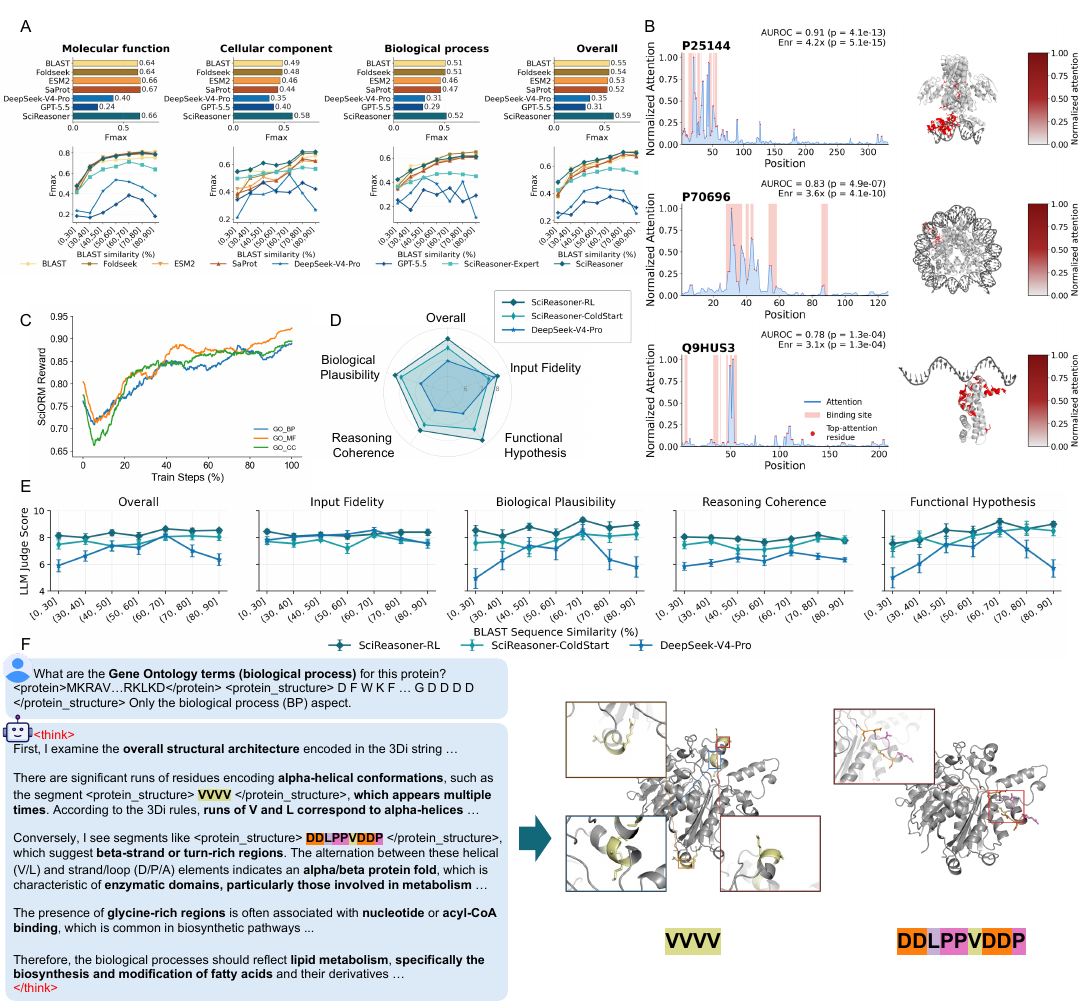}
    \caption{\textbf{\projName{} predicts protein Gene Ontology terms through structure-grounded reasoning.}
    \textbf{(A)} Performance on the CAFA-3 Gene Ontology benchmark for Molecular Function, Biological Process and Cellular Component annotations, stratified by maximum BLAST sequence identity to the training set. \projName{} shows the largest gains in low-homology regimes, particularly for Cellular Component prediction.
    \textbf{(B)} Attention analysis for DNA-binding Gene Ontology prediction. Residues receiving high attention are enriched at contact-defined DNA-binding sites and localize to protein--DNA interfaces in representative complexes.
    \textbf{(C)} Reinforcement-learning trajectories for the three ontology aspects. Rewards increase after an initial exploration phase and converge to stable plateaus, indicating consistent optimization across molecular-function, biological-process and cellular-component annotations.
    \textbf{(D)} LLM-as-judge evaluation of reasoning traces on held-out Biological Process proteins. Reinforcement learning improves input fidelity, biological plausibility, reasoning coherence and functional-hypothesis quality relative to the coldstart checkpoint and DeepSeek-V4-Pro.
    \textbf{(E)} Reasoning-quality scores stratified by BLAST similarity for Biological Process prediction. Mean GPT-5.5 judge scores are reported on five rubric axes across seven similarity bins relative to the training set ($n=20$ proteins per bin; error bars, s.e.m.). \projName{}-RL achieves the highest scores across axes and remains stable across similarity bins, suggesting that its reasoning generalizes beyond memorized homologues.
    \textbf{(F)} Structure-token visualization of the reasoning trace for \textit{E.~coli} $\beta$-ketoacyl-ACP synthase I (FabB; 2bz4\_A, UniProt P0A953). Load-bearing claims and cited 3Di-token fragments are highlighted in the trace and mapped onto PyMOL-rendered structures. The VVVV fragment maps to canonical $\alpha$-helices, whereas DDLPPVDDP maps to an extended-strand/turn region, supporting the inferred $\alpha/\beta$ fold and the resulting fatty-acid biosynthesis annotation.}
    \label{fig:go}
\end{figure}

Assigning Gene Ontology (GO) terms \citep{ashburner2000go,gene2021gene} to protein sequences is central to computational biology, translating rapidly growing sequence and structure databases \citep{abramson2024accurate} into interpretable functional annotations \citep{uniprot2023uniprot}. Since only a small fraction of proteins have experimentally verified functions \citep{friedberg2006automated,radivojac2013large}, accurate prediction across Molecular Function (MF), Biological Process (BP), and Cellular Component (CC) remains critical for protein characterization, pathway analysis, disease-gene interpretation, and therapeutic-target discovery \citep{cantalapiedra2021eggnog,gligorijevic2021deepfri}.

We evaluated GO-term prediction on DeepFRI-GO~\citep{gligorijevic2021deepfri} and, to prevent label leakage, excluded from continued pretraining all proteins with $>$30\% sequence identity to the DeepFRI-GO test set. 
As shown in Fig.~\ref{fig:go}A, \projName{} attains an overall ($F_{\max}=0.59$), outperforming sequence-based BLAST transfer (0.55) \citep{altschul1990blast}, structure-based Foldseek 3Di alignment (0.54) \citep{van2024fast}, fine-tuned protein language models including ESM2 (0.53) \citep{esm2} and SaProt 650M AF2 (0.52) \citep{su2025saprothub}, and general-purpose reasoning LLMs, DeepSeek-V4-Pro (0.35) and GPT-5.5 (0.31). The gain is strongest for CC, where \projName{} reaches 0.58 versus BLAST (0.49), and remains positive for BP (0.52 versus 0.51). MF is near saturation, with \projName{} at 0.66, comparable to SaProt (0.67). 
 
Overall, \projName{} is the only method that leads or matches the best baseline across all branches, indicating that it integrates evidence beyond any single baseline family.

To determine whether these gains reflect genuine functional inference rather than memorized homology-based transfer, we stratified the test proteins by maximum BLAST percent identity to the training set (\(e\leq10^{-3}\), query coverage \(\geq50\%\), self-hits removed). The advantage of \projName{} is largest precisely where sequence similarity becomes unreliable. 
In the CC \((0,30]\%\) identity bin, \projName{} improves over BLAST by \(+0.21\) absolute \(F_{\max}\) (0.55 versus 0.34 in Fig.~\ref{fig:go}A), and outperforms ESM2 by \(+0.13\) absolute \(F_{\max}\) (0.55 versus 0.42). 
This low-homology advantage aligns with the design of \projName{}: whereas BLAST relies on local sequence similarity, ESM2 encodes evolutionary and sequence-context patterns without explicit structural grounding, \projName{} predicts GO terms through reasoning traces that integrate domain composition, localized motifs, structural environments, and reference proteins. Such local functional cues can remain conserved even when global sequence identity is weak.

We next tested whether \projName{} grounds GO predictions in functionally relevant protein regions rather than superficial sequence patterns. Using DNA-binding proteins, where protein--DNA contacts provide structural ground truth, we extracted attention from the generated DNA-binding GO token (GO:0003677) to residue-level structure and sequence tracks and combined them into a per-residue profile. These profiles strongly align with DNA-binding sites (Fig.~\ref{fig:go}B): catabolite control protein A (UniProt: P25144) reaches AUROC (=0.91) and top-20\% fold enrichment ($=4.2\times$), histone H2B type 1-A (UniProt: P70696) reaches AUROC (=0.83) and enrichment ($=3.6\times$), and DesT (UniProt: Q9HUS3) reaches AUROC (=0.78) and enrichment ($=3.1\times$).
In all cases, the most attended residues fall within or immediately adjacent to annotated binding regions, and structural projections confirm that these residues cluster near bound DNA in three-dimensional space. Thus, \projName{} does not merely assign the correct GO term. It concentrates its evidence on residues that are physically responsible for the corresponding molecular function.

Reinforcement learning further improves GO prediction across all three ontology aspects (Fig.~\ref{fig:go}C). After a brief exploration phase during the first \(\sim\!5\%\) of training, in which the science reward temporarily decreases under noisy advantages, all three trajectories rise and converge to stable plateaus. MF reaches the highest reward (\(\sim\!0.92\)), while BP and CC converge at comparable values (\(\sim\!0.89\)).
The simultaneous improvement across all aspects indicates that RL does not merely exploit aspect-specific shortcuts, but instead strengthens a shared protein-function representation that supports functional, process, and localization annotations.

Because $F_{\max}$ captures only final GO-term overlap, we also evaluated the reasoning traces underlying those predictions. To assess grounded biological inference beyond surface fluency, we used GPT-5.5 to score held-out BP traces on Overall Quality, Input Fidelity, Biological Plausibility, Reasoning Coherence, and Functional Hypothesis Quality. \projName{} scores highest on every axis, with an Overall score of 8.33 versus 7.77 for SciReasoner-ColdStart and 6.96 for DeepSeek-V4-Pro \citep{deepseekai2026deepseekv4} (Fig.~\ref{fig:go}D). 
Relative to the pre-RL coldstart model, the largest improvement is in Biological Plausibility (8.68 vs.\ 7.83), indicating that RL sharpens not only the fluency of the trace but also the biological relevance of the inferred function. 
The gap to DeepSeek-V4-Pro is concentrated on the biology-specific axes---Biological Plausibility (8.68 vs.\ 6.63) and Functional Hypothesis Quality (8.45 vs.\ 6.77)---and is far wider there than on Input Fidelity (8.27 vs.\ 8.04). Although DeepSeek-V4-Pro is a substantially larger general-purpose model, this pattern shows that domain-tuned scientific reasoning, rather than raw model scale alone, drives the improvement.

We stratified reasoning quality by BLAST similarity to test whether \projName{} relies mainly on close homologues (Fig.~\ref{fig:go}E). Across seven similarity bins, \projName{} maintains higher judge scores than both the coldstart checkpoint and DeepSeek-V4-Pro, including in low-similarity regimes where direct homology transfer is least reliable. This helps explain the low-homology gains in Fig.~\ref{fig:go}A. 
Rather than copying annotations from close neighbours, \projName{} can combine weaker but conserved cues, such as local structural motifs, short sequence fragments, fold-level patterns, and biochemical context. These results suggest that its improved performance on low-similarity proteins arises from structure-grounded functional reasoning rather than memorization of homologous sequences.

Finally, Fig.~\ref{fig:go}F illustrates how the model's reasoning can be audited at the level of structure tokens. For \textit{E. coli} \(\beta\)-ketoacyl-ACP synthase I (UniProt: P0A953), \projName{} identifies repeated \texttt{VVVV} fragments as canonical \(\alpha\)-helical regions and the \texttt{DDLPPVDDP} fragment as an extended strand/turn-rich segment. 
The trace then links this fold architecture, together with glycine-rich and catalytic-context cues, to fatty-acid biosynthesis and related lipid-metabolic processes. This example demonstrates that the reasoning trace is not an isolated natural-language explanation: its claims can be connected back to explicit structural substrings and spatial regions of the protein.

Together, these results show that \projName{} advances GO-term prediction in three ways: it improves accuracy most in low-homology regimes, grounds predictions in localized sequence and structural evidence, and uses reinforcement learning to enhance both annotations and reasoning quality. Thus, \projName{} serves not only as a more accurate GO predictor, but also as a mechanism-aware annotation framework that explains molecular function, biological process, and cellular localization assignments.

\subsubsection{\projName~formulates chemically viable retrosynthetic routes with high precision
}

\begin{figure}[!htbp]
    \centering
    \includegraphics[width=1\linewidth]{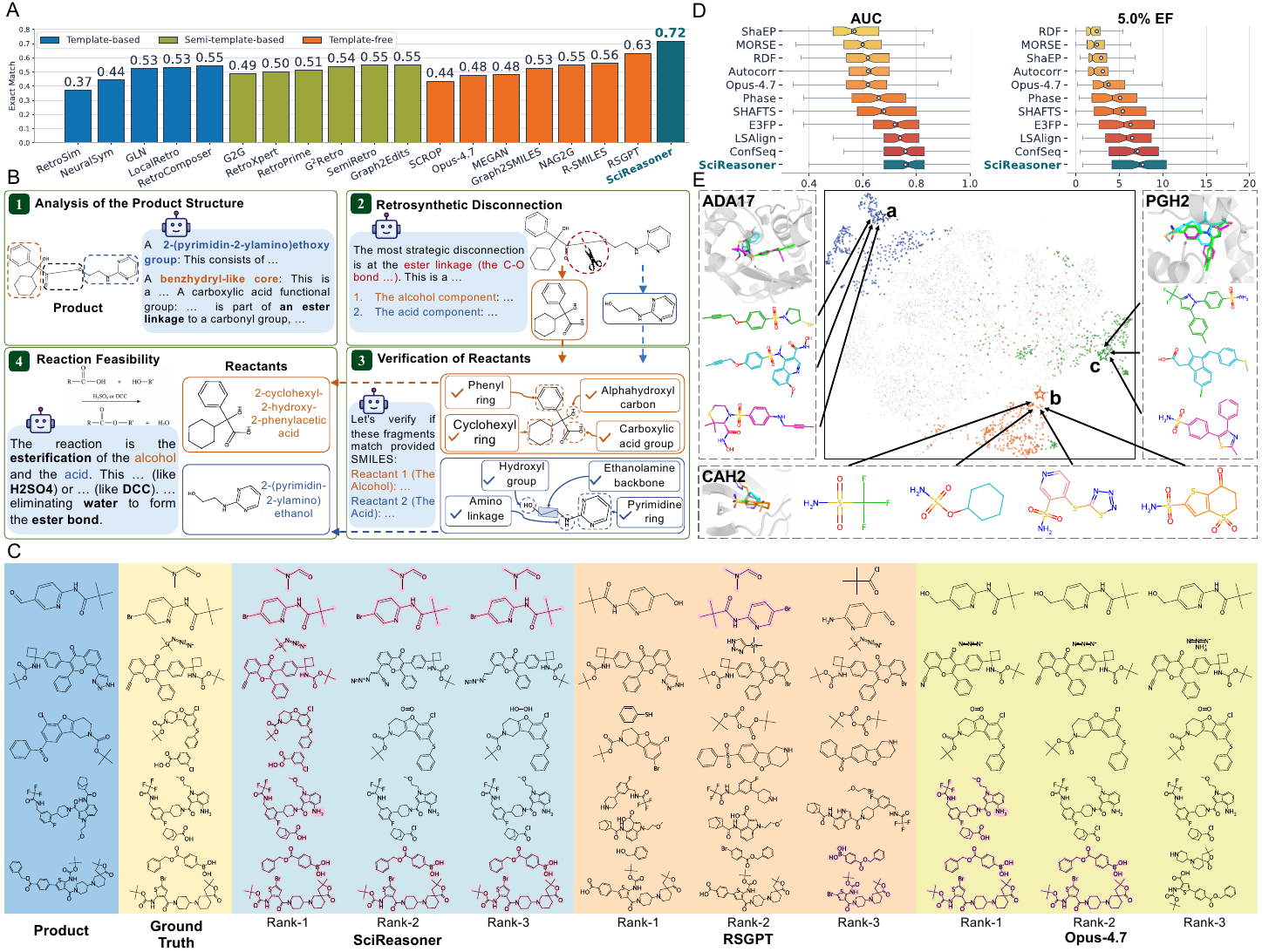}
    \caption{ 
        \textbf{\projName~formulates chemically viable retrosynthetic routes (A--C) and captures plausible 3D molecular structure for similarity representation (D, E).}
        \textbf{(A)} {{Retrosynthesis USPTO-50K performance.}}
        Across template-based, semi-template-based and template-free methods, \projName~reaches $0.72$, exceeding the prior best RSGPT by $+0.09$ points, while Opus-4.7 five-shot scores only $0.48$.
        \textbf{(B)} {{Chain-of-thought trace for a representative product.}}
        \projName~proceeds in four interpretable stages---{Analysis}, {Disconnection}, {Verification} and {Feasibility}, and recovers the two ground-truth reactants by severing the strategic ester C\textendash O bond. Every intermediate claim is itself a SMILES fragment, making the trace auditable atom-by-atom.
        \textbf{(C)} {{Cross-model Top-3 reactant predictions on Retrosynthesis USPTO-50K.}}
        Five products with their ground-truth reactants and the Top-3 predictions of \projName, RSGPT and Opus-4.7. Molecules in \textcolor{pink}{pink} match the gold canonical SMILES. \projName~recovers a correct candidate for $5/5$ products versus $2/5$ for RSGPT and $2/5$ for Opus-4.7, generalising across mechanistically distinct reaction families, rather than fitting a single template.
        \textbf{(D)} {{Per-target distributions on 3D molecule structure (DUD-E).}}
        AUC and $5.0\%$ EF box plots over $102$ DUD-E targets for nine published baselines, an Opus-4.7 reference, and \projName. \projName{} matches the previously reported best AUC and establishes a new best $5.0\%$ EF.
        \textbf{(E)} {{3D ligand-target geometry for three DUD-E targets.}}
        UMAP of the per-molecule \projName~embedding, coloured points are the $400$ closest actives for each target, labelled by DUD-E target identifiers—ADA17 (blue), CAH2 (orange), and PGH2 (green); open stars mark the reference queries. Side panels show each query and three representative actives docked into the receptor, then redrawn as 2D structures: chemically distinct scaffolds that nevertheless converge to the same 3D pharmacophore. 
    }
    \label{fig:retro_joint}
\end{figure}

Retrosynthesis (the recursive disconnection of a target molecule into commercially available precursors) is the cognitive backbone of organic-synthesis route planning. Reliable retrosynthetic models accelerate hit-to-lead optimization, and an interpretable reasoning trace lets human chemists verify and reuse the suggested routes.

We evaluate on the standard Retrosynthesis USPTO-50K \citep{schneider2016big} split: the model is given a single target SMILES and must output the set of reactant SMILES that produces it. We sample 16 stochastic completions per query ($T{=}0.6$, top-$p{=}0.95$) and rank the answers by sample frequency. A prediction is correct only when its canonical SMILES exactly matches the ground-truth reactant set, and we report Exact Match as the standard metric. Fig.~\ref{fig:retro_joint}A places \projName~against $18$ published baselines spanning template-based \citep{coley2017computer,segler2017neural,dai2019retrosynthesis,chen2021deep,yan2022retrocomposer}, semi-template-based \citep{shi2020graph,yan2020retroxpert,wang2021retroprime,chen2023g,gao2022semiretro,zhong2023retrosynthesis} and template-free \citep{zheng2019predicting,sacha2021molecule,tu2022permutation,zhong2022root,yao2024node,deng2025rsgpt,opus47} families: \projName~reaches \textbf{$0.72$} (Exact Match), exceeding the previous best template-free method RSGPT \citep{deng2025rsgpt} ($0.63$) by $+0.09$ points, while Opus-4.7 five-shot scores only $0.48$. 
To prevent direct test leakage, we removed from the pretraining corpus all reactions whose product SMILES matched any product in the Retrosynthesis USPTO-50K test set. 

Unlike generic chat models that emit a free-form natural-language explanation followed by a single SMILES guess, \projName~is a scientific reasoning model that interleaves natural-language steps with \emph{structural tokens (SMILES fragments)} inside its chain-of-thought, e.g.\ the ester C--O bond, the phenyl ring \texttt{c1ccccc1}. Because the intermediate steps are themselves chemical structures, the trace can be checked atom-by-atom and the final reactant assembly is grounded in concrete sub-fragments.

Fig.~\ref{fig:retro_joint}B illustrates this on a hydroxy-substituted benzhydryl ester. \projName~severs the strategic ester C--O bond, independently verifies the expected functional groups on each precursor, and recognises the inverse reaction as an acid-catalysed esterification. Six of the seven sub-fragments emitted in the trace are strictly smaller than either reactant, yet their union exactly reconstructs the gold answer, indicating bottom-up reasoning over chemical groups rather than template recall.

Fig.~\ref{fig:retro_joint}C extends this to five further Retrosynthesis USPTO-50K products, contrasting \projName~against the specialised retrosynthesis model RSGPT and the general-purpose large language model Opus-4.7. \projName~contains the gold reactant set in its Top-3 for $5/5$ rows; RSGPT recovers it in $2/5$, and Opus-4.7 in $2/5$.

On the $N$-tert-butyl pyridine-carboxamide aldehyde, \projName~recovers the literature Vilsmeier formylation route (DMF/Br precursor pair) with a single canonical answer across all 16 samples; RSGPT proposes a hydroxymethyl + bromopyridine pair, while Opus-4.7 commits all three ranks to oxidation of a benzyl alcohol and never proposes the C--N disconnection (Fig.~\ref{fig:retro_joint}C, Row 1). On a 1,4-disubstituted 1,2,3-triazole-fused phenyl-benzofuranone, the gold cut is a copper-catalysed azide--alkyne [3+2] cycloaddition (CuAAC) of trimethylsilyl azide with the terminal aryl-alkyne; only \projName's rank-1 recovers this pair, while RSGPT keeps the triazole intact and disconnects elsewhere and Opus-4.7 mis-treats the cycloaddition as an azide-plus-nitrile coupling (Fig.~\ref{fig:retro_joint}C, Row 2). On a chloro-arylsulfanyl tetrahydrofuro-pyridine $N$-Boc carbamate, the strategic cut is $m$-CPBA oxidation of the thioether to the sulfoxide; only \projName's rank-1 identifies it, while RSGPT and Opus-4.7 instead attempt to cleave the Boc carbamate or the tetrahydrofuran (Fig.~\ref{fig:retro_joint}C, Row 3). On a polycyclic indole-based amide, \projName~and Opus-4.7's rank-1 both propose the correct amide coupling between the bicycloalkyl carboxylic acid and the parent aniline; the remaining baselines fragment the indole or the trifluoromethyl chain (Fig.~\ref{fig:retro_joint}C, Row 4). On a thiophene--phenyl biaryl bearing a Boc-aminothiophene and a spirolactone-piperazine amide, the strategic cut is a Suzuki--Miyaura coupling of the 5-bromo Boc-aminothiophene with benzyl 4-boronate-benzoate; \projName~returns this pair across all $16$ samples and Opus-4.7's rank-1/rank-2 both match, while RSGPT recovers the gold pair only at rank-3 (Fig.~\ref{fig:retro_joint}C, Row 5).

Taken together, the structural-token chain-of-thought makes \projName's predictions both more accurate and easier to audit: the trace states inside itself why each bond is the strategic cut, rather than answering in one step. 

\subsubsection{\projName~captures plausible 3D molecular structure for similarity representation}

Identifying compounds that share a binding-pocket geometry but not a 2D scaffold is the central problem of virtual screening and ligand-based drug discovery. A representation that groups molecules by binding-pocket compatibility, rather than by simple substructure overlap, would directly enable hit expansion and scaffold hopping.

We probe whether \projName~itself already provides such a representation. From a single-molecule prompt, we extract the last layer hidden state of \projName~and average it over the 10 tokens generated immediately after the prompt to obtain a molecular embedding. Following the standard DUD-E protocol \citep{mysinger2012directory}, we ask how well Euclidean distance in this embedding space ranks actives versus decoys for a given target query, reporting AUC and the $5.0\%$ EF, and the embedding's qualitative structure is visualised via 2D UMAP projection.

\projName~natively consumes \emph{structural tokens}: a dedicated ConfSeq \citep{xiong2025bridging} tokenizer converts each 3D conformer into a sequence of geometry-aware tokens drawn from a separate vocabulary backed by a separate learned embedding table, held disjoint from the natural-language embedding. Because structural tokens are a parallel modality, geometric features are separated from those used for IUPAC names or descriptive language, so the hidden states above structural-token positions carry an uncontaminated geometric signal.

Fig.~\ref{fig:retro_joint}D quantifies this representation against nine published virtual-screening methods on $102$ DUD-E targets, spanning alignment-free descriptors and alignment-based shape-overlay methods \citep{hemmer1999deriving,devinyak20143d,klein2004topological,axen2017simple,xiong2025bridging,vainio2009shaep,sastry2011rapid,liu2011shafts,hu2018ls}, plus an Opus-4.7~\citep{opus47} reference. Used as a drop-in similarity feature, \projName's per-molecule embedding matches ConfSeq's prior best AUC of $0.76$ and improves the best $5.0\%$ EF from $7.12$ to $7.70$, obtained without docking, scoring functions, or task-specific fine-tuning.

Fig.~\ref{fig:retro_joint}E demonstrates the qualitative structure of the same embedding on three pharmacologically distinct DUD-E targets spanning unrelated protein families: ADAM17 (a zinc metalloprotease), carbonic anhydrase II (a zinc hydrolase) and cyclooxygenase-2 / PGHS-2 (a heme-containing dioxygenase). Each target's top-$400$ closest actives form a clearly separated cluster around its query ligand ($\bigstar$) on the 2D UMAP, and the three clusters occupy distinct regions of the plane despite the proteins being sequence-unrelated, and the decoys (light grey) do not infiltrate any cluster, indicating that the embedding captures family-specific chemistry rather than gross descriptors.

The boxed side panels visualise the per-target binding mode in 3D. For ADAM17, the cyclic-thiol sulfonamide query is matched by closest actives that swap the thiopyrrolidine zinc-binder for a hydroxamate amide or extend the alkyne tail into a propargyl amine while keeping the diaryl sulfonyl core; their Morgan-fingerprint Tanimoto similarities to the query are only $0.30$--$0.32$, yet USRCAT 3D-shape similarities remain greater than $0.22$ and Open3DALIGN scores reach $120$--$140$, meaning all three ligands drape over the same TACE active site. For PGHS-2, the celecoxib-like diaryl-pyrazole sulfonamide query is matched by an indanone-acetic-acid analog (Tanimoto $0.11$) and a chloro-aryl-thiazole-sulfonamide (Tanimoto $0.31$), chemically divergent COX-2 chemotypes that adopt the same 3D pharmacophore. Carbonic anhydrase II is the most striking case: the query reduces to the minimal \ce{NH2{-}SO2{-}CF3} warhead, and its three closest actives---a cyclohexyl sulfamate, a pyridyl-thiadiazole sulfonamide and a thiopheno-thiazinone, spanning very different fused-ring topologies yet all expose the canonical zinc-binding sulfonamide that defines CA-II inhibition.

Taken together, both qualitative and quantitative results indicate that \projName~has implicitly learned to cluster DUD-E actives by binding-pocket geometry rather than scaffold similarity.

\begin{figure}[!htbp]
    \centering
    \includegraphics[width=0.9\linewidth]{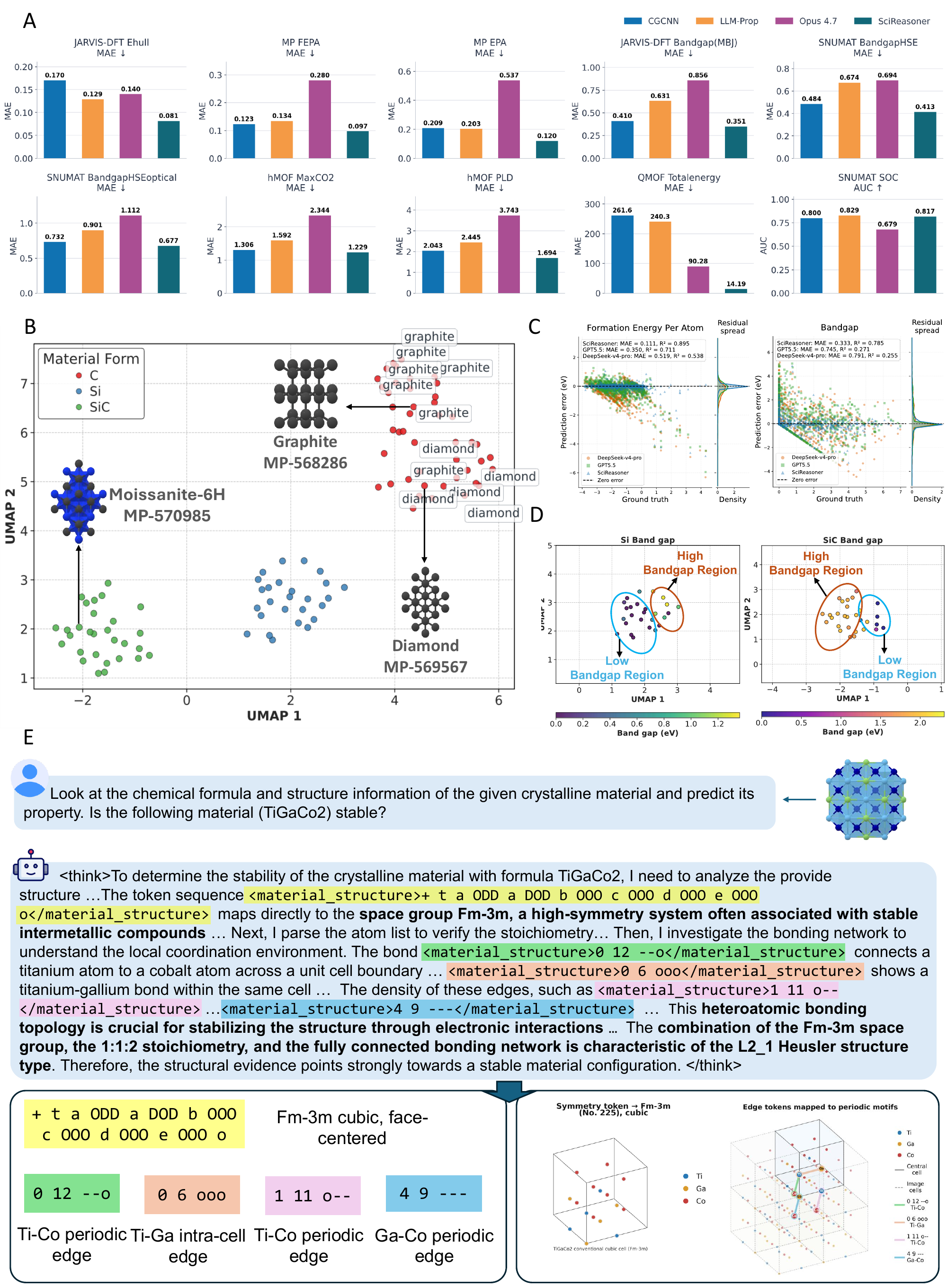}
    \caption{\textbf{Evaluation of material predictive performance and visualization of the learned structure-property latent space}.
\textbf{(A)} Performance comparison across 10 material property prediction sub-tasks, both regression  (MAE as the metric) and classifications (AUC as the metric) tasks are reported. 
\textbf{(B)} UMAP of the high-dimensional representations extracted by the model on three materials (C, Si and SiC). 
The latent space inherently clusters materials by chemical space and spontaneously distinguishes distinct structural polymorphs. 
\textbf{(C)} Parity plots comparing model-predicted formation energies per atom and bandgap against ground-truth values. The strong correlations (R$^2$
 = 0.895 and 0.785) demonstrate the model's ability to capture physically meaningful materials-property information. 
 \textbf{(D)} Local UMAP embeddings for the Si and SiC subsets, color-coded by their corresponding band gaps. The continuous transition and distinct grouping of high- and low-bandgap regions indicate that the learned representations effectively capture underlying structure-property relationships.
  \textbf{(E)} Structure-grounded reasoning trace for a representative stability prediction on TiGaCo$_2$. The model's reasoning explicitly identifies the SLICES symmetry substring, which decodes to the 
  Fm$\bar{3}$m space group, and cites edge substrings.
  These substrings are mapped back to the corresponding CIF structure, where they visualize the global cubic symmetry framework and local periodic connectivities.
 }
    \label{fig:mat_results}
\end{figure}

\subsubsection{\projName~learns structure- and property-aware representations for materials tasks}

We evaluated \projName~on ten downstream materials sub-tasks from five complementary databases: 
Materials Project\footnote{\url{https://materialsproject.org/}},
JARVIS-DFT\footnote{\url{https://jarvis.nist.gov/jarvisdft}},
SNUMAT\footnote{\url{https://www.snumat.com/}},
hMOF\footnote{\url{https://mof.tech.northwestern.edu/}},
and QMOF\footnote{\url{https://contribs.materialsproject.org/projects/qmof}}. 
These sub-tasks cover inorganic crystals, semiconductors and metal organic frameworks, and include thermodynamic stability, formation and total energies, electronic band gaps, spin orbit coupling, CO$_2$ uptake and pore geometry. For these sub-tasks, \projName~was compared with CGCNN~\citep{xie2018crystal}, LLM-Prop~\citep{niyongabo2025llm} and Opus-4.7, using Mean Absolute Error (MAE) for regression tasks and Area Under the Curve (AUC) for classification tasks. 
As shown in Fig.~\ref{fig:mat_results}A, \projName~outperforms CGCNN across all ten tasks and exceeds LLM-Prop on most numerical properties. These results indicate that its representations capture both general energetic trends and more specialized relationships involving electronic structure, relativistic effects, adsorption behaviour and pore geometry.

We next examined the latent space learned by \projName. UMAP projections show clear separation among C, Si and SiC structures, indicating that the embeddings encode global chemical identity (Fig.~\ref{fig:mat_results}B). Within each compositional domain, structures do not collapse into a single cluster. Instead, they form extended subspaces that distinguish polymorphs and polytypes with the same composition, suggesting sensitivity to local geometry and bonding topology. 
At the quantitative level, \projName~produces DFT-consistent predictions for both formation energy and band gap, with parity plots showing close agreement with ground-truth values and residuals concentrated around zero (Fig.~\ref{fig:mat_results}C). This predictive accuracy is reflected in the learned representation space: local UMAP projections for Si and SiC separate low- and high-band-gap structures while preserving continuous property gradients within each compositional manifold (Fig.~\ref{fig:mat_results}D). Together, these results show that \projName~organizes structurally related materials along physically meaningful electronic-property manifolds, rather than encoding composition alone. 

A representative TiGaCo$_2$ stability prediction illustrates how \projName~grounds its reasoning in explicit structural tokens rather than relying only on the chemical formula (Fig.~\ref{fig:mat_results}E). The model identifies the structure substring corresponding to the Fm$\bar{3}$m space group and links local edge substrings to periodic Ti--Co, Ti--Ga and Ga--Co connectivities in the reconstructed CIF structure, providing a traceable connection between token-level evidence and crystallographic features.

\subsection{Structural evidence drives native scientific reasoning}

\begin{figure}[!htbp]
    \centering
    \includegraphics[width=1\linewidth]{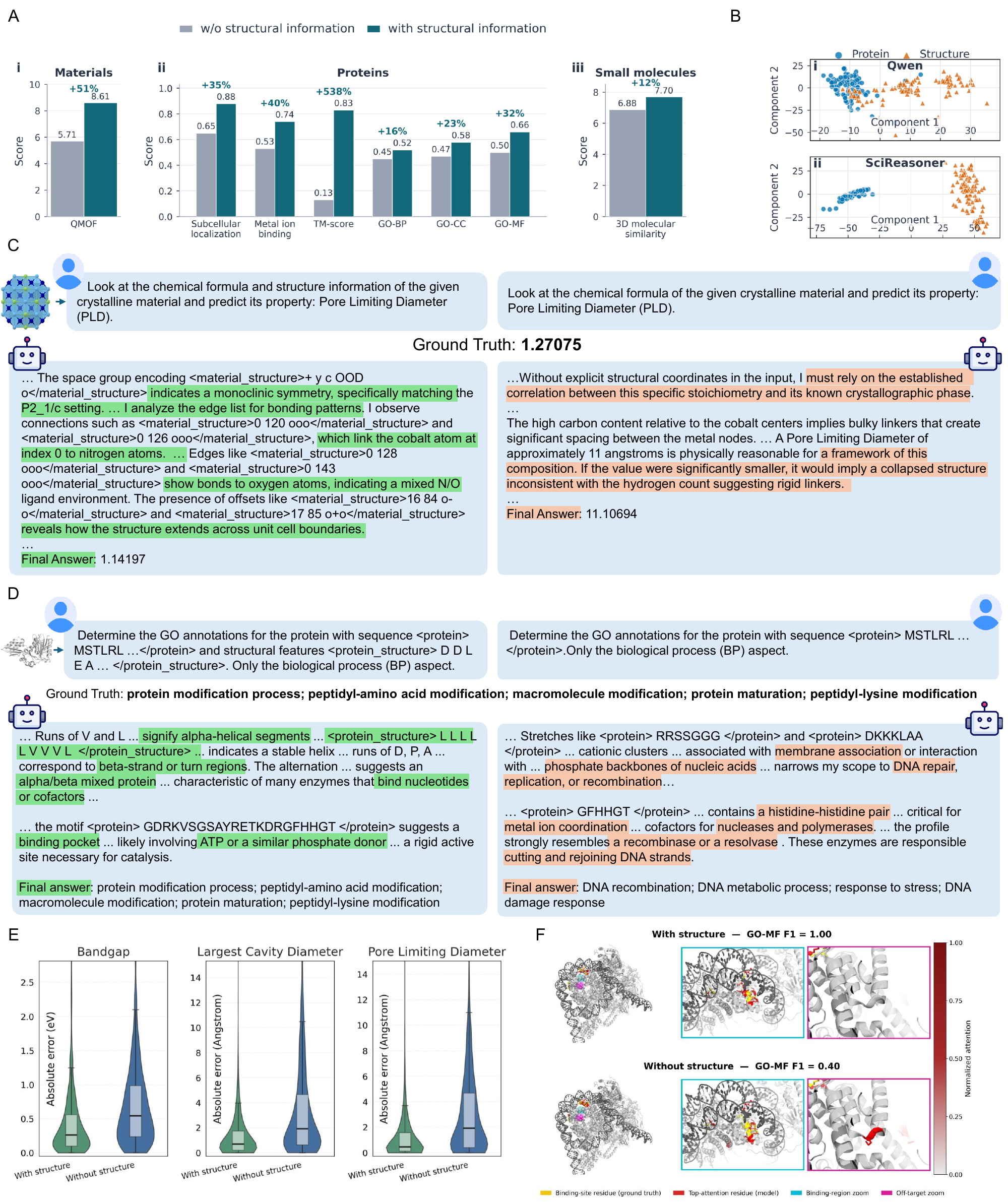}
    \caption{\textbf{Structural information empowers native scientific reasoning.}
    \textbf{(A) Structure-ablation benchmark.} Performance with and without structural inputs across materials, proteins and small molecules. Structural cues consistently improve prediction beyond sequence- or text-only inputs.
    \textbf{(B) Modality-separated representations.} PCA of Qwen and    \projName{} embeddings for paired protein sequence and structure inputs. \projName{} more clearly separates the two modalities, indicating stronger structure-aware encoding.
    \textbf{(C) Structure-grounded materials reasoning.} Chain-of-thought comparison for QMOF pore-limiting-diameter prediction. Without structure, the model relies on compositional priors and overestimates pore size; with structural tokens, it uses symmetry and bonding connectivity to recover a value closer to the ground truth.
    \textbf{(D) Structure-grounded GO reasoning.} Chain-of-thought comparison for biological-process GO prediction. Sequence-only reasoning predicts DNA-related processes, whereas structure-aware reasoning integrates secondary-structure and binding-site evidence to recover protein-modification annotations.
    \textbf{(E) Materials prediction with structure.} Absolute-error distributions for bandgap, largest-cavity-diameter and pore-limiting-diameter prediction. Structural input reduces errors across all three properties.
    \textbf{(F) Structure-guided GO attention.} Residue-level attention maps for GO molecular-function prediction. Structure-aware reasoning focuses on the functional binding site, whereas sequence-only reasoning attends to non-binding regions, consistent with its lower accuracy.}
    \label{fig:ab_study}
\end{figure}

To test whether native scientific reasoning depends on explicit structural evidence, we performed structure-ablation experiments across materials, proteins and small molecules (Fig.~\ref{fig:ab_study}A).
Removing structural information consistently weakened performance, whereas providing structural inputs improved prediction across all three domains.
The gains were especially pronounced for protein tasks, where function depends on folded geometry, binding-site organization and spatial residue interactions, but were also evident in materials and molecular-similarity tasks, where periodic topology and three-dimensional shape provide information that cannot be recovered from composition, sequence or two-dimensional descriptors alone.
Thus, the ablation shows that structural evidence is not merely an additional input channel, but a necessary substrate for the model's native scientific reasoning.

The embedding analysis further shows that \projName{} learns a distinct representational space for structural evidence (Fig.~\ref{fig:ab_study}B).
In the Qwen embedding space, paired protein sequence and structure inputs remain partially intermixed, suggesting that structural descriptions are treated largely as another form of text.
By contrast, \projName{} forms more clearly separated sequence and structure clusters, indicating that structural tokens occupy a distinct representational subspace while preserving within-modality consistency.
This modality separation suggests that the model can distinguish sequence-level evidence from structure-level evidence, a prerequisite for reasoning tasks that require explicit modelling of sequence--structure relationships.

Case studies show that structural evidence changes the reasoning path, not merely the final output.
For QMOF pore-limiting-diameter prediction, the structure-free model relies on compositional and stoichiometric priors and overestimates the pore size by nearly an order of magnitude (Fig.~\ref{fig:ab_study}C).
With structural tokens, \projName{} instead cites monoclinic symmetry, cobalt coordination to nitrogen and oxygen atoms, edge connectivity and periodic offsets, producing a prediction close to the ground truth.
Similarly, for biological-process GO prediction, sequence-only reasoning is diverted by cationic and histidine-rich motifs towards DNA recombination and stress-response terms, whereas structure-aware reasoning integrates secondary-structure context and binding-pocket evidence to recover protein-modification annotations (Fig.~\ref{fig:ab_study}D).
These examples indicate that structural evidence redirects the model from superficial sequence or composition cues towards physically and biologically meaningful explanations.

Quantitative analyses support the same conclusion.
For materials prediction, structural input reduces absolute errors for bandgap, largest-cavity-diameter and pore-limiting-diameter prediction (Fig.~\ref{fig:ab_study}E), showing that crystal geometry improves both electronic-structure and pore-geometry inference.
For GO molecular-function prediction, residue-level attention maps show that structure-aware reasoning concentrates high attention around the functional binding site, whereas sequence-only reasoning attends to non-binding regions and yields lower accuracy (Fig.~\ref{fig:ab_study}F).
Together, the ablations, representation analysis, reasoning cases and attention maps support the claim that structural evidence drives native scientific reasoning in \projName{}, enabling the model to ground its conclusions in biologically and physically meaningful structural features rather than superficial sequence, composition or textual correlations.

\subsection{Intra-domain grounding and cross-domain consolidation stimulate self-bootstrapped native structural reasoning}

\begin{figure}[!htbp]
    \centering
    \includegraphics[width=1\linewidth]{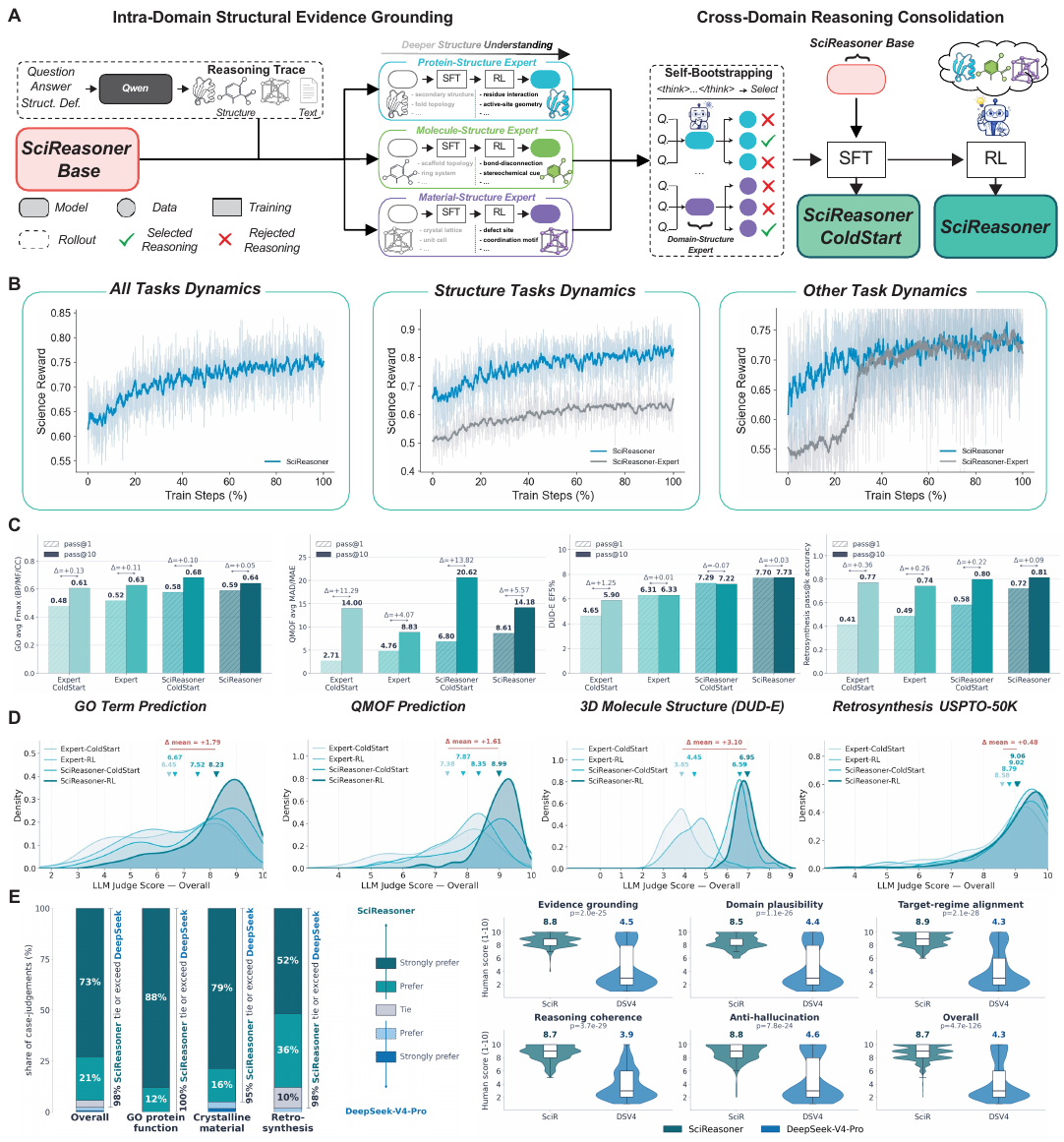}
    \caption{\textbf{Self-bootstrapped native structural reasoning unifies scientific reasoning.}
    \textbf{(A) Intra-domain structural evidence grounding and cross-domain reasoning consolidation.} Sketch reasoning traces coldstart task-specific experts, which are optimized independently by reinforcement learning (RL) to ground structural evidence within each domain. Expert-generated traces are then pooled to coldstart and reinforce a unified SciReasoner across all tasks to consolidate cross-domain reasoning patterns.
    \textbf{(B) Reward dynamics.} RL rewards rise during training, with the unified SciReasoner matching or exceeding expert variants on structure-intensive and other scientific tasks.
    \textbf{(C) Improved accuracy and sampling efficiency.} Pass@1 and pass@10 are compared across four training stages for GO term prediction, QMOF prediction, 3D molecule structure (DUD-E) and Retrosynthesis USPTO-50K. Lighter hatched bars indicate pass@1 and solid bars pass@10; $\Delta=\mathrm{pass@10}-\mathrm{pass@1}$. The post-training procedure consistently increases pass@1 and reduces $\Delta$, indicating more reliable first-sample predictions.
    \textbf{(D) Improved reasoning quality across tasks.} Kernel density estimates show LLM-judged reasoning scores for four model variants on GO term prediction, QMOF prediction, 3D molecular similarity (DUD-E), and Retrosynthesis USPTO-50K. Triangles denote means and red brackets indicate the worst-to-best mean difference. Self-bootstrapped native structural reasoning shifts score distributions towards higher quality, and unified SciReasoner generally outperforms domain-structure experts.
    \textbf{(E) Double-blinded human-expert evaluation.} Domain experts compared SciReasoner and DeepSeek-V4-Pro on GO annotation, materials-property prediction and retrosynthesis. Left, pairwise preferences overall and by task; right, score distributions for individual scoring-axis. SciReasoner was rated tie-or-better in 98\% of comparisons and scored higher on every axis.
}
    \label{fig:rl}
\end{figure}

While broad scientific pre-training provides diverse molecular, biomolecular, and materials representations, it does not establish a unified reasoning protocol across scientific tasks. We therefore adopt a self-bootstrapped post-training framework, which first grounds structural evidence within each scientific domain (i.e., intra-domain structural evidence grounding) and then integrates these domain-local reasoning patterns into a single scientific reasoner (i.e., cross-domain reasoning consolidation) (Fig.~\ref{fig:rl}A). Sketch traces provide only a coldstart prior, after which domain-structure experts generate higher-quality on-policy traces that are pooled for final training. 
Together, this two-stage procedure turns pre-trained structural priors into explicit scientific reasoning, first eliciting domain-native evidence use and then consolidating it into a unified structure-grounded reasoner.

The reward and performance dynamics in Fig.\ref{fig:rl}B support the self-bootstrapped strategy. Intra-domain structural evidence grounding  yields consistent reward gains across task families, indicating that a small set of sketch traces can seed trainable reasoning behaviours. After expert-generated traces are pooled, cross-domain reasoning consolidation enables the unified \projName{} to outperform the coldstart baseline across both structure-intensive and non-structure tasks. 
This suggests that expert-generated traces better align with the model's representation space, enabling model-native reasoning trajectories to emerge.

Fig.~\ref{fig:rl}C shows that the post-training procedure improves both first-sample accuracy and sampling efficiency across tasks. For pass@10, we used task-specific aggregation: the mean Fmax@10 across BP, MF and CC for GO annotation, an oracle best-of-10 prediction selected by minimum absolute error for QMOF regression, a label-independent 10-sample score ensemble followed by 5.0\% EF evaluation for 3D molecule structure (DUD-E), and top-10 reactant-set accuracy for retrosynthesis. In GO term prediction, intra-domain structural evidence grounding raises pass@1 from 0.48 to 0.52 and reduces the pass@10--pass@1 gap from 0.13 to 0.11, while cross-domain reasoning consolidation further achieves the highest pass@1 of 0.59 with a smaller gap than its coldstart precursor. In QMOF prediction, pass@1 increases from 2.71 to 4.76 after intra-domain structural evidence grounding and from 6.80 to 8.61 after cross-domain reasoning consolidation, with the gap reduced from 11.29 to 4.07 and from 13.82 to 5.57, respectively. 3D molecule structure (DUD-E) prediction shows the same trend: pass@1 improves from 4.65 to 6.31 after intra-domain structural evidence grounding and from 7.29 to 7.70 after cross-domain reasoning consolidation. Retrosynthesis shows the largest absolute gain, with pass@1 rising from 0.41 to 0.49 after intra-domain structural evidence grounding and to 0.72 after cross-domain reasoning consolidation, accompanied by gap reductions from 0.36 to 0.26 and from 0.22 to 0.09. 
These consistent gap reductions indicate that the procedure does not merely improve best-of-$N$ performance, but concentrates probability mass on correct first responses, making SciReasoner more reliable under single-sample scientific prediction.

Fig.~\ref{fig:rl}D further evaluates reasoning quality beyond final-answer accuracy using LLM-judge score distributions over 100 records per task. Self-bootstrapped native structural reasoning consistently shifts the distributions toward higher scores, with the largest worst-to-best mean improvement on 3D molecule structure (DUD-E) ($\Delta_{\mathrm{mean}}=+3.10$), followed by GO protein-function annotation ($+1.79$), QMOF property prediction ($+1.61$), and Retrosynthesis USPTO-50K ($+0.48$). 
The larger gains on DUD-E, GO and QMOF suggest that the self-bootstrapped procedure is especially effective when reasoning must be grounded in molecular topology, protein structural evidence or crystal-property relationships. By contrast, retrosynthesis starts from an already high reasoning-quality distribution, leaving less room for improvement. Its smaller but positive shift therefore indicates slower saturation rather than failure of the post-training procedure. Overall, these distribution-level gains show that the procedure improves not only answer correctness, but also the coherence and evidence grounding of the scientific reasoning process.

Fig.~\ref{fig:rl} shows that self-bootstrapped native structural reasoning progressively transforms domain-structure experts into a unified scientific reasoner. Intra-domain structural evidence grounding strengthens domain-specific reasoning, while cross-domain reasoning consolidation transfers these capabilities into a single cross-domain model. Improvements in reward, pass@1 performance, and reasoning quality demonstrate more accurate and stable evidence-grounded reasoning.

\subsection{Human Expert Evaluation}

The automated metrics and LLM judges establish that SciReasoner produces accurate answers, but they cannot establish whether its \emph{reasoning traces} are genuinely useful to domain experts. We therefore conducted a double-blinded human evaluation pilot in which domain experts compared SciReasoner and DeepSeek-V4-Pro reasoning traces on cases spanning three tasks --- GO protein-function annotation, crystalline-material property prediction, and single-step retrosynthesis on Retrosynthesis USPTO-50K. Evaluators were double-blinded to model identity and rated every trace against a five-axis, 1--10 rubric (Q1 evidence grounding, Q2 domain plausibility, Q3 target-regime alignment, Q4 reasoning coherence, Q5 anti-hallucination) together with a five-point pairwise preference (Q8, strongly prefer SciReasoner $\to$ strongly prefer DeepSeek-V4-Pro). The two panels of Fig.~\ref{fig:rl}E summarize the resulting $N=177$ \footnote{We are collecting more human judgments.} reliable case-judgments: the head-to-head preference distribution and the per-axis score distributions.

In the head-to-head comparison, experts rated SciReasoner tie-or-exceed than DeepSeek-V4-Pro in $98\%$ of case-judgements, with $73\%$ strongly preferring SciReasoner, $21\%$ preferring it, and $4\%$ ties. Only a small residual fraction favoured DeepSeek-V4-Pro (Fig.~\ref{fig:rl}E, pairwise preference). 
The preference is consistent across all three tasks --- $100\%$ tie-or-exceed on GO protein function, $95\%$ on crystalline-material property prediction, and $98\%$ on retrosynthesis. It demonstrates that SciReasoner's advantage is not carried by any single domain but holds across heterogeneous scientific-reasoning settings. 

The per-axis scores show that this preference reflects a broad, consistent quality gap rather than a single dominant dimension. SciReasoner averages $8.7/10$ overall versus $4.3/10$ for DeepSeek-V4-Pro, with per-axis means of $8.8$ vs $4.5$ on evidence grounding, $8.5$ vs $4.4$ on domain plausibility, $8.9$ vs $4.3$ on target-regime alignment, $8.7$ vs $3.9$ on reasoning coherence, and $8.8$ vs $4.6$ on anti-hallucination. 
Every per-axis difference is significant shown by Wilcoxon signed-rank on paired per-case differences, $P<0.001$ in Fig.~\ref{fig:rl}E, per-axis scores. 
SciReasoner's distributions are concentrated near the top of the scale on all five axes, whereas DeepSeek-V4-Pro receives more variable scores, especially on target-regime alignment and reasoning coherence. This indicates that the most reliable separation is not that DeepSeek-V4-Pro uniformly fails, but that SciReasoner is more consistently grounded, task-aligned, and coherent across cases.

Together, these results show that SciReasoner's advantage on automated metrics also holds under direct expert scrutiny: across three heterogeneous tasks, human experts prefer or tie SciReasoner reasoning in the overwhelming majority of cases, and rate it markedly higher on every quality axis, reflecting faithful, verifiable scientific reasoning.

\section{Discussion}\label{sec12}

In this work, we introduced \projName{}, a multimodal scientific foundation model for native structural reasoning across proteins, small molecules and inorganic crystals.
The central premise is that structure--property relationships cannot be fully addressed by treating scientific entities as text strings, low-dimensional descriptors or black-box inputs to property predictors.
\projName{} instead represents structures as primary objects of inference through a unified structure-aware vocabulary that discretizes three-dimensional coordinates, molecular topologies and crystallographic lattices into domain-native tokens, and integrates them with language instructions in a single autoregressive model.
This design allows residues, molecular fragments, conformational cues and crystal descriptors to function as addressable evidence units within generated reasoning trajectories, rather than as peripheral annotations appended to text.

Our results support native structural reasoning as a model-level capability for scientific AI.
Rather than using structure only as an input format or auxiliary descriptor, \projName{} makes residues, molecular fragments, conformational cues and crystal descriptors available as evidence within generated reasoning trajectories.
This shift is important because structure--property analysis requires not only accurate outputs, but intermediate claims that can be inspected against the physical object being studied.

The strongest evidence for this capability comes from regimes in which superficial similarity is insufficient.
 
For Cellular Component annotation of low-homology and orphan-like proteins, \projName{} improved $F_{\max}$ from 0.42 to 0.55, with its attention enriched at contact-defined DNA-binding residues and protein--DNA interfaces. 
In retrosynthesis, it increased single-step accuracy from 0.63 to 0.72 while identifying reaction-relevant fragments, strategic bond disconnections and chemically valid precursors.
In crystalline materials, its reasoning trajectories invoked symmetry-distinct sites, coordination environments and periodic bonding motifs when linking structure to electronic and thermodynamic behaviour.
These behaviours indicate that structural organization functions as the working substrate of inference, not merely as a representation used to produce a final score.

This distinction clarifies the contribution of \projName{} relative to existing scientific AI systems.
A Foldseek-derived structural fragment can ground a fold- or function-level hypothesis; a molecular substructure can justify a retrosynthetic disconnection; a conformational cue can support a pharmacophore-level similarity judgement; and a crystal-connectivity token can support an inference about stability or electronic response.
Because such intermediate claims are expressed in the reasoning trajectory, they can be checked against the input structure rather than accepted as opaque post hoc rationales.
This helps bridge two previously separated capabilities: domain-native structural representation and explicit scientific reasoning.
The double-blind expert evaluation, in which experts preferred or tied \projName{} over a frontier large language model in 98\% of cases, further suggests that this inspectability is practically useful for scientific assessment.

Overall, \projName{} demonstrates that a single autoregressive foundation model can unify sequence, structure, and natural-language reasoning across major scientific modalities while achieving specialist-level accuracy and interpretable scientific explanation. 
This provides a route toward scientific AI systems that move beyond prediction toward reasoning over the physical structures from which functions, properties, and mechanisms emerge.

\section{Method}\label{sec13}
\subsection{Data Source and Processing}
\subsubsection{Protein Data}

We constructed the protein component of the pretraining corpus by following established protein-language data construction protocols. 
To incorporate textual and functional protein knowledge, we used SIFTS-based PDB--UniProt mappings as indexing anchors~\citep{dana2019sifts,berman2000pdb,uniprot2023uniprot}, protein sequences were associated with protein names, PubMed/PMC literature~\citep{pubmed,pubmed_central} and curated UniProtKB/Swiss-Prot annotations, including functional descriptions, domain and subunit information, post-translational modifications, disease associations, induction information and tissue specificity when available. 

For large-scale UniProt protein sequences, we retrieved the corresponding AlphaFold-predicted structures from AlphaFoldDB whenever available~\citep{varadi2022alphafold}. Following the SaProt pretraining setting~\citep{su2025saprothub,su2024saprot}, each protein structure was converted into residue-level Foldseek 3Di tokens~\citep{van2024fast}, which describe local structural environments and are aligned one-to-one with the amino-acid sequence. The amino-acid tokens and 3Di tokens were then separated into amino acid (AA) sequences (wrapped by \verb|<protein></protein>| tag) and structural sequences (warped by \verb|<protein_structure></protein_structure>| tag) in one input, allowing the model to learn from both primary sequence and predicted three-dimensional structure. Structurally unreliable regions (pLDDT$<$70) were masked or replaced by a special structural token to reduce noise from uncertain coordinates. 

To prevent the leakage of labels in the benchmark data within the interleaved protein--text data, we excluded from continued pretraining all text-linked protein records with $>$30\% MMseq2~\citep{steinegger2017mmseqs2,kshirsagar2025rapid} sequence identity to any protein in the held-out test sets of the downstream benchmarks.

\subsubsection{Small-molecule Data}

We constructed the small-molecule component by integrating chemical text, molecular representations, structured property datasets and three-dimensional molecular conformations. The textual component was collected with US Chemistry Olympiad problems, Europe PMC full-text articles and abstracts, ChemRxiv, bioRxiv, medRxiv and arXiv submissions filtered for chemistry, medicinal chemistry and physical chemistry. These sources provide foundational chemical concepts, textbook-style explanations, research-level chemical knowledge, experimental context, safety descriptions and domain-specific terminology.

We further converted public molecular property datasets into language-interfaced templates, pairing molecular entities with curated descriptions of properties, assays, labels and task contexts. 
These sources include MoleculeNet~\citep{wu2018moleculenet}, Therapeutics Data Commons and ADMET-style datasets~\citep{huang2021therapeutics}, ChEBI, PubChem BioAssay, Tox21, hERG and CYP bioactivity datasets, CHEMDNER and other biomedical named-entity recognition corpora, and curated datasets covering ADMET, physicochemical properties, molecular toxicity and safety-related endpoints. 
Together, these resources use natural language to align standardized molecular representations with property labels, bioactivity assays, toxicity endpoints, pharmacokinetic profiles and biomedical chemical terminology. 
We also incorporated reaction, biochemical and spectral resources, including Open Reaction Database (ORD), USPTO-derived reactions, Buchwald--Hartwig datasets, Rhea/UniProt reaction annotations, MOSES~\citep{10.3389/fphar.2020.565644}, MoNA and spectrum-reasoning datasets, to support learning of reactant--product mappings, reaction procedures, enzymatic transformations, molecular understanding and spectroscopy-based reasoning.

To incorporate explicit three-dimensional molecular information, we additionally constructed a conformation-aware molecular corpus following the representation learning dataset of ConfSeq~\citep{xiong2025bridging}. Molecules collected from ChEMBL and BindingDB were converted into three-dimensional conformers using RDKit-based conformer generation. The resulting conformations were represented in a sequence-compatible format that combines molecular connectivity with internal-coordinate information and is wrapped by \verb|<molecule_structure></molecule_structure>| tag, linking with natural language to allow the model to learn geometric features such as molecular shape, bond geometry and conformational variation within a language-modeling framework. To avoid label leakage, we canonicalized molecular identifiers and excluded pretraining examples whose molecule--label or reaction product entries overlapped with the held-out test sets of the corresponding downstream benchmarks. 

\subsubsection{Materials Data}

We constructed the materials component by collecting crystal compositions, crystal information files, structure-derived textual descriptions and tabular materials properties from public materials databases. The primary sources include 
Materials Project, JARVIS-DFT, SNUMAT, hMOF, QMOF,
OQMD\footnote{\url{https://www.oqmd.org/}},
OMDB\footnote{\url{https://omdb.mathub.io/}},
JARVIS-QETB\footnote{\url{https://jarvis.nist.gov/jarvisqetb}},
GNoME\footnote{\url{https://github.com/google-deepmind/materials_discovery/blob/main/DATASET.md}}, and Cantor HEA\footnote{\url{https://doi.org/10.5281/zenodo.10854500}}
These sources cover a broad range of inorganic crystals, semiconductors, metal--organic frameworks, organic materials, high-entropy alloys and computationally discovered stable materials.
 
For each material, we retained complementary representations such as chemical formula, elemental composition, periodic crystal structure in CIF format and natural-language structural descriptions generated from crystallographic information. We then used natural language templates to link these representations to form the training samples, enabling the model to learn composition-level, geometry-level and text-level representations of periodic systems. 

For data processing, we used an 80/10/10 random split at the material-sample level, where each sample contains one material together with all associated property labels. To prevent leakage, all validation and test material samples, including their structural representations and property records, were removed from the continued-pretraining corpus before training.

\subsubsection{DNA/RNA Data}
We constructed the RNA component from RNAcentral, retaining sequences in FASTA format and wrapping them with \verb|<rna></rna>| tags. Available metadata, such as RNA family information, was incorporated to provide functional and contextual cues. 
For DNA, genomic sequences were retrieved from NCBI, sampled into 1,000-bp organism-level fragments, represented in FASTA format and wrapped with \verb|<dna></dna>| tags. Organism names and genomic coordinates were retained to anchor each sequence within its genomic context.

\subsubsection{General Text and Instruction Following Data}

To complement the scientific corpora with general-domain text and instruction-following supervision, we draw from three publicly released sources. For general web text, we sample from Nemotron-CC~v2~\citep{nvidia2025nvidianemotronnano2}, a multi-trillion-token, quality-filtered Common Crawl derivative. For reasoning-formatted instruction data, we use the mathematics and instruction-following splits of {Dolci-Think-SFT-32B}~\citep{olmo2025olmo3}, in which every response carries an explicit \texttt{<think>}-delimited deliberation trace covering multi-step symbolic reasoning and constraint-driven instruction following. For scientific instruction adherence, we additionally include the training split of {SciIF}~\citep{su2026sciif}, whose prompts impose verifiable constraints over scientific contexts.

\subsection{Model Architecture}
The overall architecture of \projName~is designed to operate as a unified causal language model capable of processing interleaved structural and textual modalities. 
The system consists of three primary components: modality-specific offline structural compressors, a structure-aware vocabulary embedding layer (acting as a discrete cross-modal projection), and a unified large language model (LLM) backbone, denoted as $f_\phi$. 
The LLM backbone is initialized with the weights of Qwen3-14B~\citep{yang2025qwen3}. 
\subsubsection{Offline Structure Encoder}
A critical limitation of applying generalist LLMs directly to scientific data is the reliance on text-based sub-word tokenizers, such as Byte-Pair Encoding~\citep{sennrich2016neural,gage1994new}, which arbitrarily fragment physical topologies. 
To prevent this, we employ a modality-specific offline compression strategy. Let $S \in \{S_{prot}, S_{chem}, S_{cryst}\}$ represent the raw structure input data for proteins, small molecules, and crystals, respectively. As shown in Fig.~\ref{fig:main_figure}B, we utilize domain-specific discrete encoders to encode $S$ into the structural information sequence $X_v$. Crucially, this offline process guarantees that fundamental physical semantics (such as local chemical motifs, three-dimensional geometry, and lattice symmetries) are preserved as distinct, holistic tokens prior to interacting with the language model. 
Specifically, we adopt: (i) Foldseek~\citep{van2024fast} for encoding the 3D structure of protein, (ii) SLICES~\citep{xiao2023invertible} for crystal, and (iii) ConfSeq~\citep{xiong2025bridging} for 3D molecular structures. 
The encoded structural information sequences are tagged with corresponding structure tag, i.e., \texttt{<material\_structure></material\_structure>},  \texttt{<molecule\_structure></molecule\_structure>},  and \texttt{<protein\_structure></protein\_structure>}. 

\subsubsection{Structure-Aware Vocabulary Embedding}
Unlike traditional multimodal frameworks that rely on heavy continuous encoders (e.g., Graph Neural Networks) during the forward pass, \projName~treats the discrete structural sequence $X_v$ identically to linguistic scientific tokens. We introduce a dedicated structure-aware vocabulary, which functions as a discrete cross-modal projection layer. Let $\mathbf{W}_v \in \mathbb{R}^{|V_v| \times d_{LLM}}$ be the learnable embedding matrix for the structural vocabulary, where $|V_v|$ is the vocabulary size of all structural motifs and features, and $d_{LLM}$ is the hidden dimension of the LLM. 
The structural sequence $X_v$ is directly projected into the dense continuous space via a discrete embedding lookup: 
\begin{equation}
    \mathbf{H}_v = \text{Embedding}(X_v, \mathbf{W}_v) \in \mathbb{R}^{L_v \times d_{LLM}}
\end{equation}
This discrete projection elegantly bypasses the computational bottleneck of continuous spatial encoders, seamlessly aligning structural inputs with the LLM's native semantic space. 

\subsubsection{Unified Autoregressive Generation}
Simultaneously, the language instruction $X_q$ is tokenized and embedded by the LLM's native embedding layer to produce the language representation $\mathbf{H}_q \in \mathbb{R}^{L_q \times d_{LLM}}$. The projected structural embeddings and language embeddings are concatenated along the sequence dimension to form the unified input prompt $\mathbf{H}_{prompt} = [\mathbf{H}_v; \mathbf{H}_q]$. The LLM backbone $f_\phi$ takes $\mathbf{H}_{prompt}$ as input and generates the response sequence $X_a$ in an autoregressive manner, ensuring the generation is strictly conditioned on both the structural evidence and the textual instruction.

\subsection{Pretraining}
\subsubsection{Optimization Objective and Multi-Stage Pretraining}

The entire multi-stage pretraining pipeline is optimized using a single, unified autoregressive causal language modeling objective (Next Token Prediction).  

Given a structural input $\mathbf{H}_v$ and a language instruction $\mathbf{H}_q$, the model generates a target response sequence $X_a = \{x_{a,1}, x_{a,2}, ..., x_{a,T}\}$. The training objective minimizes the negative log-likelihood of the target tokens conditioned on the preceding context:
\begin{equation}
    \mathcal{L}_{NTP} = -\sum_{t=1}^{T} \log P_\phi (x_{a,t} \mid x_{a,<t}, \mathbf{H}_v, \mathbf{H}_q)
\end{equation}
where $\phi$ represents the trainable parameters of the system at any given stage. To effectively align the distinct physical structural space with the natural language space without catastrophic forgetting, we implement a three-stage curriculum learning strategy, distinguished exclusively by parameter freezing configurations, data mixture, and learning rate dynamics. Let $\Theta = \{\theta_{vocab}, \theta_{emb}, \theta_{head}, \theta_{backbone}\}$ denote the complete parameter set, corresponding to the structure-aware vocabulary layer ($\mathbf{W}_v$), LLM textual token embeddings, LLM prediction head, and the core LLM transformer backbone, respectively.

\paragraph{Stage 1: Warm-up Training}
The initial stage focuses on mapping the newly introduced structural tokens into the linguistic semantic space. To prevent the untrained structure-aware vocabulary layer from corrupting the pre-trained reasoning capabilities of the LLM, we strictly freeze the core transformer backbone. 
In this stage, the trainable parameter subset is $\phi = \{\theta_{vocab}, \theta_{emb}, \theta_{head}\}$, while $\theta_{backbone}$ remains frozen. The model is trained on fundamental structure-to-text reconstruction tasks. By restricting gradient updates to the peripheral layers, this stage establishes a stable foundation in the embedding space before deeper integration. To ensure localized and stable optimization for these newly initialized parameters, Stage 1 maintains an independent optimizer state and employs its own standalone learning rate schedule.

\paragraph{Stage 2: Full-parameter Training}
Once the structural tokens are stably anchored in the embedding space, we proceed to align the modalities across the deep layers of the network. In this stage, all parameters are unfrozen ($\phi = \Theta$). For the continuous optimization dynamics spanning Stage 2 and Stage 3, we initialize a new shared optimizer state and a unified Warmup-Stable-Decay (WSD) learning rate scheduler. 

Formally, let $t$ denote the training step across the combined duration of Stage 2 and Stage 3. The shared WSD scheduler is defined by the durations of the warmup ($T_{w}$), stable ($T_{s}$), and decay ($T_{d}$) phases, with total steps $T_{total} = T_{w} + T_{s} + T_{d}$. The learning rate $\eta(t)$ is formulated as:
\begin{equation}
    \eta(t) = \begin{cases}
        \eta_{max} \cdot \frac{t}{T_{w}} & \text{if } 0 \le t < T_{w} \\
        \eta_{max} & \text{if } T_{w} \le t < T_{w} + T_{s} \\
        f_{decay}(t) & \text{if } T_{w} + T_{s} \le t \le T_{total}
    \end{cases}
\end{equation}
where $\eta_{max}$ is the peak learning rate and $f_{decay}(t)$ represents a monotonically decreasing annealing function (e.g., cosine decay). 

Stage 2 spans the interval $0 \le t < T_{stage2}$, where $T_{w} < T_{stage2} < T_{w} + T_{s}$. This indicates that Stage 2 strictly encompasses the entire warmup phase and the initial segment of the stable phase. During this period, the model is trained on diverse, broadly-sourced paired structure-text datasets, allowing the full network to assimilate domain-specific structural knowledge at a high, constant learning rate.

\paragraph{Stage 3: Annealing Training} 
In the final stage, we maintain the full-parameter optimization setting ($\phi = \Theta$) and seamlessly inherit both the optimizer state and the active WSD scheduler from Stage 2. Stage 3 covers the remainder of the training timeline ($T_{stage2} \le t \le T_{total}$). Thus, it begins by continuing the constant learning rate $\eta_{max}$ from the stable phase and eventually triggers the decay phase ($t \ge T_{w} + T_{s}$) in its later period. 
In this training stage, we increase the proportion of question--answer (QA)-style data to achieve annealed training. 
\subsection{Post-training}

The pretraining curriculum endows \projName~with broad cross-modal scientific knowledge, yet the resulting checkpoint $M_0$ behaves as a next-token continuator rather than an instruction-following reasoner. Post-training instantiates self-bootstrapped native structural reasoning through a coldstart supervised stage followed by reinforcement learning (RL). Pooling the chain-of-thought (CoT) supervision of all tasks into a single pass is, however, ill-suited here: the reasoning patterns of heterogeneous scientific tasks are divergent enough that joint training induces destructive interference, and the seed CoT traces, borrowed from external teachers, are distributionally misaligned with \projName.

We therefore organize post-training around two stages: \emph{intra-domain structural evidence grounding} followed by \emph{cross-domain reasoning consolidation}. 
In the first stage, we partition the task universe by task prior and specialize $M_0$ into a family of per-group experts that learn how structural tokens serve as domain-specific reasoning evidence. In the second stage, these experts regenerate rational CoT traces that replace the external supervision; pooled across all tasks and augmented with non-thinking data, they drive a single unified coldstart and RL pass yielding the final model. 

\subsubsection{Supervised finetuning for reasoning pattern coldstart}

The coldstart phase seeks to \emph{activate} the capabilities accumulated during pretraining: we elicit instruction following and, most critically, an explicit CoT reasoning behaviour in which the model deliberates before committing to an answer. Because \projName~is initialized from Qwen3~\citep{yang2025qwen3}, which natively supports a switchable thinking interface, we frame this stage as the re-establishment of a structured generation protocol that the backbone has already been exposed to, requiring only a lightweight intervention rather than a wholesale re-training.

\paragraph{Reasoning-formatted data construction} We curate a compact supervised corpus in which every target response adheres to an explicit deliberation template demarcated by dedicated \texttt{<think>} delimiters. We construct two complementary response modes to preserve the dual-mode behavior of the backbone. In the \emph{thinking} mode, the model is prompted with an instruction and produces an interleaved trace of the form
\begin{equation}
    X_a \;=\; \texttt{<think>}\;\backslash\texttt{n}\; c \;\backslash\texttt{n}\; \texttt{</think>}\;\backslash\texttt{n}\backslash\texttt{n}\; y,
\end{equation}
where $c$ denotes the latent chain-of-thought rationale and $y$ the final answer conditioned on $c$. In the \emph{non-thinking} mode, the instruction is appended with a \texttt{/no\_think} control token and the deliberation span is collapsed to an empty placeholder,
\begin{equation}
    X_a \;=\; \texttt{<think>}\;\backslash\texttt{n}\backslash\texttt{n}\;\texttt{</think>}\;\backslash\texttt{n}\backslash\texttt{n}\; y,
\end{equation}
yielding a direct-answer response. Exposing the model to both formats within a single corpus re-anchors the semantics of the \texttt{<think>} delimiters and the \texttt{/no\_think} switch, enabling the controllable allocation of test-time computation between explicit reasoning and direct response.

\paragraph{Supervised optimization objective} The coldstart stage retains the autoregressive next-token-prediction objective used throughout pretraining, but departs from it in the supervision mask. Rather than computing the loss over the entire sequence, we adopt a supervised-finetuning (SFT) formulation in which gradients are propagated solely through the response tokens, while the instruction and structural context serve purely as conditioning. Concretely, let the full sequence be partitioned into the prompt context $(\mathbf{H}_v, \mathbf{H}_q)$ and the target response $X_a = \{x_{a,1}, \dots, x_{a,T}\}$, where $X_a$ encompasses both the deliberation span and the final answer. The coldstart objective minimizes the negative log-likelihood restricted to the response positions:
\begin{equation}
    \mathcal{L}_{SFT} = -\sum_{t=1}^{T} m_t \, \log P_\phi \big( x_{a,t} \mid x_{a,<t}, \mathbf{H}_v, \mathbf{H}_q \big),
\end{equation}
where the binary mask $m_t \in \{0, 1\}$ activates supervision only on the response segment ($m_t = 1$ for tokens in $X_a$) and suppresses the loss contribution of the prompt tokens ($m_t = 0$). This response-only masking ensures that the optimization sharpens the conditional mapping from instruction to reasoned answer, instead of expending capacity on modelling the marginal distribution of the inputs.

\paragraph{Coldstart} In the first stage, a coarse task prior partitions the task universe $\mathcal{T}$ into $J$ disjoint groups $\{\mathcal{G}_1, \dots, \mathcal{G}_J\}$, each aggregating similar tasks with CoT corpus $\mathcal{C}_j$ whose rationale $c$ is teacher-supplied. We coldstart $M_0$ independently on each $\mathcal{C}_j$ under $\mathcal{L}_{SFT}$, yielding specialized checkpoints $\{M_j^{\text{cs}}\}_{j=1}^{J}$ so that each internalizes a coherent reasoning style rather than reconciling conflicting ones. In the second stage, after the experts $\{M_j^{\text{expert}}\}$ are obtained by the \emph{intra-domain structural evidence grounding} stage below, we replace the external supervision with \projName-generated traces: each $M_j^{\text{expert}}$ rolls out on its group's prompts, and verification-passing traces $\tilde{c}$ replace the external $c$ in $\mathcal{C}_j$, giving a new corpus $\tilde{\mathcal{C}}_j$. Augmenting each $\tilde{\mathcal{C}}_j$ with \texttt{/no\_think} direct-answer data and pooling across groups, $\tilde{\mathcal{C}} = \bigcup_{j=1}^{J} \tilde{\mathcal{C}}_j$, we perform a single unified coldstart from $M_0$ under the same $\mathcal{L}_{SFT}$. As the targets now lie on the model's own manifold, this furnishes a well-matched initialization for the \emph{cross-domain reasoning consolidation} pass.

\subsubsection{Reinforcement learning}
For each sub-task, we curate RL data with an empirical solve-rate filter that selects medium-difficulty
  instances.

  Let \(D^{\text{train}}_{k}\) denote the training set for the \(k\)-th sub-task. For each example \(x \in
  D^{\text{train}}_{k}\):
  \begin{enumerate}
      \item \textbf{Stochastic rollouts.} We generate \(N=8\) independent rollouts from \(M_0\) on \(x\) with stochastic decoding and set \texttt{do\_sample=True}. To balance response diversity and format fidelity, we use an adaptive temperature \(T\), tuned once for each sub-task before collection. Starting from \(T=0.9\), we increase \(T\) when the eight samples show insufficient diversity, such
      as low uniqueness among normalized responses or highly repetitive content. If the outputs become too diffuse and frequently violate the required format or fail to parse, we decrease \(T\). For the final eight rollouts for each \(x\), we keep \(T\) fixed and vary only the random seed, while holding all other decoding parameters constant.

      \item \textbf{Empirical solve rate.} For rollout \(i\), we compute a correctness indicator \(f_i(x)\in\{0,1\}\) using the sub-task's canonical metric, such as exact match or a thresholded score, and aggregate
      \begin{equation}
          \hat p(x) \;=\; \frac{1}{8}\sum_{i=1}^{8} f_i(x).
      \end{equation}

      \item \textbf{Mid-difficulty filtering.} We retain examples with partial success:
      \begin{equation}
          S_k \;=\; \left\{\,x \in D^{\mathrm{train}}_{k} \;\middle|\; 0.125 < \hat p(x) < 0.875 \,
          \right\}.
      \end{equation}
      The retained examples are neither trivially solved nor consistently failed, and thus provide more informative RL learning signals than cases with \(\hat p(x)\approx 1\) or \(\hat p(x)\approx 0\).
    \item \textbf{Difficulty-centred sampling.} We construct the RL subset \(T_k\) by ranking candidate examples according to their distance from intermediate success,
    \begin{equation}
        d(x) \;=\; \left|\hat p(x)-0.5\right|.
    \end{equation}
    We then select \(K=2000\) examples without replacement, first from \(S_k\) in ascending order of \(d(x)\):
    \begin{equation}
        T_k \;=\;
        \begin{cases}
        \operatorname*{arg\,top}_{K,\,x \in S_k} \bigl(-d(x)\bigr),
        & |S_k| \geq K, \\[4pt]
        S_k \cup
        \operatorname*{arg\,top}_{K-|S_k|,\,x \in D^{\mathrm{train}}_{k}\setminus S_k}
        \bigl(-d(x)\bigr),
        & |S_k| < K .
        \end{cases}
    \end{equation}
    Here, \(\operatorname*{arg\,top}_{m,\,x\in A} f(x)\) denotes the \(m\) examples in a candidate set \(A\) with the largest values of \(f(x)\). Since \(f(x)=-d(x)\), this operator selects examples with the smallest distance to \(\hat p(x)=0.5\). Thus, the procedure starts from examples closest to intermediate success and gradually expands to examples with success rates farther from \(0.5\) until \(K=2000\) examples are collected.
\end{enumerate}

The resulting 2,000-example set serves as the RL training pool for that sub-task.

Focusing RL on neither-trivial-nor-impossible instances yields stronger and more stable improvement: items with \(\hat p \in (0.125,0.875)\) (i) expose the policy to cases where exploration can flip outcomes, (ii) avoid vanishing advantages on saturated positives, and (iii) reduce uninformative gradients from uniformly failed cases. 

Model training is performed with DAPO~\citep{yu2026dapo}, which optimizes the policy with RL across heterogeneous scientific tasks. Given a prompt, we first sample \(G\) candidates from the old policy \(\pi_{\theta_{\text{old}}}\). The corresponding within-group standardized advantages are computed as
\begin{equation}
  r_{i,t}(\theta)=\frac{\pi_{\theta}(o_{i,t} \mid q, o_{i,<t})}{\pi_{\theta_{\text{old}}}(o_{i,t} \mid
  q,o_{i,<t})},\quad
  \hat{A}_{i,t} = \frac{R_i - \text{mean}(\{R_i\}_{i=1}^G)}{\text{std}(\{R_i\}_{i=1}^G)}.
\label{eq:advantage_calculation}
\end{equation}

We then optimize the following PPO-style clipped objective:
\begin{equation}
\begin{aligned}
\mathcal{J}_{\text{DAPO}}(\theta) =\quad& \mathbb{E}_{(q,a)\sim \mathcal{D}, \{o_i\}_{i=1}^G\sim
\pi_{\theta_\text{old}}(\cdot\mid q)}\\&
\Bigg[\frac{1}{\sum_{i=1}^{G}|o_i|}\sum_{i=1}^{G}\sum_{t=1}^{|o_i|}
\min \Big( r_{i,t}(\theta) \hat{A}_{i,t},
\ \text{clip} \Big( r_{i,t}(\theta), 1 - {\varepsilon_{\text{low}}}, 1 + {\varepsilon_{\text{high}}}
\Big) \hat{A}_{i,t} \Big) \Bigg]
\\
\text{s.t.}\quad& 0< \Big|\{o_i\mid\texttt{is\_equivalent}(a,o_i)\}\Big|< G,
\end{aligned}
\label{eq:dapoloss}
\end{equation}
The asymmetric \emph{Clip-Higher} schedule is used to preserve exploration. We also apply \emph{Dynamic Sampling} to remove degenerate groups whose rewards are identical, and use a token-level policy-gradient reduction.

The original DAPO reward is rule-based and binary, whereas many of our tasks, including free-form semantic judgment, information extraction, and numeric regression, cannot be evaluated by a single exact-match criterion. We therefore substitute the binary reward with a \emph{reward softening} scheme, which converts grouped task metrics to a shared \([0,1]\) scale: 
\begin{equation}
\label{eq:task_aware_soft_reward}
R_i^{\text{soft}}
\;=\;
g\!\big(m_i(q,o_i,\mathcal{A}(q))\big)
\end{equation}
Here, \(g(\cdot)\) is a monotone calibration function that keeps reward magnitudes comparable across tasks, and \(m_i\) denotes the \emph{group-specific} quality metric. Under this grouping scheme, scientific prediction tasks use distance-based rewards, retrieval and extraction tasks use matching-based rewards, and tool-verified rewards are computed with professional scientific software.

\paragraph{Intra-domain structural evidence grounding and cross-domain reasoning consolidation} The procedure above is first run independently per group as intra-domain structural evidence grounding: starting from $M_j^{\text{cs}}$, the solve-rate filtering and DAPO objective of Eqs.~\eqref{eq:advantage_calculation}--\eqref{eq:task_aware_soft_reward} are applied on the data of $\mathcal{G}_j$ alone, producing the experts $\{M_j^{\text{expert}}\}_{j=1}^{J}$ that supply the on-policy traces for the unified coldstart. It is then run once more over the merged all-task pool as cross-domain reasoning consolidation, starting from the unified coldstart checkpoint. Because the policy is now initialized on expert-generated on-policy supervision from every group, this final pass consolidates the experts' specialized competences without the interference of a naive joint optimization, yielding the final unified model $M_{\text{final}}$.

\backmatter

\clearpage

\begin{appendices}

\providecommand{\BestPerf}[1]{\textbf{#1}}
\providecommand{\SecBestPerf}[1]{\underline{#1}}

\section{Detailed experimental results}\label{secB1}

We report task-level evaluation results for all benchmark tasks with available model measurements. SciReasoner is compared with four frontier general-purpose models: Opus-4.7, GPT-5.5, Kimi-K2.6 and DeepSeek-V4-Pro. The result tables are organized by discipline: Chemistry, Material Science, and Biology. Within each table, rows are further grouped by task type: Scientific QA, Property Prediction, Property Classification, or Generation and Design.

Best and second-best results are highlighted in bold and underlined, respectively. Metrics marked with $\uparrow$ are better when larger, whereas metrics marked with $\downarrow$ are better when smaller. 

\subsection{Task and metric descriptions}\label{secB1_task_intro}

The descriptions below follow the task organization used in the result tables and clarify both the expected model behavior and the metric used for evaluation.

\paragraph{Chemistry tasks.}
\textbf{Scientific QA.}
\begin{itemize}
    \item \textbf{Chemical entity recognition} (\textbf{F1}): identifies chemical mentions in scientific or biomedical text and evaluates span/entity recovery with a precision--recall balanced score.
    \item \textbf{Chemical protein interaction extraction} (\textbf{F1}): extracts chemical--protein relation statements from text, requiring the model to identify the paired entities and the asserted interaction.
    \item \textbf{Chemical disease interaction extraction} (\textbf{F1}): extracts chemical--disease relation statements from scientific text.
    \item \textbf{Multiple choice question} (\textbf{ACC}): selects the correct answer option for scientific multiple-choice questions, testing factual knowledge and reasoning.
    \item \textbf{True or false question} (\textbf{ACC}): judges whether a scientific statement is correct, incorrect, or unsupported under the task format.
    \item \textbf{Open question} (\textbf{BertScore}): generates free-form answers for scientific questions and compares semantic similarity with reference answers.
    \item \textbf{Name conversion-s2i} (\textbf{Split Match}): converts a SMILES string into an IUPAC-style molecular name and checks component-level string agreement.
    \item \textbf{Name conversion-s2f} (\textbf{Element Match}): converts a SMILES string into a molecular formula and checks whether element composition is preserved.
    \item \textbf{Name conversion-i2s} (\textbf{Exact Match}): converts an IUPAC-style name into a SMILES string and requires exact agreement with the reference.
    \item \textbf{Name conversion-i2f} (\textbf{Element Match}): converts an IUPAC-style name into a molecular formula and evaluates element-level formula correctness.
    \item \textbf{Molecular description generation} (\textbf{ROUGE-L}): generates a molecule description and compares it with the reference text.
    \item \textbf{Molecule captioning} (\textbf{MENTOR}): produces molecule-level captions that capture chemical semantics and is evaluated with a molecule-captioning metric.
\end{itemize}

\textbf{Property Prediction.}
\begin{itemize}
    \item \textbf{Estimated solubility (ESOL) prediction} (\textbf{RMSE}): predicts aqueous solubility from molecular representation; lower root mean squared error indicates better numeric prediction.
    \item \textbf{3D molecule structure (DUD-E)} (\textbf{5.0\% EF}): ranks compounds in a DUD-E-style virtual-screening setting and evaluates early enrichment among the top 5.0\% candidates.
    \item \textbf{Lipophilicity (LIPO) prediction} (\textbf{RMSE}): predicts lipophilicity-related continuous values from molecular structure.
    \item \textbf{Physicochemical prediction} (\textbf{MAE}): predicts continuous physicochemical endpoints from molecular structure and reports average absolute error.
\end{itemize}

\textbf{Property Classification.}
\begin{itemize}
    \item \textbf{Blood-brain barrier permeability (BBBP) prediction} (\textbf{ACC}): classifies whether a molecule can pass the blood--brain barrier.
    \item \textbf{Clinical toxicity (ClinTox) prediction} (\textbf{ACC}): classifies molecular clinical toxicity labels.
    \item \textbf{HIV Prediction} (\textbf{ACC}): classifies whether a molecule is active against HIV replication.
    \item \textbf{Side Effect Resource (SIDER) Prediction} (\textbf{ACC}): classifies side-effect associations for a molecule.
\end{itemize}

\textbf{Generation and Design.}
\begin{itemize}
    \item \textbf{Forward synthesis} (\textbf{Exact Match}): generates the expected product or outcome for a synthesis prompt and checks exact string agreement.
    \item \textbf{Forward reaction prediction} (\textbf{Exact Match}): predicts products from specified reactants and reagents.
    \item \textbf{Reagent prediction} (\textbf{Exact Match}): predicts the reagent, catalyst, solvent, or auxiliary component needed for a reaction.
    \item \textbf{Retrosynthesis mol} (\textbf{Exact Match}): proposes precursor reactants for a target molecule in a Mol-Instructions-style retrosynthesis setting.
    \item \textbf{Retrosynthesis USPTO-50K} (\textbf{Exact Match}): proposes reactants for target products in the USPTO-50K retrosynthesis setting.
    \item \textbf{Retrosynthesis smol} (\textbf{Exact Match}): generates plausible precursor molecules for a target product in the SMol-style retrosynthesis setting.
    \item \textbf{Molecule generation} (\textbf{Exact Match}): generates a molecular string that satisfies a given property or constraint prompt.
    \item \textbf{Description guided molecule design} (\textbf{Exact Match}): designs a molecule from a natural-language requirement and checks exact agreement with the target answer.
\end{itemize}

\paragraph{Material science tasks.}
\textbf{Property Prediction.}
For database-level benchmarks that aggregate multiple heterogeneous properties, we follow domain LLM-Prop~\cite{niyongabo2025llm} and report the normalized score \textbf{$\frac{\text{MAD}}{\text{MAE}}$}, where larger values indicate lower error relative to target dispersion. 
\begin{itemize}
    \item \textbf{MP regression} (\textbf{$\frac{\text{MAD}}{\text{MAE}}$}): predicts continuous Materials Project properties, such as band gap, density, volume, formation energy, and stability-related quantities.
    \item \textbf{SNUMAT regression} (\textbf{$\frac{\text{MAD}}{\text{MAE}}$}): predicts SNUMAT material properties, including band-gap and spin-orbit-related targets.
    \item \textbf{JARVIS-DFT} (\textbf{$\frac{\text{MAD}}{\text{MAE}}$}): predicts DFT-derived material properties covering structural, electronic, elastic, dielectric, and thermodynamic quantities.
    \item \textbf{JARVIS-QETB} (\textbf{$\frac{\text{MAD}}{\text{MAE}}$}): predicts quantum-electronic tight-binding properties such as energy and band-gap-related values.
    \item \textbf{GNoME} (\textbf{$\frac{\text{MAD}}{\text{MAE}}$}): predicts large-scale inorganic material properties, including energy, density, volume, and band-gap-related targets.
    \item \textbf{hMOF} (\textbf{$\frac{\text{MAD}}{\text{MAE}}$}): predicts porous-material properties such as CO$_2$ adsorption, pore diameter, void fraction, and surface area.
    \item \textbf{Cantor HEA} (\textbf{$\frac{\text{MAD}}{\text{MAE}}$}): predicts high-entropy alloy properties, including formation energy, energy above hull, volume per atom, and energy per atom.
    \item \textbf{QMOF} (\textbf{$\frac{\text{MAD}}{\text{MAE}}$}): predicts quantum metal--organic framework properties such as total energy, band gap, cavity diameter, and pore-limiting diameter.
    \item \textbf{OQMD} (\textbf{$\frac{\text{MAD}}{\text{MAE}}$}): predicts Open Quantum Materials Database properties such as band gap and formation energy.
    \item \textbf{OMDB} (\textbf{$\frac{\text{MAD}}{\text{MAE}}$}): predicts organic-material database properties, mainly band-gap-related targets.
\end{itemize}

\textbf{Property Classification.}
\begin{itemize}
    \item \textbf{MP classification} (\textbf{AUC}): classifies discrete Materials Project attributes such as direct-gap status or thermodynamic stability.
    \item \textbf{SNUMAT classification} (\textbf{AUC}): classifies SNUMAT material attributes such as direct or indirect band-gap status.
\end{itemize}

\textbf{Generation and Design.}
\begin{itemize}
    \item \textbf{Composition material} (\textbf{SMACT}): generates material compositions under elemental constraints and checks chemical validity.
    \item \textbf{Bulk modulus material} (\textbf{SMACT}): generates material compositions conditioned on a target bulk modulus and evaluates chemical plausibility.
\end{itemize}

\paragraph{Biology tasks.}
\textbf{Scientific QA.}
\begin{itemize}
    \item \textbf{Function} (\textbf{ROUGE-L}): generates protein-function text from biological context and compares it with the reference description.
    \item \textbf{General function} (\textbf{ROUGE-L}): produces broader functional descriptions or annotations for biological sequences.
\end{itemize}

\textbf{Property Prediction.}
\begin{itemize}
    \item \textbf{Fluorescence} (\textbf{Spearman}): predicts protein mutant fluorescence and evaluates whether predicted rankings match reference rankings.
    \item \textbf{Stability} (\textbf{Spearman}): predicts protein stability values and evaluates rank correlation with reference stability.
    \item \textbf{Enhancer activity} (\textbf{HK-PCC}): predicts DNA enhancer activity and reports Pearson correlation for the housekeeping channel.
    \item \textbf{Isoform} (\textbf{R2}): predicts alternative polyadenylation isoform usage from RNA sequence.
    \item \textbf{Mean ribosome loading} (\textbf{R2}): predicts ribosome loading efficiency from RNA sequence.
    \item \textbf{Programmable RNA switches} (\textbf{R2}): predicts ON, OFF, and ON/OFF behavior of programmable RNA switches.
    \item \textbf{CRISPR on target} (\textbf{Spearman}): predicts on-target knockout efficacy for CRISPR guide RNAs.
    \item \textbf{siRNA efficiency} (\textbf{Mixed-score}): predicts siRNA gene-silencing efficiency using both continuous accuracy and range-level agreement.
    \item \textbf{Structural similarity} (\textbf{MAE}): predicts a numeric structural-similarity target and evaluates average absolute deviation.
    \item \textbf{TM-score} (\textbf{Spearman}): predicts or ranks structure-similarity scores and evaluates ordering consistency.
\end{itemize}

\textbf{Property Classification.}
\begin{itemize}
    \item \textbf{Solubility} (\textbf{ACC}): classifies whether a protein sequence is soluble.
    \item \textbf{gSymbol2Tissue} (\textbf{F1}): maps a gene symbol to tissue-expression labels.
    \item \textbf{gName2Cancer} (\textbf{F1}): maps a gene name to associated cancer types.
    \item \textbf{gSymbol2Cancer} (\textbf{F1}): maps a gene symbol to associated cancer types.
    \item \textbf{Antibody antigen} (\textbf{MCC}): predicts whether an antibody and antigen sequence pair interact.
    \item \textbf{RNA protein interaction} (\textbf{MCC}): predicts whether an RNA sequence and a protein sequence interact.
    \item \textbf{Epigenetic marks prediction} (\textbf{MCC}): predicts epigenetic mark presence from DNA sequence.
    \item \textbf{TF-m} (\textbf{MCC}): predicts mouse transcription-factor binding from DNA sequence.
    \item \textbf{Enhancer-promoter interaction} (\textbf{MCC}): classifies whether enhancer and promoter regions interact.
    \item \textbf{PD-prom 300 all} (\textbf{MCC}): detects promoter regions in a 300-bp setting over all examples.
    \item \textbf{PD-prom 300 notata} (\textbf{MCC}): detects 300-bp promoters without TATA motifs.
    \item \textbf{PD-prom 300 tata} (\textbf{MCC}): detects 300-bp promoters with TATA motifs.
    \item \textbf{CPD-prom core all} (\textbf{MCC}): detects core promoter regions over all examples.
    \item \textbf{CPD-prom core notata} (\textbf{MCC}): detects core promoters without TATA motifs.
    \item \textbf{CPD-prom core tata} (\textbf{MCC}): detects core promoters with TATA motifs.
    \item \textbf{TF-h} (\textbf{MCC}): predicts human transcription-factor binding from DNA sequence.
    \item \textbf{Yeast PPI} (\textbf{ACC}): predicts yeast protein--protein interactions.
    \item \textbf{Human PPI} (\textbf{ACC}): predicts human protein--protein interactions.
    \item \textbf{Protein function} (\textbf{ROUGE-L}): predicts protein function annotations or descriptions from sequence.
    \item \textbf{Domain motif} (\textbf{ROUGE-L}): predicts domain or motif descriptions for protein sequences.
    \item \textbf{Non-coding RNA family} (\textbf{ACC}): classifies non-coding RNA sequences into functional families.
    \item \textbf{Modification} (\textbf{ACC}): predicts RNA modification labels.
    \item \textbf{Fold type} (\textbf{ACC}): classifies protein structural fold type.
    \item \textbf{Subcellular localization} (\textbf{ACC}): predicts the cellular localization label of a protein.
    \item \textbf{EC number} (\textbf{Fmax}): predicts enzyme commission annotations for protein sequences.
    \item \textbf{Keywords} (\textbf{F1}): predicts UniProt-style functional keywords.
    \item \textbf{Metal ion binding} (\textbf{ACC}): predicts whether a protein binds metal ions.
    \item \textbf{GO-BP} (\textbf{Fmax}): predicts Gene Ontology biological-process terms.
    \item \textbf{GO-CC} (\textbf{Fmax}): predicts Gene Ontology cellular-component terms.
    \item \textbf{GO-MF} (\textbf{Fmax}): predicts Gene Ontology molecular-function terms.
\end{itemize}

\textbf{Generation and Design.}
\begin{itemize}
    \item \textbf{Function-guided protein design} (\textbf{Normalized SW}): generates a protein sequence from a functional prompt and evaluates sequence similarity to reference proteins.
    \item \textbf{Catalytic activity} (\textbf{ROUGE-L}): describes the enzyme-catalyzed reaction implied by a protein sequence and compares it with the reference text.
\end{itemize}

\paragraph{Metric definitions.}
\begin{itemize}
    \item \textbf{ACC} ($\uparrow$): fraction of samples whose predicted label exactly matches the reference label.
    \item \textbf{AUC} ($\uparrow$): area under the ROC curve; higher values indicate stronger ranking of positive examples above negatives.
    \item \textbf{F1} ($\uparrow$): harmonic mean of precision and recall, used when both false positives and false negatives matter.
    \item \textbf{Fmax} ($\uparrow$): maximum F1 over candidate thresholds, commonly used for multi-label functional annotation.
    \item \textbf{MCC} ($\uparrow$): Matthews correlation coefficient for binary classification; it remains informative when classes are imbalanced.
    \item \textbf{RMSE} ($\downarrow$): root mean squared error for regression, with larger errors penalized more strongly.
    \item \textbf{MAE} ($\downarrow$): mean absolute error between predicted and reference numeric values.
    \item \textbf{$\frac{\text{MAD}}{\text{MAE}}$} ($\uparrow$): ratio between target dispersion and model error; larger values indicate better prediction relative to a mean baseline.
    \item \textbf{Spearman} ($\uparrow$): rank correlation between predicted and reference values.
    \item \textbf{HK-PCC} ($\uparrow$): Pearson correlation coefficient for the housekeeping enhancer-activity output.
    \item \textbf{R2} ($\uparrow$): coefficient of determination, measuring explained variance in regression targets.
    \item \textbf{Mixed-score} ($\uparrow$): composite siRNA score combining numeric error and activity-range agreement.
    \item \textbf{BertScore} ($\uparrow$): semantic similarity between generated and reference text using contextual embeddings.
    \item \textbf{ROUGE-L} ($\uparrow$): longest-common-subsequence overlap between generated text and reference text.
    \item \textbf{MENTOR} ($\uparrow$): molecule-captioning metric for comparing generated molecular descriptions with references.
    \item \textbf{Split Match} ($\uparrow$): component-level match for molecular name conversion outputs.
    \item \textbf{Element Match} ($\uparrow$): element-composition match for molecular formula generation.
    \item \textbf{Exact Match} ($\uparrow$): strict string or structured-answer equality with the reference output.
    \item \textbf{5.0\% EF} ($\uparrow$): enrichment factor in the top 5.0\% of a virtual-screening ranking.
    \item \textbf{SMACT} ($\uparrow$): validity rate under charge-balance and chemical-plausibility checks for generated material compositions.
    \item \textbf{Normalized SW} ($\uparrow$): maximum normalized Smith--Waterman alignment score between generated and reference protein sequences.
\end{itemize}

\subsection{Detailed results}\label{secB1_task_res}

\providecommand{\BestPerf}[1]{\textbf{#1}}
\providecommand{\SecBestPerf}[1]{\underline{#1}}

\setlength{\tabcolsep}{1pt}

Table~\ref{tab:bar_chart_results} summarizes the comparison with specialist baselines. Tables~\ref{tab:res0624_chemistry_grouped}--\ref{tab:res0624_biology_grouped} provide the complete task-level comparison with frontier general-purpose models across Chemistry, Material Science, and Biology. Within each discipline, tasks are organized as Scientific QA, Property Prediction, Property Classification, or Generation and Design.

Across the full benchmark suite, \projName{} leads on \textbf{\totalSOTATasks{}} of 86 tasks. The appendix separates these comparisons for clarity. On the 33 tasks with specialist baselines, \projName{} matches or surpasses the specialist in 26 comparisons~\ref{tab:bar_chart_results}. Against LLM baselines across all 86 tasks, \projName{} is the best-performing model on 75 of 86 tasks (22 of 28 Chemistry tasks, 13 of 14 Material Science tasks, and 40 of 44 Biology tasks). The few non-leading results are concentrated in molecule captioning, selected chemistry QA and classification tasks, bulk modulus material, and a small number of protein-interaction or modification benchmarks, where competing general-purpose models remain strong.

At the discipline level, the Chemistry results indicate broad gains in chemical information extraction, name conversion, molecular description generation, reaction prediction, retrosynthesis, and molecule design (Table~\ref{tab:res0624_chemistry_grouped}). In Material Science, \projName{} achieves consistently strong regression and classification performance, with pronounced margins on JARVIS-QETB, GNoME, QMOF and OQMD (Table~\ref{tab:res0624_material_science_grouped}). In Biology, the model performs robustly across sequence-to-function generation, biological property prediction, promoter and interaction classification and RNA/protein tasks (Table~\ref{tab:res0624_biology_grouped}). Together, these results indicate that the performance gains are not restricted to a single metric or task format, but extend across structured prediction, text generation, classification, and design-oriented settings.

\begin{table}[!t]
    \centering
    \scriptsize
    \begin{tabular}{l c l c c}
    \toprule
    Task & Metric & Specialist method & Specialist & SciReasoner \\
    \midrule
    \addlinespace[0.8em]
    \multicolumn{5}{l}{\raisebox{1.2ex}[0pt][0pt]{\bfseries Generation \& Design}}\\[-0.8ex]
    \quad Retrosynthesis USPTO-50K & Exact Match$\uparrow$ & RSGPT~\cite{deng2025rsgpt} & \SecBestPerf{0.63} & \BestPerf{0.72} \\
    \addlinespace[0.8em]
    \multicolumn{5}{l}{\raisebox{1.2ex}[0pt][0pt]{\bfseries Prediction}}\\[-0.8ex]
    \quad Fluorescence & Spearman$\uparrow$ & SaprotHub~\cite{su2025saprothub} & \SecBestPerf{0.70} & \BestPerf{0.77} \\
    \quad Isoform & R2$\uparrow$ & APARENT~\cite{bogard2019deep} & \SecBestPerf{0.59} & \BestPerf{0.86} \\
    \quad TM-score & Spearman$\uparrow$ & SaprotHub~\cite{su2025saprothub} & \BestPerf{0.83} & \BestPerf{0.83} \\
    \quad ESOL & RMSE$\downarrow$ & MolCLR~\cite{wang2022molclr} & \SecBestPerf{1.11} & \BestPerf{1.03} \\
    \quad GNoME & $\frac{\text{MAD}}{\text{MAE}}\uparrow$ & LLM-Prop~\cite{niyongabo2025llm} & \SecBestPerf{15.60} & \BestPerf{21.91} \\
    \quad QMOF & $\frac{\text{MAD}}{\text{MAE}}\uparrow$ & LLM-Prop~\cite{niyongabo2025llm} & \SecBestPerf{1.96} & \BestPerf{8.61} \\
    \quad MP regression & $\frac{\text{MAD}}{\text{MAE}}\uparrow$ & LLM-Prop~\cite{niyongabo2025llm} & \SecBestPerf{4.39} & \BestPerf{5.83} \\
    \quad JARVIS-DFT & $\frac{\text{MAD}}{\text{MAE}}\uparrow$ & LLM-Prop~\cite{niyongabo2025llm} & \SecBestPerf{2.91} & \BestPerf{5.67} \\
    \quad SNUMAT regression & $\frac{\text{MAD}}{\text{MAE}}\uparrow$ & LLM-Prop~\cite{niyongabo2025llm} & \SecBestPerf{1.51} & \BestPerf{2.26} \\
    \quad hMOF & $\frac{\text{MAD}}{\text{MAE}}\uparrow$ & LLM-Prop~\cite{niyongabo2025llm} & \SecBestPerf{1.48} & \BestPerf{1.67} \\
    \quad OQMD & $\frac{\text{MAD}}{\text{MAE}}\uparrow$ & LLM-Prop~\cite{niyongabo2025llm} & \SecBestPerf{6.02} & \BestPerf{7.22} \\
    \quad OMDB & $\frac{\text{MAD}}{\text{MAE}}\uparrow$ & LLM-Prop~\cite{niyongabo2025llm} & \BestPerf{1.51} & \SecBestPerf{1.50} \\
    \quad DUD-E & 5.0\% EF$\uparrow$ & ConfSeq~\cite{xiong2025bridging} & \SecBestPerf{7.12} & \BestPerf{7.70} \\
    \quad Cantor HEA & $\frac{\text{MAD}}{\text{MAE}}\uparrow$ & LLM-Prop~\cite{niyongabo2025llm} & \BestPerf{8.40} & \SecBestPerf{7.79} \\
    \quad LIPO & RMSE$\downarrow$ & MolCLR~\cite{wang2022molclr} & \BestPerf{0.65} & \SecBestPerf{0.80} \\
    \addlinespace[0.8em]
    \multicolumn{5}{l}{\raisebox{1.2ex}[0pt][0pt]{\bfseries Classification}}\\[-0.8ex]
    \quad BBBP & ACC$\uparrow$ & MolCLR~\cite{wang2022molclr} & \SecBestPerf{0.74} & \BestPerf{0.84} \\
    \quad ClinTox & ACC$\uparrow$ & MolCLR~\cite{wang2022molclr} & \SecBestPerf{0.93} & \BestPerf{0.95} \\
    \quad HIV Prediction & ACC$\uparrow$ & MolCLR~\cite{wang2022molclr} & \SecBestPerf{0.81} & \BestPerf{0.92} \\
    \quad SIDER & ACC$\uparrow$ & MolCLR~\cite{wang2022molclr} & \SecBestPerf{0.68} & \BestPerf{0.74} \\
    \quad MP classification & AUC$\uparrow$ & LLM-Prop~\cite{niyongabo2025llm} & \SecBestPerf{0.72} & \BestPerf{0.73} \\
    \quad TF-m & MCC$\uparrow$ & NT~\cite{dalla2025nucleotide} & \SecBestPerf{0.57} & \BestPerf{0.64} \\
    \quad PD-prom 300 all & MCC$\uparrow$ & NT~\cite{dalla2025nucleotide} & \BestPerf{0.91} & \SecBestPerf{0.89} \\
    \quad CPD-prom core all & MCC$\uparrow$ & NT~\cite{dalla2025nucleotide} & \SecBestPerf{0.67} & \BestPerf{0.68} \\
    \quad RNA protein interaction & MCC$\uparrow$ & RPI-Pred~\cite{suresh2015rpi} & \SecBestPerf{0.74} & \BestPerf{0.81} \\
    \quad Non-coding RNA family & ACC$\uparrow$ & RNA-MSM~\cite{zhang2024multiple} & \SecBestPerf{0.89} & \BestPerf{0.90} \\
    \quad GO-BP & Fmax$\uparrow$ & SaprotHub~\cite{su2025saprothub} & \SecBestPerf{0.49} & \BestPerf{0.52} \\
    \quad GO-CC & Fmax$\uparrow$ & SaprotHub~\cite{su2025saprothub} & \SecBestPerf{0.48} & \BestPerf{0.58} \\
    \quad GO-MF & Fmax$\uparrow$ & SaprotHub~\cite{su2025saprothub} & \BestPerf{0.67} & \SecBestPerf{0.66} \\
    \quad Human PPI & ACC$\uparrow$ & ESM2~\cite{esm2} & \BestPerf{0.77} & \SecBestPerf{0.73} \\
    \quad Subcellular localization & ACC$\uparrow$ & ESM2~\cite{esm2} & \SecBestPerf{0.84} & \BestPerf{0.88} \\
    \quad Metal ion binding & ACC$\uparrow$ & ESM2~\cite{esm2} & \SecBestPerf{0.73} & \BestPerf{0.74} \\
    \quad Solubility & ACC$\uparrow$ & DeepLoc~\cite{almagro2017deeploc} & \BestPerf{0.77} & \SecBestPerf{0.72} \\
    \bottomrule
    \end{tabular}
    \caption{Per-task comparison of SciReasoner against specialist baselines. \BestPerf{Bold} indicates the best performance, and \SecBestPerf{underline} indicates the second best.}
    \label{tab:bar_chart_results}
    \end{table}

\begin{table}[!t]
\centering
\scriptsize
\setlength{\tabcolsep}{1pt}
\begin{tabular}{l c c c c c c}
\toprule
Task & Metric & Opus-4.7 & GPT-5.5 & Kimi-K2.6 & DeepSeek-V4-Pro & SciReasoner \\
\midrule
\multicolumn{7}{l}{\textbf{Scientific QA}}\\
\midrule
\quad Chemical entity recognition & F1$\uparrow$ & \SecBestPerf{0.84} & 0.69 & 0.72 & 0.63 & \BestPerf{0.88} \\
\quad Chemical protein interaction extraction & F1$\uparrow$ & \SecBestPerf{0.35} & 0.05 & 0.18 & 0.03 & \BestPerf{0.36} \\
\quad Chemical disease interaction extraction & F1$\uparrow$ & \SecBestPerf{0.41} & 0.31 & 0.34 & 0.27 & \BestPerf{0.54} \\
\quad Multiple choice question & ACC$\uparrow$ & \SecBestPerf{0.90} & \BestPerf{0.91} & 0.87 & 0.89 & 0.88 \\
\quad True or false question & ACC$\uparrow$ & \BestPerf{0.63} & \SecBestPerf{0.60} & 0.58 & \BestPerf{0.63} & 0.55 \\
\quad Open question & BertScore$\uparrow$ & 0.77 & \SecBestPerf{0.82} & \SecBestPerf{0.82} & \SecBestPerf{0.82} & \BestPerf{0.85} \\
\quad Name conversion-s2i & Split Match$\uparrow$ & \SecBestPerf{0.34} & 0.07 & 0.02 & 0.01 & \BestPerf{0.50} \\
\quad Name conversion-s2f & Element Match$\uparrow$ & \SecBestPerf{0.85} & 0.84 & 0.04 & 0.28 & \BestPerf{0.96} \\
\quad Name conversion-i2s & Exact Match$\uparrow$ & \SecBestPerf{0.69} & 0.58 & 0.19 & 0.35 & \BestPerf{0.87} \\
\quad Name conversion-i2f & Element Match$\uparrow$ & 0.90 & \SecBestPerf{0.91} & 0.40 & 0.64 & \BestPerf{0.93} \\
\quad Molecular description generation & ROUGE-L$\uparrow$ & \SecBestPerf{0.46} & 0.08 & 0.09 & 0.31 & \BestPerf{0.75} \\
\quad Molecule captioning & MENTOR$\uparrow$ & \BestPerf{0.46} & 0.30 & 0.14 & 0.16 & \SecBestPerf{0.45} \\
\midrule
\multicolumn{7}{l}{\textbf{Property Prediction}}\\
\midrule
\quad ESOL & RMSE$\downarrow$ & \BestPerf{0.89} & 8.72 & 137.05 & 5.24 & \SecBestPerf{1.03} \\
\quad DUD-E & 5.0\% EF$\uparrow$ & 3.77 & \SecBestPerf{4.98} & 1.85 & 1.28 & \BestPerf{7.70} \\
\quad LIPO & RMSE$\downarrow$ & \SecBestPerf{0.97} & 1.48 & 16.25 & 3.29 & \BestPerf{0.80} \\
\quad Physicochemical prediction & MAE$\downarrow$ & \SecBestPerf{19.52} & 25.79 & 2158.02 & 73.24 & \BestPerf{4.58} \\
\midrule
\multicolumn{7}{l}{\textbf{Property Classification}}\\
\midrule
\quad BBBP & ACC$\uparrow$ & 0.73 & \SecBestPerf{0.81} & 0.54 & 0.69 & \BestPerf{0.84} \\
\quad ClinTox & ACC$\uparrow$ & \SecBestPerf{0.88} & 0.41 & 0.57 & 0.77 & \BestPerf{0.95} \\
\quad HIV Prediction & ACC$\uparrow$ & \BestPerf{0.94} & 0.61 & 0.91 & 0.90 & \SecBestPerf{0.92} \\
\quad SIDER & ACC$\uparrow$ & 0.70 & \BestPerf{0.84} & 0.69 & 0.71 & \SecBestPerf{0.74} \\
\midrule
\multicolumn{7}{l}{\textbf{Generation and Design}}\\
\midrule
\quad Forward synthesis & Exact Match$\uparrow$ & \SecBestPerf{0.56} & \SecBestPerf{0.56} & 0.35 & 0.25 & \BestPerf{0.73} \\
\quad Forward reaction prediction & Exact Match$\uparrow$ & 0.84 & 0.82 & \SecBestPerf{0.90} & 0.80 & \BestPerf{0.98} \\
\quad Reagent prediction & Exact Match$\uparrow$ & \SecBestPerf{0.07} & 0.05 & 0.04 & 0.04 & \BestPerf{0.12} \\
\quad Retrosynthesis mol & Exact Match$\uparrow$ & \SecBestPerf{0.43} & 0.29 & \SecBestPerf{0.43} & 0.26 & \BestPerf{0.67} \\
\quad Retrosynthesis USPTO-50K & Exact Match$\uparrow$ & \SecBestPerf{0.48} & 0.31 & 0.10 & 0.15 & \BestPerf{0.72} \\
\quad Retrosynthesis smol & Exact Match$\uparrow$ & \SecBestPerf{0.23} & 0.13 & 0.03 & 0.03 & \BestPerf{0.39} \\
\quad Molecule generation & Exact Match$\uparrow$ & \SecBestPerf{0.44} & 0.43 & 0.15 & 0.30 & \BestPerf{0.50} \\
\quad Description guided molecule design & Exact Match$\uparrow$ & \SecBestPerf{0.10} & \SecBestPerf{0.10} & 0.08 & 0.08 & \BestPerf{0.12} \\
\bottomrule
\end{tabular}
\caption{Evaluation results on Chemistry tasks grouped by task type. \BestPerf{Bold} indicates the best performance, and \SecBestPerf{underline} indicates the second best. 
}
\label{tab:res0624_chemistry_grouped}
\end{table}

\begin{table}[!t]
\centering
\scriptsize
\setlength{\tabcolsep}{1pt}
\begin{tabular}{l c c c c c c}
\toprule
Task & Metric & Opus-4.7 & GPT-5.5 & Kimi-K2.6 & DeepSeek-V4-Pro & SciReasoner \\
\midrule
\multicolumn{7}{l}{\textbf{Property Prediction}}\\
\midrule
\quad MP regression & $\frac{\text{MAD}}{\text{MAE}}\uparrow$ & 2.11 & \SecBestPerf{2.67} & 1.76 & 1.58 & \BestPerf{5.83} \\
\quad SNUMAT regression & $\frac{\text{MAD}}{\text{MAE}}\uparrow$ & 1.51 & \SecBestPerf{1.67} & 1.41 & 1.46 & \BestPerf{2.26} \\
\quad JARVIS-DFT & $\frac{\text{MAD}}{\text{MAE}}\uparrow$ & 1.48 & \SecBestPerf{1.78} & 1.31 & 1.38 & \BestPerf{5.67} \\
\quad JARVIS-QETB & $\frac{\text{MAD}}{\text{MAE}}\uparrow$ & 0.73 & 0.84 & \SecBestPerf{0.88} & 0.86 & \BestPerf{108.98} \\
\quad GNoME & $\frac{\text{MAD}}{\text{MAE}}\uparrow$ & 1.94 & \SecBestPerf{5.39} & 1.56 & 1.22 & \BestPerf{21.91} \\
\quad hMOF & $\frac{\text{MAD}}{\text{MAE}}\uparrow$ & \SecBestPerf{1.08} & 0.99 & 1.00 & 0.92 & \BestPerf{1.67} \\
\quad Cantor HEA & $\frac{\text{MAD}}{\text{MAE}}\uparrow$ & 1.40 & \SecBestPerf{2.72} & 1.44 & 1.29 & \BestPerf{7.79} \\
\quad QMOF & $\frac{\text{MAD}}{\text{MAE}}\uparrow$ & 1.66 & \SecBestPerf{3.12} & 1.56 & 1.35 & \BestPerf{8.61} \\
\quad OQMD & $\frac{\text{MAD}}{\text{MAE}}\uparrow$ & \SecBestPerf{1.97} & 1.84 & 1.03 & 1.46 & \BestPerf{7.22} \\
\quad OMDB & $\frac{\text{MAD}}{\text{MAE}}\uparrow$ & 1.26 & \SecBestPerf{1.43} & 1.07 & 1.10 & \BestPerf{1.50} \\
\midrule
\multicolumn{7}{l}{\textbf{Property Classification}}\\
\midrule
\quad MP classification & AUC$\uparrow$ & 0.61 & 0.65 & \SecBestPerf{0.66} & 0.59 & \BestPerf{0.73} \\
\quad SNUMAT classification & AUC$\uparrow$ & 0.58 & 0.59 & 0.56 & \SecBestPerf{0.60} & \BestPerf{0.68} \\
\midrule
\multicolumn{7}{l}{\textbf{Generation and Design}}\\
\midrule
\quad Composition material & SMACT$\uparrow$ & \SecBestPerf{0.89} & \BestPerf{0.90} & 0.35 & 0.24 & \BestPerf{0.90} \\
\quad Bulk modulus material & SMACT$\uparrow$ & \BestPerf{0.99} & \BestPerf{0.99} & 0.17 & 0.11 & \SecBestPerf{0.88} \\
\bottomrule
\end{tabular}
\caption{Evaluation results on Material Science tasks grouped by task type. \BestPerf{Bold} indicates the best performance, and \SecBestPerf{underline} indicates the second best. 
}
\label{tab:res0624_material_science_grouped}
\end{table}

\begin{table}[!t]
\centering
\scriptsize
\setlength{\tabcolsep}{1pt}
\begin{tabular}{l c c c c c c}
\toprule
Task & Metric & Opus-4.7 & GPT-5.5 & Kimi-K2.6 & DeepSeek-V4-Pro & SciReasoner \\
\midrule
\multicolumn{7}{l}{\textbf{Scientific QA}}\\
\midrule
\quad Function & ROUGE-L$\uparrow$ & \SecBestPerf{0.30} & 0.10 & 0.01 & 0.03 & \BestPerf{0.80} \\
\quad General function & ROUGE-L$\uparrow$ & \SecBestPerf{0.52} & 0.10 & 0.18 & 0.15 & \BestPerf{0.77} \\
\midrule
\multicolumn{7}{l}{\textbf{Property Prediction}}\\
\midrule
\quad Fluorescence & Spearman$\uparrow$ & 0.44 & \SecBestPerf{0.50} & 0.10 & 0.04 & \BestPerf{0.77} \\
\quad Stability & Spearman$\uparrow$ & \SecBestPerf{0.36} & 0.13 & 0.06 & -0.02 & \BestPerf{0.61} \\
\quad Enhancer activity & HK-PCC$\uparrow$ & 0.07 & -0.06 & -0.05 & \SecBestPerf{0.13} & \BestPerf{0.64} \\
\quad Isoform & R2$\uparrow$ & \SecBestPerf{0.09} & 0.01 & 0.05 & 0.01 & \BestPerf{0.86} \\
\quad Mean ribosome loading & R2$\uparrow$ & 0.03 & \SecBestPerf{0.10} & 0.01 & 0.00 & \BestPerf{0.60} \\
\quad Programmable RNA switches & R2$\uparrow$ & \SecBestPerf{0.04} & 0.02 & 0.02 & 0.01 & \BestPerf{0.46} \\
\quad CRISPR on target & Spearman$\uparrow$ & 0.14 & \SecBestPerf{0.28} & 0.05 & \BestPerf{0.40} & 0.12 \\
\quad siRNA efficiency & Mixed-score$\uparrow$ & 0.00 & \SecBestPerf{0.33} & 0.07 & 0.22 & \BestPerf{0.61} \\
\quad Structural similarity & MAE$\downarrow$ & \SecBestPerf{0.18} & 0.27 & 0.47 & 0.56 & \BestPerf{0.05} \\
\quad TM-score & Spearman$\uparrow$ & -0.06 & \SecBestPerf{0.15} & -0.15 & -0.12 & \BestPerf{0.83} \\
\midrule
\multicolumn{7}{l}{\textbf{Property Classification}}\\
\midrule
\quad Solubility & ACC$\uparrow$ & \SecBestPerf{0.58} & 0.54 & 0.53 & \SecBestPerf{0.58} & \BestPerf{0.72} \\
\quad gSymbol2Tissue & F1$\uparrow$ & 0.43 & \SecBestPerf{0.44} & 0.23 & 0.34 & \BestPerf{0.53} \\
\quad gName2Cancer & F1$\uparrow$ & 0.06 & \SecBestPerf{0.12} & 0.00 & 0.01 & \BestPerf{0.73} \\
\quad gSymbol2Cancer & F1$\uparrow$ & \SecBestPerf{0.16} & 0.12 & 0.00 & 0.00 & \BestPerf{0.71} \\
\quad Antibody antigen & MCC$\uparrow$ & 0.08 & \SecBestPerf{0.12} & 0.02 & 0.06 & \BestPerf{0.28} \\
\quad RNA protein interaction & MCC$\uparrow$ & -0.44 & -0.13 & \SecBestPerf{0.05} & -0.38 & \BestPerf{0.81} \\
\quad Epigenetic marks prediction & MCC$\uparrow$ & -0.02 & \SecBestPerf{0.12} & -0.04 & -0.13 & \BestPerf{0.18} \\
\quad TF-m & MCC$\uparrow$ & \SecBestPerf{0.28} & 0.22 & 0.06 & 0.04 & \BestPerf{0.64} \\
\quad Enhancer-promoter interaction & MCC$\uparrow$ & \SecBestPerf{0.07} & -0.03 & 0.00 & 0.00 & \BestPerf{0.24} \\
\quad PD-prom 300 all & MCC$\uparrow$ & \SecBestPerf{0.30} & -0.10 & 0.23 & -0.07 & \BestPerf{0.89} \\
\quad PD-prom 300 notata & MCC$\uparrow$ & \SecBestPerf{0.23} & -0.08 & -0.04 & 0.13 & \BestPerf{0.93} \\
\quad PD-prom 300 tata & MCC$\uparrow$ & \SecBestPerf{0.30} & 0.00 & 0.05 & 0.22 & \BestPerf{0.55} \\
\quad CPD-prom core all & MCC$\uparrow$ & \SecBestPerf{0.35} & 0.05 & 0.02 & -0.13 & \BestPerf{0.68} \\
\quad CPD-prom core notata & MCC$\uparrow$ & \SecBestPerf{0.34} & 0.21 & 0.05 & -0.14 & \BestPerf{0.68} \\
\quad CPD-prom core tata & MCC$\uparrow$ & \SecBestPerf{0.40} & 0.09 & 0.06 & 0.02 & \BestPerf{0.67} \\
\quad TF-h & MCC$\uparrow$ & 0.21 & \SecBestPerf{0.22} & 0.14 & -0.02 & \BestPerf{0.52} \\
\quad Yeast PPI & ACC$\uparrow$ & 0.42 & \BestPerf{0.56} & 0.48 & 0.48 & \SecBestPerf{0.54} \\
\quad Human PPI & ACC$\uparrow$ & 0.70 & \BestPerf{0.82} & 0.50 & 0.55 & \SecBestPerf{0.73} \\
\quad Protein function & ROUGE-L$\uparrow$ & \SecBestPerf{0.43} & 0.10 & 0.14 & 0.14 & \BestPerf{0.51} \\
\quad Domain motif & ROUGE-L$\uparrow$ & \SecBestPerf{0.42} & 0.06 & 0.24 & 0.15 & \BestPerf{0.55} \\
\quad Non-coding RNA family & ACC$\uparrow$ & 0.21 & \SecBestPerf{0.24} & 0.16 & 0.12 & \BestPerf{0.90} \\
\quad Modification & ACC$\uparrow$ & 0.52 & \BestPerf{0.55} & 0.52 & \SecBestPerf{0.53} & 0.51 \\
\quad Fold type & ACC$\uparrow$ & \SecBestPerf{0.00} & \SecBestPerf{0.00} & \SecBestPerf{0.00} & \SecBestPerf{0.00} & \BestPerf{0.49} \\
\quad Subcellular localization & ACC$\uparrow$ & 0.48 & \SecBestPerf{0.65} & 0.11 & 0.13 & \BestPerf{0.88} \\
\quad EC number & Fmax$\uparrow$ & 0.09 & 0.12 & \SecBestPerf{0.15} & 0.13 & \BestPerf{0.78} \\
\quad Keywords & F1$\uparrow$ & \SecBestPerf{0.58} & 0.36 & 0.03 & 0.08 & \BestPerf{0.83} \\
\quad Metal ion binding & ACC$\uparrow$ & \BestPerf{0.74} & \SecBestPerf{0.70} & 0.43 & 0.60 & \BestPerf{0.74} \\
\quad GO-BP & Fmax$\uparrow$ & \SecBestPerf{0.43} & 0.29 & 0.27 & 0.31 & \BestPerf{0.52} \\
\quad GO-CC & Fmax$\uparrow$ & 0.33 & \SecBestPerf{0.40} & 0.28 & 0.35 & \BestPerf{0.58} \\
\quad GO-MF & Fmax$\uparrow$ & \SecBestPerf{0.47} & 0.24 & 0.16 & 0.40 & \BestPerf{0.66} \\
\midrule
\multicolumn{7}{l}{\textbf{Generation and Design}}\\
\midrule
\quad Function-guided protein design & Normalized SW$\uparrow$ & \SecBestPerf{0.74} & 0.67 & 0.73 & 0.73 & \BestPerf{0.94} \\
\quad Catalytic activity & ROUGE-L$\uparrow$ & \SecBestPerf{0.60} & 0.14 & 0.17 & 0.19 & \BestPerf{0.70} \\
\bottomrule
\end{tabular}
\caption{Evaluation results on Biology tasks grouped by task type. \BestPerf{Bold} indicates the best performance, and \SecBestPerf{underline} indicates the second best. 
}
\label{tab:res0624_biology_grouped}
\end{table}

\clearpage

\section{Human Evaluation Form}\label{secC1}

This appendix shows representative questionnaire items used for double-blinded human evaluation.
One sample is selected from each task category: crystal-material property
prediction, Gene Ontology prediction, and retrosynthesis. For each item,
evaluators read the input prompt, two anonymized model reasoning traces (named Model A and Model B correspondingly) and final
outputs, and a read-only ground-truth fact sheet. They then score each model on
five trace-quality axes and complete the overall comparison questions.

\subsection{General Evaluation Instructions}

Evaluators should assess the quality of the reasoning trace, not only whether
the final answer is numerically or symbolically close to the ground truth. The
main criteria are evidence grounding, domain plausibility, alignment with the
ground-truth regime or reaction/function region, reasoning coherence, and
hallucination risk.

\begin{longtable}{P{0.16\linewidth}P{0.70\linewidth}}
\toprule
\textbf{Verdict label} & \textbf{Meaning} \\
\midrule
Correct & The claim is supported by the provided input, the ground truth, the read-only Part~A fact sheet, or standard domain knowledge used only for verification. \\
Minor & A local or peripheral defect. The relevant entity or direction is mostly correct, and the defect does not materially change the quality of that scoring axis. \\
Major & A substantive axis-level defect, such as an input misread treated as fact, a domain inference contradicting the ground truth or standard knowledge, a wrong target region or reaction family, or a broken evidence-to-conclusion chain. Independent major defects on the same axis are counted separately. \\
Critical & Fabrication: the trace names a concrete entity, index, reference, topology, structure, GO term, reaction, reagent, or mechanism that is absent from the input/ground truth and cannot be verified from standard sources. Real-but-misapplied entities are not Critical; they are routed to Q2 or Q3. \\
\bottomrule
\end{longtable}

\begin{longtable}{P{0.10\linewidth}P{0.275\linewidth}P{0.48\linewidth}}
\toprule
\textbf{Axis} & \textbf{Focus} & \textbf{Scoring rule} \\
\midrule
Q1 & Evidence grounding in the provided input & Check whether entities, tokens, sequence positions, atom-map indices, product groups, or cited input facts actually appear in the task input. Score in \(\{1,\ldots,10\}\) or N.A. using the count-to-score table. \\
Q2 & Domain plausibility relative to ground truth & Check whether the trace's scientific inferences are plausible given the ground truth, and standard domain knowledge. Score in \(\{1,\ldots,10\}\) or N.A. using the count-to-score table. \\
Q3 & Target alignment & Judge whether the committed conclusion falls in the correct materials regime, GO region, or retrosynthesis reaction class/formed bond. Score in \(\{1,\ldots,10\}\) or N.A. using the count-to-score table. \\
Q4 & Reasoning coherence & Check whether the trace builds a relevant evidence-to-conclusion chain without unresolved contradiction, circularity, padding, or unused observations. Score in \(\{1,\ldots,10\}\) or N.A. using the count-to-score table. \\
Q5 & Unsupported overclaiming or hallucination & Check for fabricated or over-specific unsupported claims. Score in \(\{1,\ldots,10\}\) or N.A. using the count-to-score table. \\
\bottomrule
\end{longtable}

\begin{longtable}{P{0.68\linewidth}P{0.18\linewidth}}
\toprule
\textbf{Verdict counts on Q1/Q2/Q3/Q4/Q5} & \textbf{Score} \\
\midrule
\(Critical \ge 2\), or the core conclusion depends on a fabricated entity & 1 \\
\(Critical=1, Major\ge 1\) & 2 \\
\(Critical=1, Major=0\) & 3 \\
\(Critical=0, Major\ge 4\) & 2 \\
\(Critical=0, Major=3\) & 3 \\
\(Critical=0, Major=2\) & 4 \\
\(Critical=0, Major=1\) & 5 \\
\(Critical=0, Major=0, Minor\ge 4\) & 6 \\
\(Critical=0, Major=0, Minor=3\) & 7 \\
\(Critical=0, Major=0, Minor=2\) & 8 \\
\(Critical=0, Major=0, Minor=1\) & 9 \\
\(Critical=0, Major=0, Minor=0\), and at least one claim is verified Correct & 10 \\
No checkable claim on the axis & N.A. \\
\bottomrule
\end{longtable}

\begin{longtable}{P{0.10\linewidth}P{0.25\linewidth}P{0.25\linewidth}P{0.25\linewidth}}
\toprule
\textbf{Axis} & \textbf{Materials} & \textbf{Gene Ontology} & \textbf{Retrosynthesis} \\
\midrule
Q1 & Evidence grounding in formula, SLICES-PLUS space-group tokens, atom lists, edge lists, periodic offsets, and coordination counts. &
Evidence grounding in amino-acid sequence length, residue positions, motifs, and the Foldseek 3Di structural alphabet. &
Evidence grounding in product SMILES, product functional groups, atom-map indices, product connectivity, and cited atom-map sets. \\
Q2 & Materials plausibility relative to crystal chemistry, property constraints, and the ground-truth property regime. &
Biological plausibility relative to the ground-truth protein identity, family, motifs, domains, and GO region. &
Disconnection and reactant plausibility relative to the gold route, atom-map balance, oxidation/protection state, and chemical feasibility. \\
Q3 & Property-regime alignment, including boundary handling for values near regime edges. &
GO-space alignment within the evaluated branch, including correct, adjacent, wrong-region, wrong-super-class, and no-commit cases. &
Reaction-class alignment: gold reaction family plus gold formed bond scores highest; neighbouring families sharing the formed bond are borderline; different valid disconnections are lower. \\
Q4 & Coherence from decoded structure and target-specific mechanism to the committed property call. &
Coherence from sequence, 3Di, motif, domain, or family evidence to the committed GO-function hypothesis. &
Coherence from product parsing to retrosynthetic disconnection and reactant proposal. \\
Q5 & Unsupported materials overclaiming, such as invented topology, SBU, phase transition, atom index, or citation. &
Unsupported biological overclaiming, such as invented residues, motifs, domains, GO terms, protein identity, or citation. &
Unsupported chemical overclaiming, such as invented product groups, atom-map indices, reagents, named reactions, mechanisms, or citations. \\
\bottomrule
\end{longtable}

The questionnaire also contains four overall fields. Q6 asks how Model A compares
with expert expectation, Q7 asks the same for Model B, Q8 asks for a direct
Model A versus Model B comparison, and Q9 records evaluator confidence on a
1--10 scale with a free-text note.

\subsection{Blank Scoring Sheet Used for Each Sample}

\begin{longtable}{P{0.12\linewidth}P{0.18\linewidth}P{0.18\linewidth}P{0.44\linewidth}}
\toprule
\textbf{Axis} & \textbf{Model A score} & \textbf{Model B score} & \textbf{Evidence, claim verdicts, and notes} \\
\midrule
Q1 & \scoreblank & \scoreblank & \\
Q2 & \scoreblank & \scoreblank & \\
Q3 & \scoreblank & \scoreblank & \\
Q4 & \scoreblank & \scoreblank & \\
Q5 & \scoreblank & \scoreblank & \\
\midrule
Sum & \scoreblank & \scoreblank & \\
Mean & \scoreblank & \scoreblank & \\
\bottomrule
\end{longtable}

\begin{longtable}{P{0.26\linewidth}P{0.66\linewidth}}
\toprule
\textbf{Overall question} & \textbf{Allowed response} \\
\midrule
Q6. Model A versus expert expectation &
Significantly falls short / Falls short / Comparable / Exceeds / Significantly exceeds. \\
Q7. Model B versus expert expectation &
Significantly falls short / Falls short / Comparable / Exceeds / Significantly exceeds. \\
Q8. Direct comparison &
A much better / A slightly better / Tie / B slightly better / B much better. \\
Q9. Evaluator confidence &
Integer confidence score from 1 to 10, plus a free-text confidence note. \\
\bottomrule
\end{longtable}

\sampletitle{Materials}{Ag2HgI4, shear modulus}

\begin{longtable}{P{0.28\linewidth}P{0.64\linewidth}}
\toprule
\textbf{Field} & \textbf{Value} \\
\midrule
Dataset / task & JARVIS-DFT / \(G_v\). \\
Sample ID & 1169. \\
Chemical formula & \texttt{Ag2HgI4}. \\
Target property & \texttt{shear\_modulus\_gv}. \\
Property description & Shear modulus: resistance to shear deformation; related to directional bonding, framework rigidity, and elastic anisotropy. \\
Ground truth & \(5.77\) GPa. \\
Model A final output & \texttt{\{shear\_modulus\_gv: 5.62\}}. \\
Model B final output & \texttt{\{shear\_modulus\_gv: 8.00\}}. \\
\bottomrule
\end{longtable}

\paragraph{Input prompt.}
\begin{PromptBlock}
You are a material scientist. Look at the chemical formula and structure information of the given crystalline material and predict its property. The output must be in a JSON format. For example: {property_name: predicted_property_value}. Answer as precise as possible and in as few words as possible.
chemical formula: Ag2HgI4
structure information: <material_structure>o w b OOO m OOO c OOO o Ag Ag Hg I I I I 0 6 -oo 0 3 -oo 0 5 ooo 0 4 ooo 1 4 o-o 1 5 ooo 1 3 o-o 1 6 ooo 2 3 --o 2 6 -o- 2 4 o-- 2 5 ooo</material_structure>
property name: shear_modulus_gv
\end{PromptBlock}

\paragraph{Read-only ground-truth fact sheet.}
\begin{longtable}{P{0.10\linewidth}P{0.28\linewidth}P{0.54\linewidth}}
\toprule
\textbf{ID} & \textbf{Reference fact} & \textbf{Value} \\
\midrule
A1 & Decoded space group & No primer match. The leading tokens \texttt{o w b OOO m OOO c OOO o} do not exactly match any listed SLICES-PLUS primer row, so a concrete space-group label is not a checkable Q1 claim. \\
A2 & Number of atoms & 7; any cited atom index \(\geq 7\) is fabricated. \\
A3 & Formula consistency & The atom list Ag Ag Hg I I I I corresponds to 2 Ag, 1 Hg, and 4 I, matching \texttt{Ag2HgI4}. \\
A4 & Target and ground truth & \texttt{shear\_modulus\_gv} \(= 5.77\) GPa. \\
A5 & Ground-truth regime & Soft shear modulus, \(G_v < 20\) GPa; far from the 20 GPa boundary. The hard physical constraint is \(G_v \lesssim K_v\). \\
A6 & Atom-index table & 0=Ag, 1=Ag, 2=Hg, 3=I, 4=I, 5=I, 6=I. \\
A7 & Key neighbors & Ag(0) connects to I(6,3,5,4); Ag(1) connects to I(4,5,3,6); Hg(2) connects to I(3,6,4,5). Each iodine connects to Ag(0), Ag(1), and Hg(2). \\
A8 & Coordination summary & Ag(0), Ag(1), and Hg(2) are each four-coordinate, tetrahedral by iodine; each iodine is three-coordinate. \\
\bottomrule
\end{longtable}

\paragraph{Example claim prompts shown to the evaluator.}
\begin{longtable}{P{0.12\linewidth}P{0.38\linewidth}P{0.38\linewidth}}
\toprule
\textbf{Axis} & \textbf{Model A claim examples} & \textbf{Model B claim examples} \\
\midrule
Q1 & The trace cites the atom list, selected edge tokens, and tetrahedral metal-iodine coordination. &
The trace parses indices 0 and 1 as Ag, 2 as Hg, and 3--6 as I, and lists the twelve metal-iodine edges. \\
Q2 & Heavy and polarizable iodide ions imply a compliant lattice with low shear stiffness. &
The trace treats the material as a soft iodide solid and invokes tetrahedral coordination. \\
Q3 & The committed value 5.62 GPa lies in the correct soft regime and is close to the ground truth. &
The committed value 8.00 GPa also lies in the correct soft regime. \\
Q4 & The trace connects structure, coordination, heavy halide chemistry, and the target mechanical property. &
The trace contains longer speculative discussion; evaluators check whether the final estimate follows from the cited evidence. \\
Q5 & No concrete fabricated entity is apparent if the trace remains within decoded structure and standard halide chemistry. &
Evaluators check any named phase, database identifier, or literature-like statement for fabrication. \\
\bottomrule
\end{longtable}

\sampletitle{Gene Ontology}{1bd8\_A-P55273, biological process}

\begin{longtable}{P{0.28\linewidth}P{0.64\linewidth}}
\toprule
\textbf{Field} & \textbf{Value} \\
\midrule
Dataset / task & GO / BP. \\
Protein name in sample & \texttt{1bd8\_A-P55273}. \\
Chain & A. \\
Model A performance metadata & F1 \(=0.967\), precision \(=0.993\), recall \(=0.943\); 139 predicted BP terms. \\
Model B performance metadata & F1 \(=0.209\), precision \(=0.314\), recall \(=0.156\); 70 predicted BP terms. \\
Number of true BP terms & 145. \\
\bottomrule
\end{longtable}

\paragraph{Input prompt.}
\begin{PromptBlock}
What are the Gene Ontology terms (biological process) for this protein?
<protein>RAGDRLSGAAARGDVQEVRRLLHRELVHPDALNRFGKTALQVMMFGSTAIALELLKQGASPNVQDTSGTSPVHDAARTGFLDTLKVLVEHGADV NVPDGTGALPIHLAVQEGHTAVVSFLAAESDLHRRDARGLTPLELALQRGAQDLVDILQGHM</protein>
<protein_structure>D L Q Q Q L L V C Q L V L N L P S N C C S C P V V V D Q Q Q D A D P V R G G S L A R G P L L N L S S L L S S L V S P N D L C D A D P Q Q D G S L L V N L L V L V V S N N V S S V V S P H D Q A R A G N Q Q D G S L L N N L Q N V N L V S N L V C L Q P Y P Q C G A G N V R D G S L R S N V V V V V V S S N V S N V V S D</protein_structure>
Only the biological process (BP) aspect.
\end{PromptBlock}

\paragraph{Model outputs shown in the questionnaire.}
\begin{longtable}{P{0.20\linewidth}P{0.72\linewidth}}
\toprule
\textbf{Output} & \textbf{Abbreviated display} \\
\midrule
Model A final prediction & A long semicolon-separated list of 139 BP terms, including \texttt{regulation of signal transduction}, \texttt{regulation of cell cycle}, \texttt{negative regulation of endopeptidase activity}, \texttt{proteolysis}, \texttt{regulation of apoptotic process}, \texttt{DNA damage response}, \texttt{regulation of kinase activity}, and \texttt{negative regulation of cell cycle}. \\
Model B final prediction & A JSON-style output with 70 BP terms, dominated by \texttt{regulation of DNA-templated transcription}, \texttt{gene expression}, \texttt{positive regulation of RNA biosynthetic process}, \texttt{positive regulation of DNA-templated transcription}, and related transcriptional or signaling terms. \\
Ground truth & A long semicolon-separated list of 145 BP terms centered on negative regulation of cell cycle, G1/S transition, CDK activity regulation, apoptotic or programmed cell-death regulation, DNA-damage response and repair, and stress-response processes. \\
\bottomrule
\end{longtable}

\paragraph{Read-only ground-truth fact sheet.}
\begin{longtable}{P{0.10\linewidth}P{0.28\linewidth}P{0.54\linewidth}}
\toprule
\textbf{ID} & \textbf{Reference fact} & \textbf{Value} \\
\midrule
A1 & Sequence length & 156 residues. \\
A2 & 3Di length and low-confidence stretches & 156 tokens, aligned 1:1 with the amino-acid sequence; no \texttt{\#} low-confidence tokens are present. \\
A3 & Diagnostic motifs & Basic-rich segments \texttt{RRLLHRE} at residue 19 and \texttt{RRDARGL} at residue 128. The sequence is built from tandem ankyrin repeats; it does not contain a real DNA-binding or bZIP motif. \\
A4 & Ground-truth identity / family & Cyclin-dependent kinase 4 inhibitor D (p19INK4d / CDKN2D), human. It is an ankyrin-repeat CDK inhibitor and tumor suppressor. \\
A5 & Ground-truth BP region & Negative regulation of cell cycle and cell-cycle phase transition, especially G1/S; regulation of CDK or protein-serine/threonine kinase activity; apoptotic or programmed cell-death regulation; DNA-damage response and repair; response to stress, radiation, or chemical stimulus. Not transcription or DNA-templated gene expression. \\
A6 & Number of true terms & \(N_{\mathrm{true}} = 145\). \\
A7 & UniProt / InterPro constraints & InterPro IPR050776 Ank\_Repeat/CDKN\_Inhibitor, IPR002110 Ankyrin\_rpt, and IPR036770 Ankyrin\_rpt-contain\_sf. Keywords include Cell cycle, Tumor suppressor, ANK repeat, Nucleus, and Cytoplasm. The protein inhibits CDK4 and CDK6. \\
\bottomrule
\end{longtable}

\paragraph{Example claim prompts shown to the evaluator.}
\begin{longtable}{P{0.12\linewidth}P{0.38\linewidth}P{0.38\linewidth}}
\toprule
\textbf{Axis} & \textbf{Model A claim examples} & \textbf{Model B claim examples} \\
\midrule
Q1 & The trace cites 3Di runs, loop-like 3Di segments, and basic sequence clusters such as \texttt{RRLLHRE}. &
The trace quotes the full sequence and identifies a basic region, but also claims a heptad or leucine-zipper-like pattern. \\
Q2 & The trace infers a regulatory protein and predicts cell-cycle, apoptosis, kinase-regulation, and stress-response terms. &
The trace identifies the protein as a bZIP, WRKY, or transcription factor-like protein and predicts transcriptional regulation terms. \\
Q3 & Model A's committed terms largely overlap the CDK inhibitor BP region. &
Model B's committed terms center on transcription and gene expression, which is outside the ground-truth BP region. \\
Q4 & Evaluators check whether the trace moves from sequence and 3Di evidence to the committed GO region without unsupported leaps. &
Evaluators check format-oriented reasoning, identity contradictions, and whether the final term list follows from grounded evidence. \\
Q5 & Named biological entities and GO terms should be checked for unsupported specificity or fabrication. &
The bZIP, WRKY, leucine-zipper, and transcription-factor claims must be judged as real-but-misassigned or fabricated, following the rubric. \\
\bottomrule
\end{longtable}

\sampletitle{Retrosynthesis}{USPTO-50K sample 4, other}

\begin{longtable}{P{0.28\linewidth}P{0.64\linewidth}}
\toprule
\textbf{Field} & \textbf{Value} \\
\midrule
Dataset / task & USPTO-50K / retrosynthesis. \\
Sample ID & 4. \\
Reaction class & Other / uncategorized retrosynthesis reaction. \\
Matches gold & Model A matches the gold reactants; Model B proposes a related but not gold reactant set. \\
\bottomrule
\end{longtable}

\paragraph{Input prompt.}
\begin{PromptBlock}
Please suggest potential reactants for the given product.
<SMILES> [C:1](=[O:2])([C:3]([F:4])([F:5])[F:6])[NH:7][CH2:8][c:9]1[cH:10][cH:11] 
[cH:12][cH:13][c:14]1[S:15](=[O:16])(=[O:17])[CH:18]1[CH2:19][CH2:20]1 </SMILES>
\end{PromptBlock}

\paragraph{Model outputs and gold reactants.}
\begin{longtable}{P{0.20\linewidth}P{0.72\linewidth}}
\toprule
\textbf{Field} & \textbf{SMILES} \\
\midrule
Product & \smiles{[C:1](=[O:2])([C:3]([F:4])([F:5])[F:6])[NH:7][CH2:8][c:9]1[cH:10][cH:11][cH:12][cH:13][c:14]1[S:15](=[O:16])(=[O:17])[CH:18]1[CH2:19][CH2:20]1}. \\
Gold reactants & \smiles{O=C(O[C:1](=[O:2])[C:3]([F:4])([F:5])[F:6])C(F)(F)F.[NH2:7][CH2:8][c:9]1[cH:10][cH:11][cH:12][cH:13][c:14]1[S:15](=[O:16])(=[O:17])[CH:18]1[CH2:19][CH2:20]1}. \\
Model A final prediction & \smiles{O=C(O[C:1](=[O:2])[C:3]([F:4])([F:5])[F:6])C(F)(F)F.[NH2:7][CH2:8][c:9]1[cH:10][cH:11][cH:12][cH:13][c:14]1[S:15](=[O:16])(=[O:17])[CH:18]1[CH2:19][CH2:20]1}. \\
Model B final prediction & \smiles{NCc1ccccc1S(=O)(=O)C1CC1.OC(=O)C(F)(F)F}. \\
\bottomrule
\end{longtable}

\paragraph{Read-only ground-truth fact sheet.}
\begin{longtable}{P{0.10\linewidth}P{0.28\linewidth}P{0.54\linewidth}}
\toprule
\textbf{ID} & \textbf{Reference fact} & \textbf{Value} \\
\midrule
A1 & Product functional-group inventory & Trifluoroacetyl \(\mathrm{C(=O)CF_3}\) at atom maps 1--6; secondary amide N-H at atom 7; benzylic \(\mathrm{CH_2}\) at atom 8; ortho-disubstituted benzene ring at atoms 9--14; aryl sulfone at atom 15 with oxygens 16 and 17; cyclopropyl atoms 18--20. \\
A2 & Highest atom-map index & 20; any cited atom-map index \(>20\) is fabricated. \\
A3 & Gold reaction class and formed bond & Trifluoroacetylation of a primary amine. The forward reaction forms the amide C--N bond between carbonyl carbon [C:1] and nitrogen [NH:7]. The acyl source is trifluoroacetic anhydride (TFAA). \\
A4 & Gold reactant atom-map set and legitimate lost atoms & The TFAA fragment carries atoms 1--6, while the amine fragment carries atoms 7--20. The union covers the full product atom map. The unmapped leaving group is a trifluoroacetate or water-equivalent by-product. \\
\bottomrule
\end{longtable}

\paragraph{Example claim prompts shown to the evaluator.}
\begin{longtable}{P{0.12\linewidth}P{0.38\linewidth}P{0.38\linewidth}}
\toprule
\textbf{Axis} & \textbf{Model A claim examples} & \textbf{Model B claim examples} \\
\midrule
Q1 & The trace identifies the trifluoroacetyl group, amide N, benzyl group, sulfone, and cyclopropyl group. &
The trace parses the product as a trifluoroacetamide attached to an ortho-substituted aryl sulfone and cyclopropyl group. \\
Q2 & The trace disconnects the C1--N7 amide bond and proposes TFAA plus the primary amine. &
The trace disconnects the same amide bond but proposes trifluoroacetic acid plus the amine. \\
Q3 & Model A matches the gold amide-forming reaction family and gold formed bond. &
Model B finds the same C--N disconnection but uses a less activated acyl source than the gold TFAA route. \\
Q4 & The trace follows product parsing to amide disconnection to reactant selection. &
Evaluators check for length padding, repeated route restatement, and wobble among acid, acid chloride, and anhydride before commitment. \\
Q5 & TFAA, primary amine acylation, and amide formation are real and supported by the product. &
Trifluoroacetic acid and the amine are real; evaluators check whether any unsupported reaction claims or invented reagents appear. \\
\bottomrule
\end{longtable}

\begin{fullcotsection}{inputbg}{Input prompt}
\begin{cotverbatim}
Chemical formula: BaLaFe2Cu2O11

Target property: band_gap

Material structure:
<material_structure>o x b OOO j OOO o Ba La Fe Fe Cu Cu O O O O O O O O O O O 0 8 -oo 0 12 -oo 0 16 -oo 0 16 ooo 0 7 -oo 0 11 -oo 0 14 ooo 0 14 o+o 0 6 ooo 0 10 ooo 0 9 ooo 0 13 ooo 1 9 o-o 1 13 o-o 1 14 ooo 1 14 +oo 1 6 ooo 1 10 ooo 1 16 o-o 1 16 ooo 1 7 o-o 1 11 o-o 1 8 ooo 1 12 ooo 2 7 --o 2 8 -oo 2 9 o-o 2 6 ooo 2 14 ooo 3 10 oo- 3 13 oo- 3 12 oo- 3 11 oo- 3 15 ooo 3 16 oo- 4 11 --- 4 12 -o- 4 13 o-- 4 10 oo- 4 14 oo- 5 6 ooo 5 9 ooo 5 8 ooo 5 15 ooo 5 16 ooo 5 7 ooo 6 14 ooo 6 16 ooo 7 14 ++o 7 16 ooo 8 14 +oo 8 16 ooo 9 14 o+o 9 16 ooo 10 14 ooo 10 16 ooo 11 14 ++o 11 16 ooo 12 14 +oo 12 16 ooo 13 14 o+o 13 16 ooo</material_structure> \\

Question:
You are a material scientist. Look at the chemical formula and structure information of the given crystalline material and predict its property. The output must be in a json format. For example: {property_name: predicted_property_value}. Answer as precise as possible and in as few words as possible. \\
chemical formula: BaLaFe2Cu2O11 \\
structure information: <material_structure>o x ...</material_structure> \\
property name: band_gap
\end{cotverbatim}
\end{fullcotsection}

\vspace{0.6em}

\begin{fullcotsection}{cotbg}{SciReasoner reasoning trace}
\begin{cotverbatim}
<think>
To determine the band gap of this material, I first need to decode the structural information provided in the SLICES string to understand the chemical environment and bonding topology. The input string begins with the space group encoding <material_structure>o x b OOO j OOO o</material_structure>, which defines the symmetry operations, followed immediately by the atom list. Reading the atom symbols in order, I see <material_structure>Ba La Fe Fe Cu Cu O O O O O O O O O O O</material_structure>. This confirms the stoichiometry matches the formula BaLaFe2Cu2O11, consisting of one Barium, one Lanthanum, two Iron, two Copper, and eleven Oxygen atoms per unit cell. \\
My primary focus for predicting the electronic property is the transition metal sublattice, specifically the Iron and Copper sites, which correspond to indices 2, 3, 4, and 5 in the atom list. I need to examine the bonding environment to see how these metals interact with the oxygen ligands. Scanning the edge list, I find connections such as <material_structure>2 6 ooo</material_structure> and <material_structure>2 14 ooo</material_structure>, which indicate that the Iron atom at index 2 is bonded to Oxygen atoms at indices 6 and 14 within the same unit cell. Similarly, the Copper atom at index 5 shows connections like <material_structure>5 15 ooo</material_structure> and <material_structure>5 16 ooo</material_structure>, linking it to Oxygen sites 15 and 16. \\
Crucially, I observe that the Oxygen sites are not isolated; they act as bridges connecting the different metal centers. For instance, Oxygen site 14 is connected to both Iron (index 2 via <material_structure>2 14 ooo</material_structure>) and Copper (index 1 via <material_structure>1 14 ooo</material_structure>), creating a Fe-O-Cu pathway. This heteroatomic networking allows for significant orbital overlap between the transition metal d-orbitals and the oxygen p-orbitals. In materials containing mixed transition metals like Iron and Copper, the presence of these continuous pathways often leads to broadened energy bands. If the Fermi level intersects these bands, the material exhibits metallic behavior. The dense connectivity evident in the edge list, with multiple intra-cell and inter-cell bonds like <material_structure>0 16 -oo</material_structure> and <material_structure>13 14 o+o</material_structure>, supports a scenario where charge carriers can delocalize across the lattice. Therefore, based on the structural evidence of a connected transition metal-oxygen framework, the most consistent prediction for the band gap is that it closes completely. \\
{band_gap: 0.0}
</think>
\end{cotverbatim}
\end{fullcotsection}

\vspace{0.6em}

\begin{fullcotsection}{finalbg}{SciReasoner final output}
\begin{cotverbatim}
{band_gap: 0.0}
\end{cotverbatim}
\end{fullcotsection}

\vspace{0.6em}

\captionof{figure}{
Full example of a structure-aware reasoning trace. The input contains the
chemical formula, target property, and SLICES-PLUS structural string. The
reasoning trace explicitly cites structural substrings from the input and
connects them to a material-property mechanism before producing the final
prediction.
}
\label{fig:full-cot-example}

\clearpage

\begin{fullcotsection}{inputbg}{Input prompt}

\begin{cotverbatim}

Protein name: 1t4b_A-P0A9Q9

Protein structure: 

<protein_structure>D F E E E E A C L V F P L N V V L L V V C L V V V V L V V Y N Y A Y E D C P Q F Q D F D D V S S V D R D T H H H L Q D L V V V L R G Q E Y A Y P N E D V S C V P R V V V N V V V V R L H A Y E E L Y P N C P P P P L E A E F D C L Q F V V V V V V S V V V S R R Y Y Y H A A L V L L L C C L Q Q S L C V V V Q F F Q A K E K D K A F A Q C S L G D V S L V L L V Q Q V V V L C V Q C V V L V V P P P R D V V V S V V S S V C C Q P V P P R D C V R V V H D A H V D K D F A D A D A D P V R H H P V Q V S Q Q V N N C R S N V D P D G H Y Y G Y I D M Y H S A H F K M K M K M W T F T P D Q D D L V V S Q V S S C V S D V Q N A E Q E Q D D P S C V Q C A D S V N A E P H S H W Y K Y P W D A D P V D R R I T I I M I M H G C S D S V G N P R S S V V V V V V D</protein_structure>

Question: 
Given the protein sequence <protein>MKNVGFIGWRGMVGSVLMQRMVEERDFDAIRPVFFSTSQLGQAAPSFGGTTGTLQDAFDLEALKALDIIVTCQGGDYTNEIYPKLRESGWQGYWIDAASSLRMKDDAIIILDPVNQDVITDGLNNGIRTFVGGNCTVSLMLMSLGGLFANDLVDWVSVATYQAASGGGARHMRELLTQMGHLYGHVADELATPSSAILDIERKVTTLTRSGELPVDNFGVPLAGSLIPWIDKQLDNGQSREEWKGQAETNKILNTSSVIPVDGLCVRVGALRCHSQAFTIKLKKDVSIPTVEELLAAHNPWAKVVPNDREITMRELTPAAVTGTLTTPVGRLRKLNMGPEFLSAFTVGDQLLWGAAEPLRRMLRQLA</protein> and its structure representation <protein_structure>D F E ... </protein_structure>, predict its Gene Ontology annotations for biological process. Only the biological process (BP) aspect.

\end{cotverbatim}

\end{fullcotsection}

\par\medskip

\begin{fullcotsection}{cotbg}{SciReasoner reasoning trace}

\begin{cotverbatim}

<think>

To determine the appropriate Gene Ontology (GO) biological process annotations for this protein, I need to analyze the relationship between its amino acid sequence and its predicted three-dimensional structure. The core task is to infer the protein's function---specifically its role in cellular processes---based on the structural motifs and sequence signatures provided in the input. I will start by assessing the overall structural class defined by the 3Di representation, then identify specific sequence motifs that indicate enzymatic activity, and finally map these findings to the relevant biological processes.

First, I examine the structural representation to understand the protein's fold. The 3Di string provides a residue-level encoding of the local geometry. Scanning through the structure, I observe a significant prevalence of residues encoding alpha-helical conformations. For instance, the segment <protein_structure>V V L L V V C L V V V V L V V</protein_structure> near the N-terminal region shows a continuous run of V and L characters, which according to the decoding rules, signifies a stable alpha-helix. Similarly, towards the C-terminus, the pattern <protein_structure>V V V V V V D</protein_structure> indicates another helical segment. The presence of these helical bundles suggests a globular protein with a compact core, typical of enzymes that require a defined active site pocket. While there are regions with mixed characters like <protein_structure>D F E E E E A C L V F P L N</protein_structure>, which likely represent loop or coil regions connecting the secondary structure elements, the dominant feature is the alpha-helical architecture. This structural class is commonly associated with metabolic enzymes that bind small molecules.

Next, I turn to the amino acid sequence to identify specific functional motifs that correlate with the structural observations. The sequence contains several patterns indicative of cofactor binding or catalytic activity. Notably, there is a glycine-rich region <protein>GWQGYWIDAASSLRMKDDAIIILDPVNQDVITDGLNNGIRTFVGGNCTVSLMLMSLGGLFANDLVDWVSVATYQAASGGGARHMRELLTQMGHLYGHVADELATPSSAILDIERKVTTLTRSGELPVDNFGVPLAGSLIPWIDKQLDNGQSREEWKGQAETNKILNTSSVIPVDGLCVRVGALRCHSQAFTIKLKKDVSIPTVEELLAAHNPWAKVVPNDREITMRELTPAAVTGTLTTPVGRLRKLNMGPEFLSAFTVGDQLLWGAAEPLRRMLRQLA</protein>, specifically the <protein>VGGNCTVSLMLMSLGGLFANDLVDWVSVATYQAASGGGARHMRELLTQMGHLYGHVADELATPSSAILDIERKVTTLTRSGELPVDNFGVPLAGSLIPWIDKQLDNGQSREEWKGQAETNKILNTSSVIPVDGLCVRVGALRCHSQAFTIKLKKDVSIPTVEELLAAHNPWAKVVPNDREITMRELTPAAVTGTLTTPVGRLRKLNMGPEFLSAFTVGDQLLWGAAEPLRRMLRQLA</protein> stretch which includes conserved residues often found in nucleotide-binding domains. The presence of lysine and arginine residues, such as in the segment <protein>KDDAIIILDPVNQDVITDGLNNGIRTFVGGNCTVSLMLMSLGGLFANDLVDWVSVATYQAASGGGARHMRELLTQMGHLYGHVADELATPSSAILDIERKVTTLTRSGELPVDNFGVPLAGSLIPWIDKQLDNGQSREEWKGQAETNKILNTSSVIPVDGLCVRVGALRCHSQAFTIKLKKDVSIPTVEELLAAHNPWAKVVPNDREITMRELTPAAVTGTLTTPVGRLRKLNMGPEFLSAFTVGDQLLWGAAEPLRRMLRQLA</protein>, suggests potential interaction with phosphate groups, a hallmark of enzymes involved in phosphorylation or nucleotide metabolism. Furthermore, the sequence contains motifs like <protein>MGHLYGHVADELATPSSAILDIERKVTTLTRSGELPVDNFGVPLAGSLIPWIDKQLDNGQSREEWKGQAETNKILNTSSVIPVDGLCVRVGALRCHSQAFTIKLKKDVSIPTVEELLAAHNPWAKVVPNDREITMRELTPAAVTGTLTTPVGRLRKLNMGPEFLSAFTVGDQLLWGAAEPLRRMLRQLA</protein> which resemble those found in synthase or transferase families. The combination of the alpha-helical structural scaffold and these specific sequence motifs strongly points towards an enzyme that participates in the synthesis or modification of amino acids and nucleotides.

Based on this structural and sequence evidence, I can deduce the biological processes. The protein's ability to bind nucleotides and its enzymatic nature imply involvement in metabolic pathways. Specifically, the motifs suggest it may catalyze reactions in the biosynthesis of amino acids, particularly those involving sulfur or serine derivatives, and potentially interact with pyruvate or related intermediates. The structural stability indicates it functions within the cell cytoplasm or a specific compartment where these metabolites are abundant. Therefore, the annotations should reflect its role in metabolic processes, specifically focusing on the synthesis and regulation of amino acids and nucleobase compounds. I will verify this by checking for consistency with the observed structural constraints; the alpha-helical dominance supports a soluble enzyme function rather than a membrane transporter, and the specific motifs align with biosynthetic pathways rather than degradation.

</think>

\end{cotverbatim}

\end{fullcotsection}

\par\medskip

\begin{fullcotsection}{finalbg}{\projName{} final output}

\begin{cotverbatim}

amino acid metabolic process; carboxylic acid biosynthetic process; lysine metabolic process; obsolete diaminopimelate metabolic process; dicarboxylic acid metabolic process; sulfur compound metabolic process; dicarboxylic acid biosynthetic process; methionine biosynthetic process; diaminopimelate biosynthetic process; isoleucine biosynthetic process; sulfur compound biosynthetic process; threonine metabolic process; methionine metabolic process; small molecule biosynthetic process; lysine biosynthetic process; sulfur amino acid biosynthetic process; branched-chain amino acid metabolic process; alpha-amino acid metabolic process; alpha-amino acid biosynthetic process; sulfur amino acid metabolic process; isoleucine metabolic process; amino acid biosynthetic process; carboxylic acid metabolic process; lysine biosynthetic process via diaminopimelate; aspartate family amino acid biosynthetic process; L-methionine biosynthetic process; organic acid biosynthetic process; organic acid metabolic process; obsolete aspartate family amino acid metabolic process; branched-chain amino acid biosynthetic process; oxoacid metabolic process

\end{cotverbatim}

\end{fullcotsection}

\captionof{figure}{
Full example of a structure-aware reasoning trace for biological-process Gene Ontology prediction. The input contains the amino-acid sequence of \texttt{1t4b\_A-P0A9Q9} and its residue-level 3Di structural representation. The reasoning trace cites sequence and structural substrings and connects them to metabolic and amino-acid-biosynthetic processes before producing the final set of predicted BP annotations.
}

\label{fig:full-cot-1t4b-a-p0a9q9-bp}

\begingroup

\begin{fullcotsection}{inputbg}{Input prompt}

\begin{cotverbatim}

Reaction class: alkylation

Question:
Please suggest potential reactants for the given product.
<SMILES> [CH2:1]([CH2:2][CH2:3][C:4]([c:5]1[c:6]([CH3:7])[s:8][c:9]2[c:10]1[CH2:11][N:12]([C:13]([CH3:14])=[O:15])[CH2:16][CH2:17]2)=[O:18])[N:28]1[CH2:27][CH2:26][CH:25]([c:24]2[c:23]3[cH:22][cH:21][c:20]([F:19])[cH:34][c:33]3[o:32][n:31]2)[CH2:30][CH2:29]1 </SMILES>

\end{cotverbatim}

\end{fullcotsection}

\par\medskip

\begin{fullcotsection}{cotbg}{SciReasoner reasoning trace}

\begin{cotverbatim}

<think>

Let's break down the retrosynthetic analysis for the provided product.

1. Analysis of the Product Structure
The product is a complex molecule with two main structural domains connected by a propyl chain:
- Domain A (Left): A thienopyridine ring system substituted with an acetyl group and a methyl group. This domain is attached to the propyl chain via a carbonyl group (ketone).
- Domain B (Right): A piperidine ring substituted with a fluorinated benzisoxazole group. This domain is attached to the propyl chain via a nitrogen atom (tertiary amine).

The connection between these domains is a propyl linker: `[CH2:1]-[CH2:2]-[CH2:3]`.

2. Retrosynthetic Disconnection
The most strategic bond to disconnect is the C-N bond between the propyl chain and the piperidine nitrogen (bond between `[CH2:1]` and `[N:28]`).
- This is a classic alkylation disconnection.
- The propyl chain acts as the alkylating agent (electrophile).
- The piperidine nitrogen acts as the nucleophile.

3. Identifying the Reactants
Based on this disconnection, we can propose the two reactants:

*   Reactant 1 (The Alkylating Agent):
    The propyl chain must be attached to a good leaving group. In organic synthesis, **chloride** is a standard leaving group for such alkylations.
    - Structure: `Cl-CH2-CH2-CH2-Domain A`.
    - In SMILES: `Cl[CH2:1][CH2:2][CH2:3][C:4](=[O:18])[c:5]1[c:6]([CH3:7])[s:8][c:9]2[c:10]1[CH2:11][N:12]([C:13]([CH3:14])=[O:15])[CH2:16][CH2:17]2`.

*   Reactant 2 (The Nucleophile):
    The piperidine ring must have a free hydrogen on the nitrogen to act as a nucleophile.
    - Structure: `Domain B-H`.
    - In SMILES: `[F:19][c:20]1[cH:21][cH:22][c:23]2[c:24]([CH:25]3[CH2:26][CH2:27][NH:28][CH2:29][CH2:30]3)[n:31][o:32][c:33]2[cH:34]1`.

4. Sanity Check
- Chemical Feasibility: The alkylation of a secondary amine (piperidine) with a primary alkyl chloride is a high-yielding, standard reaction, typically performed in the presence of a base (like K2CO3) to scavenge the HCl byproduct.
- Selectivity: The piperidine nitrogen is the most nucleophilic site in Reactant 2. The amide nitrogen in the thienopyridine ring (Reactant 1) is much less nucleophilic due to delocalization and is not a competing site for alkylation under these conditions.
- Connectivity: The disconnection correctly places the propyl chain on the nitrogen, matching the product structure.

5. Final Answer
The possible reactants are the alkyl chloride derivative of the thienopyridine domain and the substituted piperidine.

<SMILES> Cl[CH2:1][CH2:2][CH2:3][C:4]([c:5]1[c:6]([CH3:7])[s:8][c:9]2[c:10]1[CH2:11][N:12]([C:13]([CH3:14])=[O:15])[CH2:16][CH2:17]2)=[O:18].[F:19][c:20]1[cH:21][cH:22][c:23]2[c:24]([CH:25]3[CH2:26][CH2:27][NH:28][CH2:29][CH2:30]3)[n:31][o:32][c:33]2[cH:34]1 </SMILES>

</think>

\end{cotverbatim}

\end{fullcotsection}

\par\medskip

\begin{fullcotsection}{finalbg}{SciReasoner final output}

\begin{cotverbatim}

<SMILES> Cl[CH2:1][CH2:2][CH2:3][C:4]([c:5]1[c:6]([CH3:7])[s:8][c:9]2[c:10]1[CH2:11][N:12]([C:13]([CH3:14])=[O:15])[CH2:16][CH2:17]2)=[O:18].[F:19][c:20]1[cH:21][cH:22][c:23]2[c:24]([CH:25]3[CH2:26][CH2:27][NH:28][CH2:29][CH2:30]3)[n:31][o:32][c:33]2[cH:34]1 </SMILES>

\end{cotverbatim}

\end{fullcotsection}

\captionof{figure}{
Full example of a retrosynthetic reasoning trace for an alkylation reaction from the USPTO-50K dataset. The input contains the atom-mapped SMILES representation of the target product. The reasoning trace identifies the bond between \texttt{[CH2:1]} and \texttt{[N:28]} as the strategic C--N disconnection, interprets the transformation as alkylation of a secondary piperidine amine with a primary alkyl chloride, and produces the corresponding reactant SMILES.
}

\label{fig:full-cot-sample-44-alkylation}

\endgroup

\end{appendices}

\clearpage

\bibliography{sn-bibliography}

@article{segler2018planning,
  title={Planning chemical syntheses with deep neural networks and symbolic AI},
  author={Segler, Marwin HS and Preuss, Mike and Waller, Mark P},
  journal={Nature},
  volume={555},
  number={7698},
  pages={604--610},
  year={2018},
  publisher={Nature Publishing Group UK London}
}

@article{huang2021therapeutics,
  title={Therapeutics Data Commons: Machine Learning Datasets and Tasks for Drug Discovery and Development},
  author={Huang, Kexin and Fu, Tianfan and Gao, Wenhao and Zhao, Yue and Roohani, Yusuf and Leskovec, Jure and Coley, Connor and Xiao, Cao and Sun, Jimeng and Zitnik, Marinka},
  journal={Advances in neural information processing systems},
  year={2021}
}

@article{anfinsen1973principles,
  title={Principles that govern the folding of protein chains},
  author={Anfinsen, Christian B},
  journal={Science},
  volume={181},
  number={4096},
  pages={223--230},
  year={1973},
  publisher={American Association for the Advancement of Science}
}

@article{berman2000pdb,
  author = {Berman, Helen M. and Westbrook, John and Feng, Zukang and Gilliland, Gary and Bhat, Thomas N. and Weissig, Helge and Shindyalov, Ilya N. and Bourne, Philip E.},
  title = {The Protein Data Bank},
  journal = {Nucleic Acids Research},
  year = {2000},
  volume = {28},
  number = {1},
  pages = {235--242},
  doi = {10.1093/nar/28.1.235}
}

@article{su2024saprot,
  author = {Su, Jin and Han, Chenchen and Zhou, Yuyang and Shan, Junjie and Zhou, Xibin and Yuan, Fajie},
  title = {SaProt: Protein Language Modeling with Structure-aware Vocabulary},
  booktitle = {Proceedings of the International Conference on Learning Representations (ICLR)},
  year = {2024}
}

@article{varadi2022alphafold,
  author = {Varadi, Mihaly and Anyango, Stephen and Deshpande, Mandar and Nair, Sreenath and Natassia, Cindy and Yordanova, Galabina and Yuan, David and Stroe, Oana and Wood, Gemma and Laydon, Agata and Zidek, Augustin and Green, Tim and Tunyasuvunakool, Kathryn and Petersen, Stig and Jumper, John and Clancy, Ellen and Green, Richard and Vora, Ankur and Lutfi, Mira and Figurnov, Michael and Cowie, Andrew and Hobbs, Nicole and Kohli, Pushmeet and Kleywegt, Gerard J. and Birney, Ewan and Hassabis, Demis and Velankar, Sameer},
  title = {AlphaFold Protein Structure Database: massively expanding the structural coverage of protein-sequence space with high-accuracy models},
  journal = {Nucleic Acids Research},
  year = {2022},
  volume = {50},
  number = {D1},
  pages = {D439--D444},
  doi = {10.1093/nar/gkab1061}
}

@misc{pubmed_central,
  title = {PubMed Central},
  organization = {National Library of Medicine (NIH), USA},
  year = {2026},
  howpublished = {\url{https://www.ncbi.nlm.nih.gov/pmc/}},
  note = {Accessed: 2026-06-21}
}

@article{kshirsagar2025rapid,
  title={Rapid and accurate prediction of protein homo-oligomer symmetry using Seq2Symm},
  author={Kshirsagar, Meghana and Meller, Artur and Humphreys, Ian R and Sledzieski, Samuel and Xu, Yixi and Dodhia, Rahul and Horvitz, Eric and Berger, Bonnie and Bowman, Gregory R and Ferres, Juan Lavista and others},
  journal={Nature Communications},
  volume={16},
  number={1},
  pages={2017},
  year={2025},
  publisher={Nature Publishing Group UK London}
}

@misc{pubmed,
  title = {PubMed},
  organization = {National Library of Medicine (NIH), USA},
  year = {2026},
  howpublished = {\url{https://pubmed.ncbi.nlm.nih.gov/}},
  note = {Accessed: 2026-06-21}
}

@article{dana2019sifts,
  author = {Dana, Jose M. and Gutmanas, Aleksandras and Tyagi, Nidhi and Qi, Guoying and O'Donovan, Claire and Martin, Maria and Velankar, Sameer},
  title = {SIFTS: updated Structure Integration with Function, Taxonomy and Sequences resource allows 40-fold increase in coverage of structure-based annotations for proteins},
  journal = {Nucleic Acids Research},
  year = {2019},
  volume = {47},
  number = {D1},
  pages = {D482--D489},
  doi = {10.1093/nar/gky1114}
}

@article{zhao2024efficient,
  title={Efficient retrosynthetic planning with MCTS exploration enhanced A* search},
  author={Zhao, Dengwei and Tu, Shikui and Xu, Lei},
  journal={Communications Chemistry},
  volume={7},
  number={1},
  pages={52},
  year={2024},
  publisher={Nature Publishing Group UK London}
}

@article{duek2021functionathon,
  title={Functionathon: a manual data mining workflow to generate functional hypotheses for uncharacterized human proteins and its application by undergraduate students},
  author={Duek, Paula and Mary, Camille and Zahn-Zabal, Monique and Bairoch, Amos and Lane, Lydie},
  journal={Database},
  volume={2021},
  pages={baab046},
  year={2021},
  publisher={Oxford University Press UK}
}

@article{strieth2024artificial,
  title={Artificial intelligence for retrosynthetic planning needs both data and expert knowledge},
  author={Strieth-Kalthoff, Felix and Szymkuc, Sara and Molga, Karol and Aspuru-Guzik, Alan and Glorius, Frank and Grzybowski, Bartosz A},
  journal={Journal of the American Chemical Society},
  volume={146},
  number={16},
  pages={11005--11017},
  year={2024},
  publisher={ACS Publications}
}

@article{aygun2026ai,
  title={An AI system to help scientists write expert-level empirical software},
  author={Ayg{\"u}n, Eser and Belyaeva, Anastasiya and Comanici, Gheorghe and Coram, Marc and Cui, Hao and Garrison, Jake and Johnston, Renee and Kast, Anton and McLean, Cory Y and Norgaard, Peter and others},
  journal={Nature},
  pages={1--3},
  year={2026},
  publisher={Nature Publishing Group UK London}
}

@article{gottweis2026accelerating,
  title={Accelerating scientific discovery with Co-Scientist},
  author={Gottweis, Juraj and Weng, Wei-Hung and Daryin, Alexander and Tu, Tao and Sirkovic, Petar and Myaskovsky, Artiom and Glowaty, Grzegorz and Weissenberger, Felix and Orlandi, Alessio and Popovici, Dan and others},
  journal={Nature},
  pages={1--3},
  year={2026},
  publisher={Nature Publishing Group UK London}
}

@article{ghareeb2026multi,
  title={A multi-agent system for automating scientific discovery},
  author={Ghareeb, Ali Essam and Chang, Benjamin and Mitchener, Ludovico and Yiu, Angela and Szostkiewicz, Caralyn J and Shved, Dmytro and Gyimesi, Gavin J and Laurent, Jon M and Wright, Samantha M and Razzak, Muhammed T and others},
  journal={Nature},
  pages={1--3},
  year={2026},
  publisher={Nature Publishing Group UK London}
}

@article{uniprot2023uniprot,
  title={UniProt: the universal protein knowledgebase in 2023},
  author = {The UniProt Consortium },
  journal={Nucleic acids research},
  volume={51},
  number={D1},
  pages={D523--D531},
  year={2023},
  publisher={Oxford University Press}
}

@article{niyongabo2025llm,
  title={LLM-Prop: predicting the properties of crystalline materials using large language models},
  author={Niyongabo Rubungo, Andre and Arnold, Craig and Rand, Barry P and Dieng, Adji Bousso},
  journal={npj Computational Materials},
  volume={11},
  number={1},
  pages={186},
  year={2025},
  publisher={Nature Publishing Group UK London}
}

@article{yang2025qwen3,
  title={Qwen3 technical report},
  author={Yang, An and Li, Anfeng and Yang, Baosong and Zhang, Beichen and Hui, Binyuan and Zheng, Bo and Yu, Bowen and Gao, Chang and Huang, Chengen and Lv, Chenxu and others},
  journal={arXiv preprint arXiv:2505.09388},
  year={2025}
}

@article{yu2026dapo,
  title={Dapo: An open-source llm reinforcement learning system at scale},
  author={Yu, Qiying and Zhang, Zheng and Zhu, Ruofei and Yuan, Yufeng and Zuo, Xiaochen and Yue, Yu and Dai, Weinan and Fan, Tiantian and Liu, Gaohong and Liu, Lingjun and others},
  journal={Advances in Neural Information Processing Systems},
  volume={38},
  pages={113222--113244},
  year={2026}
}

@article{xiao2023invertible,
  title={An invertible, invariant crystal representation for inverse design of solid-state materials using generative deep learning},
  author={Xiao, Hang and Li, Rong and Shi, Xiaoyang and Chen, Yan and Zhu, Liangliang and Chen, Xi and Wang, Lei},
  journal={Nature Communications},
  volume={14},
  number={1},
  pages={7027},
  year={2023},
  publisher={Nature Publishing Group UK London}
}

@article{van2024fast,
  title={Fast and accurate protein structure search with Foldseek},
  author={Van Kempen, Michel and Kim, Stephanie S and Tumescheit, Charlotte and Mirdita, Milot and Lee, Jeongjae and Gilchrist, Cameron LM and S{\"o}ding, Johannes and Steinegger, Martin},
  journal={Nature biotechnology},
  volume={42},
  number={2},
  pages={243--246},
  year={2024},
  publisher={Nature Publishing Group US New York}
}

@article{sennrich2016neural,
  title={Neural machine translation of rare words with subword units},
  author={Sennrich, Rico and Haddow, Barry and Birch, Alexandra},
  booktitle={Proceedings of the 54th annual meeting of the association for computational linguistics (volume 1: long papers)},
  pages={1715--1725},
  year={2016},
}

@article{gage1994new,
  title={A new algorithm for data compression},
  author={Gage, Philip},
  journal={The C Users Journal},
  volume={12},
  number={2},
  pages={23--38},
  year={1994},
  publisher={R \& D Publications, Inc. Lawrence, KS, USA}
}

@article{mirza2025superhuman,
  title = {Are large language models superhuman chemists?},
  author={Mirza, Adrian and Alampara, Nawaf and Kunchapu, Sreekanth and R{\'\i}os-Garc{\'\i}a, Marti{\~n}o and Emoekabu, Benedict and Krishnan, Aswanth and Gupta, Tanya and Schilling-Wilhelmi, Mara and Okereke, Macjonathan and Aneesh, Anagha and others},
  journal = {Nature Chemistry},
  year = {2025},
  month = {Jul},
  publisher = {Nature Publishing Group}
}

@article{achiam2023gpt,
  title={Gpt-4 technical report},
  author={Achiam, Josh and Adler, Steven and Agarwal, Sandhini and Ahmad, Lama and Akkaya, Ilge and Aleman, Florencia Leoni and Almeida, Diogo and Altenschmidt, Janko and Altman, Sam and Anadkat, Shyamal and others},
  journal={arXiv preprint arXiv:2303.08774},
  year={2023}
}

@article{boiko2023autonomous,
  title={Autonomous chemical research with large language models},
  author={Boiko, Daniil A and MacKnight, Robert and Kline, Ben and Gomes, Gabe},
  journal={Nature},
  volume={624},
  number={7992},
  pages={570--578},
  year={2023},
  publisher={Nature Publishing Group UK London}
}

@article{wang2023scientific,
  title={Scientific discovery in the age of artificial intelligence},
  author={Wang, Hanchen and Fu, Tianfan and Du, Yuanqi and Gao, Wenhao and Huang, Kexin and Liu, Ziming and Chandak, Payal and Liu, Shengchao and Van Katwyk, Peter and Deac, Andreea and others},
  journal={Nature},
  volume={620},
  number={7972},
  pages={47--60},
  year={2023},
  publisher={Nature Publishing Group UK London}
}

@article{abramson2024accurate,
  title={Accurate structure prediction of biomolecular interactions with AlphaFold 3},
  author={Abramson, Josh and Adler, Jonas and Dunger, Jack and Evans, Richard and Green, Tim and Pritzel, Alexander and Ronneberger, Olaf and Willmore, Lindsay and Ballard, Andrew J and Bambrick, Joshua and others},
  journal={Nature},
  volume={630},
  number={8016},
  pages={493--500},
  year={2024},
  publisher={Nature Publishing Group UK London}
}

@article{xie2018crystal,
  title={Crystal graph convolutional neural networks for an accurate and interpretable prediction of material properties},
  author={Xie, Tian and Grossman, Jeffrey C},
  journal={Physical review letters},
  volume={120},
  number={14},
  pages={145301},
  year={2018},
  publisher={APS}
}

@article{yang2019analyzing,
  title={Analyzing learned molecular representations for property prediction},
  author={Yang, Kevin and Swanson, Kyle and Jin, Wengong and Coley, Connor and Eiden, Philipp and Gao, Hua and Guzman-Perez, Angel and Hopper, Timothy and Kelley, Brian and Mathea, Miriam and others},
  journal={Journal of chemical information and modeling},
  volume={59},
  number={8},
  pages={3370--3388},
  year={2019},
  publisher={ACS Publications}
}

@article{merchant2023scaling,
  title={Scaling deep learning for materials discovery},
  author={Merchant, Amil and Batzner, Simon and Schoenholz, Samuel S and Aykol, Muratahan and Cheon, Gowoon and Cubuk, Ekin Dogus},
  journal={Nature},
  volume={624},
  number={7990},
  pages={80--85},
  year={2023},
  publisher={Nature Publishing Group UK London}
}

@article{gligorijevic2021deepfri,
  title={Structure-based protein function prediction using graph convolutional networks},
  author={Gligorijevi{\'c}, Vladimir and Renfrew, P Douglas and Kosciolek, Tomasz and Leman, Julia Koehler and Berenberg, Daniel and Vatanen, Tommi and Chandler, Chris and Taylor, Bryn C and Fisk, Ian M and Vlamakis, Hera and others},
  journal={Nature communications},
  volume={12},
  number={1},
  pages={3168},
  year={2021},
  publisher={Nature Publishing Group UK London}
}

@article{su2025saprothub,
  title={Democratizing protein language model training, sharing and collaboration},
  author={Su, Jin and Li, Zhikai and Tao, Tianli and Han, Chenchen and He, Yan and Dai, Fengyuan and Yuan, Qingyan and Gao, Yuan and Si, Tong and Zhang, Xuting and others},
  journal={Nature Biotechnology},
  pages={1--7},
  year={2025},
  publisher={Nature Publishing Group US New York}
}

@article{altschul1990blast,
  title={Basic local alignment search tool},
  author={Altschul, Stephen F and Gish, Warren and Miller, Webb and Myers, Eugene W and Lipman, David J},
  journal={Journal of molecular biology},
  volume={215},
  number={3},
  pages={403--410},
  year={1990},
  publisher={Elsevier}
}

@misc{openai2026gpt55,
  title        = {Introducing {GPT-5.5}},
  author       = {{OpenAI}},
  year         = {2026},
  month        = apr,
  day          = {23},
  howpublished = {\url{https://openai.com/index/introducing-gpt-5-5/}},
  note         = {Accessed: 2026-06-05}
}

@misc{kimik26,
  title        = {Kimi K2.6 Tech Blog: Advancing Open-Source Coding},
  author       = {{Kimi}},
  year         = {2026},
  month        = apr,
  day          = {23},
  howpublished = {\url{https://www.kimi.com/blog/kimi-k2-6}},
  note         = {Accessed: 2026-07-07}
}

@misc{opus47,
  title        = {System Card: Claude Opus 4.7},
  author       = {{Anthropic}},
  year         = {2026},
  month        = apr,
  day          = {23},
  howpublished = {\url{https://anthropic.com/claude-opus-4-7-system-card}},
}

@misc{deepseekai2026deepseekv4,
      title={DeepSeek-V4: Towards Highly Efficient Million-Token Context Intelligence},
      author={DeepSeek-AI},
      year={2026},
}

@article{
esm2,
author = {Zeming Lin  and Halil Akin  and Roshan Rao  and Brian Hie  and Zhongkai Zhu  and Wenting Lu  and Nikita Smetanin  and Robert Verkuil  and Ori Kabeli  and Yaniv Shmueli  and Allan dos Santos Costa  and Maryam Fazel-Zarandi  and Tom Sercu  and Salvatore Candido  and Alexander Rives },
title = {Evolutionary-scale prediction of atomic-level protein structure with a language model},
journal = {Science},
volume = {379},
number = {6637},
pages = {1123-1130},
year = {2023},
doi = {10.1126/science.ade2574},
URL = {https://www.science.org/doi/abs/10.1126/science.ade2574},
eprint = {https://www.science.org/doi/pdf/10.1126/science.ade2574}}

@article{coley2017computer,
  title={Computer-assisted retrosynthesis based on molecular similarity},
  author={Coley, Connor W and Rogers, Luke and Green, William H and Jensen, Klavs F},
  journal={ACS central science},
  volume={3},
  number={12},
  pages={1237--1245},
  year={2017},
  publisher={ACS Publications}
}

@article{segler2017neural,
  title={Neural-symbolic machine learning for retrosynthesis and reaction prediction},
  author={Segler, Marwin HS and Waller, Mark P},
  journal={Chemistry--A European Journal},
  volume={23},
  number={25},
  pages={5966--5971},
  year={2017},
  publisher={Wiley Online Library}
}

@article{dai2019retrosynthesis,
  title={Retrosynthesis prediction with conditional graph logic network},
  author={Dai, Hanjun and Li, Chengtao and Coley, Connor and Dai, Bo and Song, Le},
  journal={Advances in Neural Information Processing Systems},
  volume={32},
  year={2019}
}

@article{chen2021deep,
  title={Deep retrosynthetic reaction prediction using local reactivity and global attention},
  author={Chen, Shuan and Jung, Yousung},
  journal={JACS Au},
  volume={1},
  number={10},
  pages={1612--1620},
  year={2021},
  publisher={ACS Publications}
}

@article{yan2022retrocomposer,
  title={RetroComposer: composing templates for template-based retrosynthesis prediction},
  author={Yan, Chaochao and Zhao, Peilin and Lu, Chan and Yu, Yang and Huang, Junzhou},
  journal={Biomolecules},
  volume={12},
  number={9},
  pages={1325},
  year={2022},
  publisher={MDPI}
}

@article{shi2020graph,
  title={A graph to graphs framework for retrosynthesis prediction},
  author={Shi, Chence and Xu, Minkai and Guo, Hongyu and Zhang, Ming and Tang, Jian},
  booktitle={International conference on machine learning},
  pages={8818--8827},
  year={2020}
}

@article{yan2020retroxpert,
  title={Retroxpert: Decompose retrosynthesis prediction like a chemist},
  author={Yan, Chaochao and Ding, Qianggang and Zhao, Peilin and Zheng, Shuangjia and Yang, Jinyu and Yu, Yang and Huang, Junzhou},
  journal={Advances in Neural Information Processing Systems},
  volume={33},
  pages={11248--11258},
  year={2020}
}

@article{wang2021retroprime,
  title={Retroprime: A diverse, plausible and transformer-based method for single-step retrosynthesis predictions},
  author={Wang, Xiaorui and Li, Yuquan and Qiu, Jiezhong and Chen, Guangyong and Liu, Huanxiang and Liao, Benben and Hsieh, Chang-Yu and Yao, Xiaojun},
  journal={Chemical Engineering Journal},
  volume={420},
  pages={129845},
  year={2021},
  publisher={Elsevier}
}

@article{chen2023g,
  title={G 2 Retro as a two-step graph generative models for retrosynthesis prediction},
  author={Chen, Ziqi and Ayinde, Oluwatosin R and Fuchs, James R and Sun, Huan and Ning, Xia},
  journal={Communications Chemistry},
  volume={6},
  number={1},
  pages={102},
  year={2023},
  publisher={Nature Publishing Group UK London}
}

@article{gao2022semiretro,
  title={SemiRetro: Semi-template framework boosts deep retrosynthesis prediction},
  author={Gao, Zhangyang and Tan, Cheng and Wu, Lirong and Li, Stan Z},
  journal={arXiv preprint arXiv:2202.08205},
  year={2022}
}

@article{zhong2023retrosynthesis,
  title={Retrosynthesis prediction using an end-to-end graph generative architecture for molecular graph editing},
  author={Zhong, Weihe and Yang, Ziduo and Chen, Calvin Yu-Chian},
  journal={Nature Communications},
  volume={14},
  number={1},
  pages={3009},
  year={2023},
  publisher={Nature Publishing Group UK London}
}

@article{zheng2019predicting,
  title={Predicting retrosynthetic reactions using self-corrected transformer neural networks},
  author={Zheng, Shuangjia and Rao, Jiahua and Zhang, Zhongyue and Xu, Jun and Yang, Yuedong},
  journal={Journal of chemical information and modeling},
  volume={60},
  number={1},
  pages={47--55},
  year={2019},
  publisher={ACS Publications}
}

@article{sacha2021molecule,
  title={Molecule edit graph attention network: modeling chemical reactions as sequences of graph edits},
  author={Sacha, Miko{\l}aj and B{\l}az, Miko{\l}aj and Byrski, Piotr and Dabrowski-Tumanski, Pawe{\l} and Chrominski, Miko{\l}aj and Loska, Rafa{\l} and W{\l}odarczyk-Pruszynski, Pawe{\l} and Jastrzebski, Stanis{\l}aw},
  journal={Journal of Chemical Information and Modeling},
  volume={61},
  number={7},
  pages={3273--3284},
  year={2021},
  publisher={ACS Publications}
}

@article{tu2022permutation,
  title={Permutation invariant graph-to-sequence model for template-free retrosynthesis and reaction prediction},
  author={Tu, Zhengkai and Coley, Connor W},
  journal={Journal of chemical information and modeling},
  volume={62},
  number={15},
  pages={3503--3513},
  year={2022},
  publisher={ACS Publications}
}

@article{zhong2022root,
  title={Root-aligned SMILES: a tight representation for chemical reaction prediction},
  author={Zhong, Zipeng and Song, Jie and Feng, Zunlei and Liu, Tiantao and Jia, Lingxiang and Yao, Shaolun and Wu, Min and Hou, Tingjun and Song, Mingli},
  journal={Chemical Science},
  volume={13},
  number={31},
  pages={9023--9034},
  year={2022},
  publisher={Royal Society of Chemistry}
}

@article{yao2024node,
  title={Node-aligned graph-to-graph: elevating template-free deep learning approaches in single-step retrosynthesis},
  author={Yao, Lin and Guo, Wentao and Wang, Zhen and Xiang, Shang and Liu, Wentan and Ke, Guolin},
  journal={JACS Au},
  volume={4},
  number={3},
  pages={992--1003},
  year={2024},
  publisher={ACS Publications}
}

@article{deng2025rsgpt,
  title={RSGPT: a generative transformer model for retrosynthesis planning pre-trained on ten billion datapoints},
  author={Deng, Yafeng and Zhao, Xinda and Sun, Hanyu and Chen, Yu and Wang, Xiaorui and Xue, Xi and Li, Liangning and Song, Jianfei and Hsieh, Chang-Yu and Hou, Tingjun and others},
  journal={Nature communications},
  volume={16},
  number={1},
  pages={7012},
  year={2025},
  publisher={Nature Publishing Group UK London}
}

@article{schneider2016big,
  title={Big data from pharmaceutical patents: a computational analysis of medicinal chemists’ bread and butter},
  author={Schneider, Nadine and Lowe, Daniel M and Sayle, Roger A and Tarselli, Michael A and Landrum, Gregory A},
  journal={Journal of medicinal chemistry},
  volume={59},
  number={9},
  pages={4385--4402},
  year={2016},
  publisher={ACS Publications}
}

@article{mysinger2012directory,
  title={Directory of useful decoys, enhanced (DUD-E): better ligands and decoys for better benchmarking},
  author={Mysinger, Michael M and Carchia, Michael and Irwin, John J and Shoichet, Brian K},
  journal={Journal of medicinal chemistry},
  volume={55},
  number={14},
  pages={6582--6594},
  year={2012},
  publisher={ACS Publications}
}

@article{xiong2025bridging,
  title={Bridging three-dimensional molecular structures and artificial intelligence with a conformation description language},
  author={Xiong, Jiacheng and Shi, Yuqi and Wu, Min and Shao, Panpan and Wang, Zhaokun and Zhang, Wei and Zhang, Runze and Chen, Zhiyi and Zeng, Chuanlong and Jiang, Xun and others},
  journal={Nature Machine Intelligence},
  pages={1--15},
  year={2026},
  publisher={Nature Publishing Group}
}

@misc{vainio2009shaep,
  title={ShaEP: molecular overlay based on shape and electrostatic potential},
  author={Vainio, Mikko J and Puranen, J Santeri and Johnson, Mark S},
  year={2009},
  publisher={ACS Publications}
}

@article{sastry2011rapid,
  title={Rapid shape-based ligand alignment and virtual screening method based on atom/feature-pair similarities and volume overlap scoring},
  author={Sastry, G Madhavi and Dixon, Steven L and Sherman, Woody},
  journal={Journal of chemical information and modeling},
  volume={51},
  number={10},
  pages={2455--2466},
  year={2011},
  publisher={ACS Publications}
}

@article{liu2011shafts,
  title={SHAFTS: a hybrid approach for 3D molecular similarity calculation. 1. Method and assessment of virtual screening},
  author={Liu, Xiaofeng and Jiang, Hualiang and Li, Honglin},
  journal={Journal of chemical information and modeling},
  volume={51},
  number={9},
  pages={2372--2385},
  year={2011},
  publisher={ACS Publications}
}

@article{hu2018ls,
  title={LS-align: an atom-level, flexible ligand structural alignment algorithm for high-throughput virtual screening},
  author={Hu, Jun and Liu, Zi and Yu, Dong-Jun and Zhang, Yang},
  journal={Bioinformatics},
  volume={34},
  number={13},
  pages={2209--2218},
  year={2018},
  publisher={Oxford University Press}
}

@article{devinyak20143d,
  title={3D-MoRSE descriptors explained},
  author={Devinyak, Oleg and Havrylyuk, Dmytro and Lesyk, Roman},
  journal={Journal of Molecular Graphics and Modelling},
  volume={54},
  pages={194--203},
  year={2014},
  publisher={Elsevier}
}

@article{axen2017simple,
  title={A simple representation of three-dimensional molecular structure},
  author={Axen, Seth D and Huang, Xi-Ping and C{\'a}ceres, Elena L and Gendelev, Leo and Roth, Bryan L and Keiser, Michael J},
  journal={Journal of medicinal chemistry},
  volume={60},
  number={17},
  pages={7393--7409},
  year={2017},
  publisher={ACS Publications}
}

@article{hemmer1999deriving,
  title={Deriving the 3D structure of organic molecules from their infrared spectra},
  author={Hemmer, Markus C and Steinhauer, Valentin and Gasteiger, Johann},
  journal={Vibrational spectroscopy},
  volume={19},
  number={1},
  pages={151--164},
  year={1999},
  publisher={Elsevier}
}

@article{klein2004topological,
  title={Topological distance based 3D descriptors for use in QSAR and diversity analysis},
  author={Klein, Christian T and Kaiser, Dominik and Ecker, Gerhard},
  journal={Journal of chemical information and computer sciences},
  volume={44},
  number={1},
  pages={200--209},
  year={2004},
  publisher={ACS Publications}
}

@article{ashburner2000go,
  title={Gene ontology: tool for the unification of biology},
  author={Ashburner, Michael and Ball, Catherine A and Blake, Judith A and Botstein, David and Butler, Heather and Cherry, J Michael and Davis, Allan P and Dolinski, Kara and Dwight, Selina S and Eppig, Janan T and others},
  journal={Nature genetics},
  volume={25},
  number={1},
  pages={25--29},
  year={2000},
  publisher={Nature Publishing Group}
}

@article{gene2021gene,
  title={The Gene Ontology resource: enriching a GOld mine},
  journal={Nucleic acids research},
  volume={49},
  number={D1},
  pages={D325--D334},
  year={2021},
  publisher={Oxford University Press}
}

@article{friedberg2006automated,
  title={Automated protein function prediction—the genomic challenge},
  author={Friedberg, Iddo},
  journal={Briefings in bioinformatics},
  volume={7},
  number={3},
  pages={225--242},
  year={2006},
  publisher={Oxford University Press}
}

@article{radivojac2013large,
  title={A large-scale evaluation of computational protein function prediction},
  author={Radivojac, Predrag and Clark, Wyatt T and Oron, Tal Ronnen and Schnoes, Alexandra M and Wittkop, Tobias and Sokolov, Artem and Graim, Kiley and Funk, Christopher and Verspoor, Karin and Ben-Hur, Asa and others},
  journal={Nature methods},
  volume={10},
  number={3},
  pages={221--227},
  year={2013},
  publisher={Nature Publishing Group US New York}
}

@article{cantalapiedra2021eggnog,
  title={eggNOG-mapper v2: functional annotation, orthology assignments, and domain prediction at the metagenomic scale},
  author={Cantalapiedra, Carlos P and Hern{\'a}ndez-Plaza, Ana and Letunic, Ivica and Bork, Peer and Huerta-Cepas, Jaime},
  journal={Molecular biology and evolution},
  volume={38},
  number={12},
  pages={5825--5829},
  year={2021},
  publisher={Oxford University Press}
}

@article{steinegger2017mmseqs2,
  title={MMseqs2 enables sensitive protein sequence searching for the analysis of massive data sets},
  author={Steinegger, Martin and S{\"o}ding, Johannes},
  journal={Nature biotechnology},
  volume={35},
  number={11},
  pages={1026--1028},
  year={2017},
  publisher={Nature Publishing Group US New York}
}

@article{10.3389/fphar.2020.565644,
  title={{M}olecular {S}ets ({MOSES}): {A} {B}enchmarking {P}latform for {M}olecular {G}eneration {M}odels},
  author={Polykovskiy, Daniil and Zhebrak, Alexander and Sanchez-Lengeling, Benjamin and Golovanov, Sergey and Tatanov, Oktai and Belyaev, Stanislav and Kurbanov, Rauf and Artamonov, Aleksey and Aladinskiy, Vladimir and Veselov, Mark and Kadurin, Artur and Johansson, Simon and  Chen, Hongming and Nikolenko, Sergey and Aspuru-Guzik, Alan and Zhavoronkov, Alex},
  journal={Frontiers in Pharmacology},
  year={2020}
}

@article{wu2018moleculenet,
  title={MoleculeNet: a benchmark for molecular machine learning},
  author={Wu, Zhenqin and Ramsundar, Bharath and Feinberg, Evan N and Gomes, Joseph and Geniesse, Caleb and Pappu, Aneesh S and Leswing, Karl and Pande, Vijay},
  journal={Chemical science},
  volume={9},
  number={2},
  pages={513--530},
  year={2018},
  publisher={Royal Society of Chemistry}
}

@misc{nvidia2025nvidianemotronnano2,
      title={NVIDIA Nemotron Nano 2: An Accurate and Efficient Hybrid Mamba-Transformer Reasoning Model}, 
      author={NVIDIA and : and Aarti Basant and Abhijit Khairnar and Abhijit Paithankar and Abhinav Khattar and Adithya Renduchintala and Aditya Malte and Akhiad Bercovich and Akshay Hazare and Alejandra Rico and Aleksander Ficek and Alex Kondratenko and Alex Shaposhnikov and Alexander Bukharin and Ali Taghibakhshi and Amelia Barton and Ameya Sunil Mahabaleshwarkar and Amy Shen and Andrew Tao and Ann Guan and Anna Shors and Anubhav Mandarwal and Arham Mehta and Arun Venkatesan and Ashton Sharabiani and Ashwath Aithal and Ashwin Poojary and Ayush Dattagupta and Balaram Buddharaju and Banghua Zhu and Barnaby Simkin and Bilal Kartal and Bita Darvish Rouhani and Bobby Chen and Boris Ginsburg and Brandon Norick and Brian Yu and Bryan Catanzaro and Charles Wang and Charlie Truong and Chetan Mungekar and Chintan Patel and Chris Alexiuk and Christian Munley and Christopher Parisien and Dan Su and Daniel Afrimi and Daniel Korzekwa and Daniel Rohrer and Daria Gitman and David Mosallanezhad and Deepak Narayanan and Dima Rekesh and Dina Yared and Dmytro Pykhtar and Dong Ahn and Duncan Riach and Eileen Long and Elliott Ning and Eric Chung and Erick Galinkin and Evelina Bakhturina and Gargi Prasad and Gerald Shen and Haifeng Qian and Haim Elisha and Harsh Sharma and Hayley Ross and Helen Ngo and Herman Sahota and Hexin Wang and Hoo Chang Shin and Hua Huang and Iain Cunningham and Igor Gitman and Ivan Moshkov and Jaehun Jung and Jan Kautz and Jane Polak Scowcroft and Jared Casper and Jian Zhang and Jiaqi Zeng and Jimmy Zhang and Jinze Xue and Jocelyn Huang and Joey Conway and John Kamalu and Jonathan Cohen and Joseph Jennings and Julien Veron Vialard and Junkeun Yi and Jupinder Parmar and Kari Briski and Katherine Cheung and Katherine Luna and Keith Wyss and Keshav Santhanam and Kezhi Kong and Krzysztof Pawelec and Kumar Anik and Kunlun Li and Kushan Ahmadian and Lawrence McAfee and Laya Sleiman and Leon Derczynski and Luis Vega and Maer Rodrigues de Melo and Makesh Narsimhan Sreedhar and Marcin Chochowski and Mark Cai and Markus Kliegl and Marta Stepniewska-Dziubinska and Matvei Novikov and Mehrzad Samadi and Meredith Price and Meriem Boubdir and Michael Boone and Michael Evans and Michal Bien and Michal Zawalski and Miguel Martinez and Mike Chrzanowski and Mohammad Shoeybi and Mostofa Patwary and Namit Dhameja and Nave Assaf and Negar Habibi and Nidhi Bhatia and Nikki Pope and Nima Tajbakhsh and Nirmal Kumar Juluru and Oleg Rybakov and Oleksii Hrinchuk and Oleksii Kuchaiev and Oluwatobi Olabiyi and Pablo Ribalta and Padmavathy Subramanian and Parth Chadha and Pavlo Molchanov and Peter Dykas and Peter Jin and Piotr Bialecki and Piotr Januszewski and Pradeep Thalasta and Prashant Gaikwad and Prasoon Varshney and Pritam Gundecha and Przemek Tredak and Rabeeh Karimi Mahabadi and Rajen Patel and Ran El-Yaniv and Ranjit Rajan and Ria Cheruvu and Rima Shahbazyan and Ritika Borkar and Ritu Gala and Roger Waleffe and Ruoxi Zhang and Russell J. Hewett and Ryan Prenger and Sahil Jain and Samuel Kriman and Sanjeev Satheesh and Saori Kaji and Sarah Yurick and Saurav Muralidharan and Sean Narenthiran and Seonmyeong Bak and Sepehr Sameni and Seungju Han and Shanmugam Ramasamy and Shaona Ghosh and Sharath Turuvekere Sreenivas and Shelby Thomas and Shizhe Diao and Shreya Gopal and Shrimai Prabhumoye and Shubham Toshniwal and Shuoyang Ding and Siddharth Singh and Siddhartha Jain and Somshubra Majumdar and Soumye Singhal and Stefania Alborghetti and Syeda Nahida Akter and Terry Kong and Tim Moon and Tomasz Hliwiak and Tomer Asida and Tony Wang and Tugrul Konuk and Twinkle Vashishth and Tyler Poon and Udi Karpas and Vahid Noroozi and Venkat Srinivasan and Vijay Korthikanti and Vikram Fugro and Vineeth Kalluru and Vitaly Kurin and Vitaly Lavrukhin and Wasi Uddin Ahmad and Wei Du and Wonmin Byeon and Ximing Lu and Xin Dong and Yashaswi Karnati and Yejin Choi and Yian Zhang and Ying Lin and Yonggan Fu and Yoshi Suhara and Zhen Dong and Zhiyu Li and Zhongbo Zhu and Zijia Chen},
      year={2025},
      eprint={2508.14444},
      archivePrefix={arXiv},
      primaryClass={cs.CL},
      url={https://arxiv.org/abs/2508.14444}, 
}

@misc{olmo2025olmo3,
title={Olmo 3},
author={Team Olmo and Allyson Ettinger and Amanda Bertsch and Bailey Kuehl and David Graham and David Heineman and Dirk Groeneveld and Faeze Brahman and Finbarr Timbers and Hamish Ivison and Jacob Morrison and Jake Poznanski and Kyle Lo and Luca Soldaini and Matt Jordan and Mayee Chen and Michael Noukhovitch and Nathan Lambert and Pete Walsh and Pradeep Dasigi and Robert Berry and Saumya Malik and Saurabh Shah and Scott Geng and Shane Arora and Shashank Gupta and Taira Anderson and Teng Xiao and Tyler Murray and Tyler Romero and Victoria Graf and Akari Asai and Akshita Bhagia and Alexander Wettig and Alisa Liu and Aman Rangapur and Chloe Anastasiades and Costa Huang and Dustin Schwenk and Harsh Trivedi and Ian Magnusson and Jaron Lochner and Jiacheng Liu and Lester James V. Miranda and Maarten Sap and Malia Morgan and Michael Schmitz and Michal Guerquin and Michael Wilson and Regan Huff and Ronan Le Bras and Rui Xin and Rulin Shao and Sam Skjonsberg and Shannon Zejiang Shen and Shuyue Stella Li and Tucker Wilde and Valentina Pyatkin and Will Merrill and Yapei Chang and Yuling Gu and Zhiyuan Zeng and Ashish Sabharwal and Luke Zettlemoyer and Pang Wei Koh and Ali Farhadi and Noah A. Smith and Hannaneh Hajishirzi},
year={2025},
eprint={2512.13961},
archivePrefix={arXiv},
primaryClass={cs.CL},
url={https://arxiv.org/abs/2512.13961},
}

@article{su2026sciif,
  title={SciIF: Benchmarking Scientific Instruction Following Towards Rigorous Scientific Intelligence},
  author={Su, Encheng and Wu, Jianyu and Tang, Chen and Wang, Lintao and Li, Pengze and Wang, Aoran and Zhang, Jinouwen and Wang, Yizhou and Meng, Yuan and Ma, Xinzhu and others},
  journal={arXiv preprint arXiv:2601.04770},
  year={2026}
}

@article{thakur2026composition,
  title={Composition--structure--property relationships in MXenes},
  author={Thakur, Anupma and Kim, Jongyoun and Wyatt, Brian C and Gogotsi, Yury and Anasori, Babak},
  journal={Nature Reviews Materials},
  pages={1--16},
  year={2026},
  publisher={Nature Publishing Group UK London}
}

@article{le2012quantitative,
  title={Quantitative structure--property relationship modeling of diverse materials properties},
  author={Le, Tu and Epa, V Chandana and Burden, Frank R and Winkler, David A},
  journal={Chemical reviews},
  volume={112},
  number={5},
  pages={2889--2919},
  year={2012},
  publisher={ACS Publications}
}

@article{van2008complete,
  title={A complete representation of structure--property relationships in crystals},
  author={Van de Walle, Axel},
  journal={Nature materials},
  volume={7},
  number={6},
  pages={455--458},
  year={2008},
  publisher={Nature Publishing Group UK London}
}

@book{creighton1993proteins,
  title={Proteins: structures and molecular properties},
  author={Creighton, Thomas E},
  year={1993},
  publisher={Macmillan}
}

@article{tai2022leveraging,
  title={Leveraging isothermal titration calorimetry to explore structure--property relationships of protein immobilization in metal--organic frameworks},
  author={Tai, Tzu-Yi and Sha, Fanrui and Wang, Xiaoliang and Wang, Xingjie and Ma, Kaikai and Kirlikovali, Kent O and Su, Shengyi and Islamoglu, Timur and Kato, Satoshi and Farha, Omar K},
  journal={Angewandte Chemie International Edition},
  volume={61},
  number={37},
  pages={e202209110},
  year={2022},
  publisher={Wiley Online Library}
}

@article{osadchy2011maps,
  title={Maps of protein structure space reveal a fundamental relationship between protein structure and function},
  author={Osadchy, Margarita and Kolodny, Rachel},
  journal={Proceedings of the National Academy of Sciences},
  volume={108},
  number={30},
  pages={12301--12306},
  year={2011},
  publisher={National Academy of Sciences}
}

@article{narayanan2026ether0,
  title={Training a scientific reasoning model for chemistry},
  author={Narayanan, Siddharth and Braza, James and Griffiths, Ryan-Rhys and Bou, Albert and Wellawatte, Geemi and Caldas Ramos, Mayk and Mitchener, Ludovico and Pieler, Michael and Rodriques, Sam and White, Andrew},
  journal={Advances in Neural Information Processing Systems},
  volume={38},
  pages={157671--157710},
  year={2026}
}

@article{wang2022molclr,
  title={Molecular contrastive learning of representations via graph neural networks},
  author={Wang, Yuyang and Wang, Jianren and Cao, Zhonglin and Barati Farimani, Amir},
  journal={Nature Machine Intelligence},
  pages={1--9},
  year={2022},
  publisher={Nature Publishing Group},
  doi={10.1038/s42256-022-00447-x}
}

@article{suresh2015rpi,
  title={RPI-Pred: predicting ncRNA-protein interaction using sequence and structural information},
  author={Suresh, V and Liu, Liang and Adjeroh, Donald and Zhou, Xiaobo},
  journal={Nucleic acids research},
  volume={43},
  number={3},
  pages={1370--1379},
  year={2015},
  publisher={Oxford University Press}
}

@article{zhang2024multiple,
  title={Multiple sequence alignment-based RNA language model and its application to structural inference},
  author={Zhang, Yikun and Lang, Mei and Jiang, Jiuhong and Gao, Zhiqiang and Xu, Fan and Litfin, Thomas and Chen, Ke and Singh, Jaswinder and Huang, Xiansong and Song, Guoli and others},
  journal={Nucleic acids research},
  volume={52},
  number={1},
  pages={e3--e3},
  year={2024},
  publisher={Oxford University Press}
}

@article{bogard2019deep,
  title={A deep neural network for predicting and engineering alternative polyadenylation},
  author={Bogard, Nicholas and Linder, Johannes and Rosenberg, Alexander B and Seelig, Georg},
  journal={Cell},
  volume={178},
  number={1},
  pages={91--106},
  year={2019},
  publisher={Elsevier}
}

@article{almagro2017deeploc,
  title={DeepLoc: prediction of protein subcellular localization using deep learning},
  author={Almagro Armenteros, Jos{\'e} Juan and S{\o}nderby, Casper Kaae and S{\o}nderby, S{\o}ren Kaae and Nielsen, Henrik and Winther, Ole},
  journal={Bioinformatics},
  volume={33},
  number={21},
  pages={3387--3395},
  year={2017},
  publisher={Oxford University Press}
}

@article{dalla2025nucleotide,
  title={Nucleotide transformer: building and evaluating robust foundation models for human genomics},
  author={Dalla-Torre, Hugo and Gonzalez, Liam and Mendoza-Revilla, Javier and Lopez Carranza, Nicolas and Grzywaczewski, Adam Henryk and Oteri, Francesco and Dallago, Christian and Trop, Evan and De Almeida, Bernardo P and Sirelkhatim, Hassan and others},
  journal={Nature Methods},
  volume={22},
  number={2},
  pages={287--297},
  year={2025},
  publisher={Nature Publishing Group US New York}
}

\end{document}